\documentclass{article}

\PassOptionsToPackage{numbers, compress}{natbib}
\usepackage[final,main]{neurips_2026}

\usepackage[utf8]{inputenc} 
\usepackage[T1]{fontenc}    
\usepackage{url}           
\usepackage{booktabs}      
\usepackage{amsfonts}   
\usepackage{nicefrac}     
\usepackage{microtype}    
\usepackage[table]{xcolor}  
\usepackage{wrapfig}
\definecolor{my-slate-blue}{RGB}{56, 84, 146}
\definecolor{meangray}{gray}{0.92}
\definecolor{oursrow}{HTML}{FEF5F5}
\definecolor{linkpink}{HTML}{E6007E}

\usepackage[
  colorlinks=true,         
  linkcolor=my-slate-blue, 
  citecolor=my-slate-blue, 
  urlcolor=my-slate-blue   ,
  hyperfootnotes=false
]{hyperref}
\usepackage{enumitem}
\usepackage{amsmath}
\usepackage{amsthm}
\usepackage{amssymb}
\usepackage{makecell}
\usepackage{graphicx}
\usepackage{multirow}
\usepackage{caption}
\usepackage{longtable}
\usepackage{subcaption}
\usepackage{xspace}
\usepackage[font=small]{caption}
\usepackage[most]{tcolorbox}
\usepackage[export]{adjustbox}
\usepackage{fontawesome5}

\newcommand{\method}{DynaFLIP\xspace}
\xspaceaddexceptions{'}

\title{\method: Rethinking Robotics Perception via Tri-Modal-Dynamics Guided Representation}

\author{
  Jusuk Lee $^{1}$
  \quad\ Seungjae Lee$^{2}$ 
  \quad\ Jonghun Shin$^{1}$ 
  \quad Hoseong Jung$^{1}$ 
  \quad Sungha Kim$^{1}$\\
  \textbf{Daesol Cho$^{3}$ 
  \quad H. Jin Kim$^{1}$
  \quad Jia-Bin Huang$^{2,\dagger}$
  \quad Furong Huang$^{2, 4, \dagger}$}
  \vspace{.8em}  \\
  $^1$ Seoul National University 
  \quad $^2$ University of Maryland, College Park\\
  $^3$ Georgia Institute of Technology
  \quad $^4$ All Purpose AI
  \vspace{.8em} 
  \\
  {\hypersetup{urlcolor=linkpink}\url{https://dynaflip-robotics.github.io}}
}

\begin{document}
\raggedbottom

\maketitle
\renewcommand{\thefootnote}{}\footnotetext{$\dagger$ Equal Advising.}\renewcommand{\thefootnote}{\arabic{footnote}}

% Disable ToC
\let\oldaddcontentsline\addcontentsline
\renewcommand{\addcontentsline}[3]{}

\begin{figure}[ht]
\vspace{-0.5cm}
\centering
\includegraphics[width=1.0\textwidth]{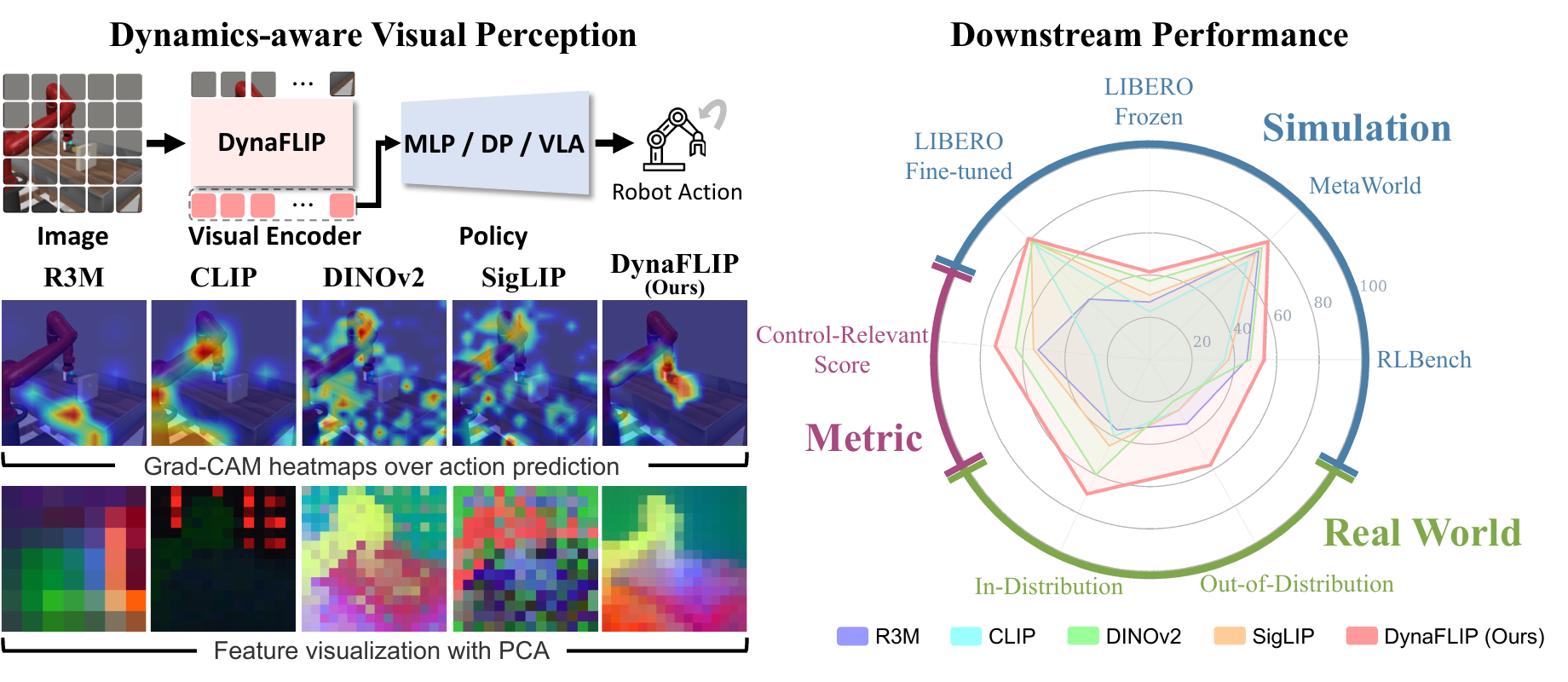}
\caption{\textbf{\method learns dynamics-aware visual representations that focus on control-relevant regions and capture spatially coherent structure, leading to strong downstream performance.} \method serves as a visual backbone for diverse downstream policies (MLP, diffusion policy, VLA). 
Grad-CAM shows \method attending to manipulated objects and interaction regions, while PCA reveals coherent object-level structure. 
\method outperforms baselines across simulation, real-world tasks, and the control-relevant metric.}
\label{fig:thumbnail}
% \vspace{-0.5em}
\end{figure}

\begin{abstract}
Robot manipulation critically depends on perception that preserves the action-relevant aspects of a scene.
Yet most robot learning pipelines are built upon visual encoders pre-trained for static recognition or vision-language alignment, leaving motion understanding to downstream policies. 
We introduce \method, a dynamics-aware multimodal pre-training framework that pushes motion understanding upstream into perception. 
We construct image--language--3D flow triplets from heterogeneous human and robot videos, and use these triplets as training-time supervision to shape an image-only encoder. 
Our key idea is to encourage the three modalities to span a small simplex volume in the shared hyperspherical space---a smaller simplex volume indicating stronger alignment. 
To avoid the geometric ambiguity and trivial collapse of naive volume minimization, we combine simplex-volume minimization with a cosine regularizer and a contrastive objective. 
Our analyses show that \method focuses on control-relevant regions critical for manipulation. 
The resulting dynamics-aware representations serve as reusable visual backbones and consistently outperform baselines across diverse downstream policies, including VLAs. 
We validate this across diverse simulation and real-world setups, with gains reaching +22.5\% under out-of-distribution scenarios. 
Our results suggest that robot generalization improves when visual representations are trained to encode not just what is present, but how the world changes under action.
\end{abstract}

\section{Introduction}
\label{sec:intro}

A central goal of robot learning is to build agents that generalize across diverse real-world environments and tasks---new objects, backgrounds, and distractors. 
Recent robot learning systems increasingly pursue this goal by reusing powerful vision encoders such as CLIP, SigLIP, and DINOv2~\cite{radford2021learning, zhai2023sigmoid, oquab2024dinov2} inside diverse policies, ranging from imitation learning to Vision-Language-Action (VLA) models~\cite{kim2025openvla, lee2025tracegen, bu2025univla, black2024pi_0, intelligence2025pi_}. 
This practice inherits a key assumption: perception can be borrowed from encoders pre-trained for mainstream computer-vision objectives, while motion and dynamics are handled mainly by downstream planning or control. 
We argue that this assumption fundamentally limits robot generalization. 
In particular, manipulation is about how actions induce state transitions, yet existing visual encoders are not exposed to motion and dynamics during pre-training. 
As a result, they often attend to visually salient but control-irrelevant regions instead of the manipulated object or contact area. 
We therefore rethink the robotic pipeline by pushing \emph{dynamics awareness} upstream into perception, so that visual encoders represent not only what is in the scene, but also how the scene changes under action.

The challenge is then how to inject dynamics awareness into a visual encoder when the encoder ultimately operates on a single image at test time. Images alone do not always reveal which aspects of a scene are causally relevant for action, whereas other modalities can provide complementary evidence about intended and realized state changes. This suggests using such modalities not as additional inputs at test time, but as supervision to shape the visual encoder's representation during training. In this work, we focus on three such modalities, each contributing information that the others cannot. \emph{Image transitions} provide the most direct visual evidence of what changed between states, but cannot explain why a change occurred. \emph{Language} fills this gap by describing the intended transition at a semantic level. \emph{3D flow} then adds what neither image transitions nor language can provide: an explicit, viewpoint-invariant account of how the scene moves in physical space, decoupled from 2D appearance. We deliberately select these three modalities because all of them can be extracted from action-free video data, allowing pre-training to leverage large-scale human and robot videos rather than the limited robot-collected datasets.

With the three modalities identified, the remaining challenge is how to transfer their supervisory signal into the latent space of an image-only encoder. Standard anchor-based multimodal objectives~\cite{girdhar2023imagebind, zhu2023languagebind, 
ruan2023accommodating}---even when the image serves as the anchor---do not ensure mutual alignment among the remaining modalities. 
An alternative strategy, inspired by prior work in multimodal retrieval~\cite{yin2026towards, cicchettigramian, cicchetti2025triangle}, is to constrain all modality embeddings jointly through the simplex they span. However, naive simplex-volume minimization is itself prone to two pitfalls. First, geometric ambiguity: a low-volume simplex does not guarantee mutual alignment, since the simplex volume can shrink even when some modality pairs remain far apart. Second, trivial collapse: in the absence of negative tuples, the simplex volume is minimized when all modality embeddings collapse to a single point. A useful robotics representation must therefore exploit higher-order multimodal geometry to learn a coherent, control-relevant visual latent space, while avoiding these degeneracies.

In this paper, we propose \method, a \textbf{Dyna}mics-aware 3D \textbf{F}low-\textbf{L}anguage-\textbf{I}mage \textbf{P}re-training framework that uses image transitions, language, and 3D flow as training-time supervision to shape the latent space of an image-only encoder, yielding control-relevant visual representations for downstream manipulation. Building on simplex-based alignment~\cite{yin2026towards, cicchettigramian, cicchetti2025triangle}, we minimize the volume of the simplex spanned by the three modalities in a shared embedding space (a triangle area in our three-modal setting). To address the two pitfalls of naive simplex-volume minimization, we resolve geometric ambiguity through a cosine regularizer between selected modality pairs, and prevent trivial collapse by embedding the cosine-augmented energy in an InfoNCE-style contrastive framework~\cite{oord2018representation}. We further introduce two auxiliary objectives---a temporal contrastive loss and an actor loss---to reinforce trajectory-level temporal structure and strengthen dynamics-aware visual representations. Extensive experiments in both simulation and real-world environments show that the resulting encoder outperforms strong baselines, transfers effectively as a visual backbone across diverse downstream policies, and is especially robust under out-of-distribution variations.

In summary, our contributions are threefold: \textbf{(i)} We recast robot generalization partly as a perception problem: robust manipulation requires visual representations that encode dynamics- and control-relevant structure, rather than merely what is most visually salient. \textbf{(ii)} We introduce \method that distills supervision from image transitions, language, and 3D flow into an image-only encoder through higher-order multimodal alignment while preventing geometric ambiguity and trivial collapse. \textbf{(iii)} We construct image--language--3D flow triplets from human and robot videos and show that \method transfers strongly as a reusable backbone across simulation and real-world manipulation, achieving up to 22.5\% improvement over the strongest baseline under real-world OOD perturbations.

\section{Method}

\begin{figure*}[t]
\centering
\includegraphics[width=1.0\textwidth]{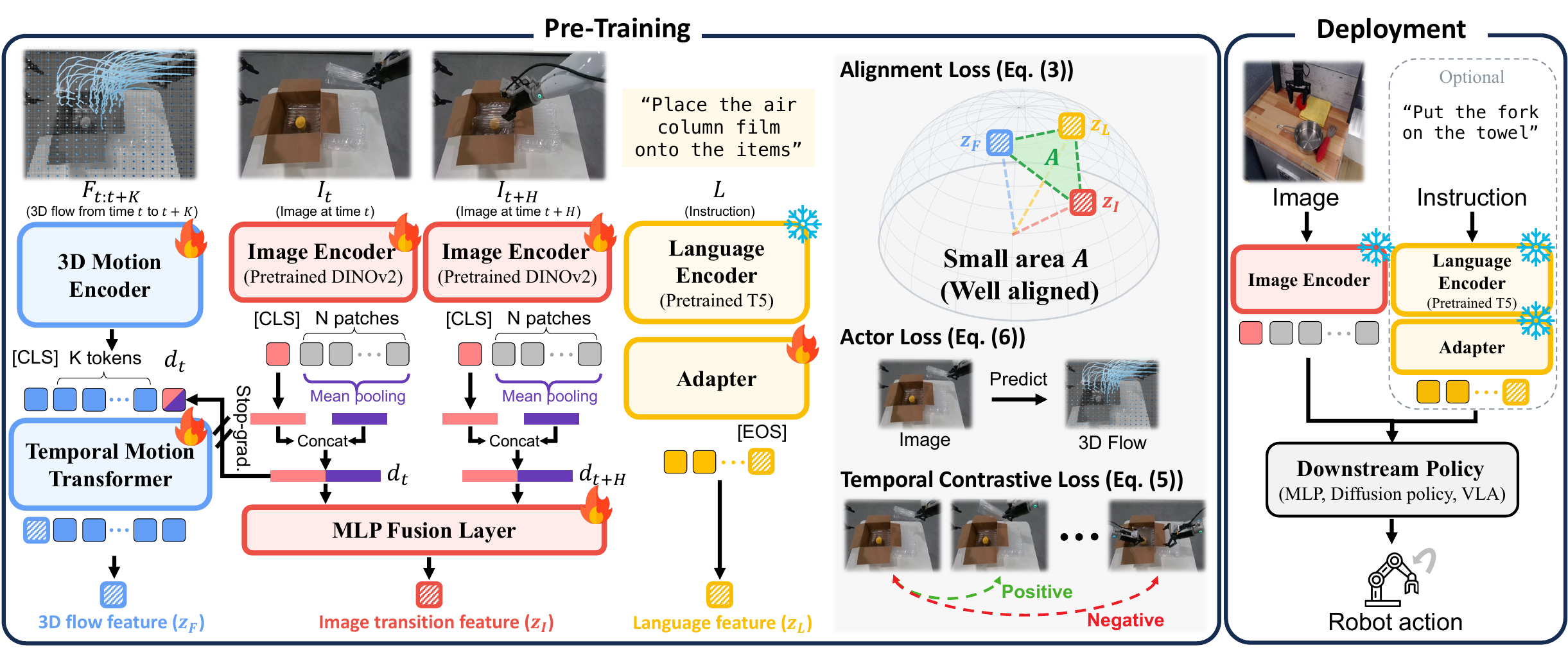}
\caption{\textbf{Overview of \method.} 
Three modalities are encoded into embeddings in a shared hyperspherical space. 
The image encoder (initialized from DINOv2 and fully fine-tuned) produces per-frame features from $I_t, I_{t+H}$ via CLS and mean-pooled patch tokens, which are then fused into $z_I$. 
A frozen T5 with a learnable adapter produces $z_L$ from the EOS token of $L$, and a 3D flow encoder produces $z_F$ from $F_{t:t+K}$. 
The alignment loss minimizes the area $A$ spanned by these embeddings, with auxiliary actor and temporal contrastive losses reinforcing dynamics-aware representations. 
Our pre-trained image encoder serves as a visual backbone for diverse downstream policies, with the language encoder optionally included for instruction-conditioned policies.}
\label{fig:architecture}
\vspace{-0.5em}
\end{figure*}

\method shifts visual pre-training from static scene understanding to motion-induced state transitions.
Section~\ref{sec:simplex_guided_multimodal_learning} introduces a simplex-guided multimodal alignment objective that aligns image transitions, language, and 3D flow into a shared embedding space while resolving two optimization pitfalls: geometric ambiguity and trivial collapse. Section~\ref{sec:full_pretraining_objective} then presents the auxiliary objectives---temporal contrastive and actor losses---that further strengthen dynamics-aware visual representations. Finally, Section~\ref{sec:dataset_construction} describes how we construct large-scale image--language--3D flow triplets from human and robot videos.
\subsection{Simplex-Guided Multimodal Alignment for Dynamics-Aware Representation}
\label{sec:simplex_guided_multimodal_learning}

% Goal + 3 modalities
We aim to learn dynamics-aware visual representations by aligning three transition-based modalities---\emph{image transitions, language, and 3D flow}. Image transitions capture visual state changes, language specifies the intended transition at a semantic level, and 3D flow encodes physical motion in the scene. We map each modality to an $\ell_2$-normalized embedding on the unit sphere: $z_I$ for the image transition, $z_L$ for the language, and $z_F$ for the 3D flow.

% Background + simplex volume -> triangle
A common strategy for aligning multiple modalities is anchor-based contrastive learning, where one modality serves as a reference and each auxiliary modality is independently aligned to it~\cite{girdhar2023imagebind, zhu2023languagebind, ruan2023accommodating}. However, this design enforces pairwise alignment only with the anchor and does not constrain the non-anchor modalities relative to each other. To capture mutual alignment among all three modalities, we adopt a simplex-volume-based formulation~\cite{yin2026towards, cicchettigramian, cicchetti2025triangle}. For an $m$-modal tuple of $\ell_2$-normalized embeddings, the generalized simplex volume $\mathcal{V}_m$ measures the volume of the simplex spanned by the embeddings in the shared latent space, with smaller $\mathcal{V}_m$ indicating stronger joint alignment. In our three-modal setting, $\mathcal{V}_m$ reduces to the \textit{triangle} area
\begin{equation}
    \mathcal{V}_3(z_L, z_I, z_F) = A(z_L, z_I, z_F) = \frac{1}{2}\sqrt
    {\langle u, u \rangle \langle v, v \rangle - \langle u, v \rangle^2}, 
    \quad u = z_I - z_L, \, v = z_F - z_L,
    \label{eq:triangle_area}
\end{equation}
spanned by the three embeddings. A small triangle area thus indicates joint alignment among all three modalities, capturing higher-order multimodal geometry beyond anchor-based pairwise alignment. The general $m$-modal formulation is provided in Appendix~\ref{app:simplex_volume_definition}.

\begin{figure*}[t!]
\centering
\includegraphics[width=1.0\textwidth]{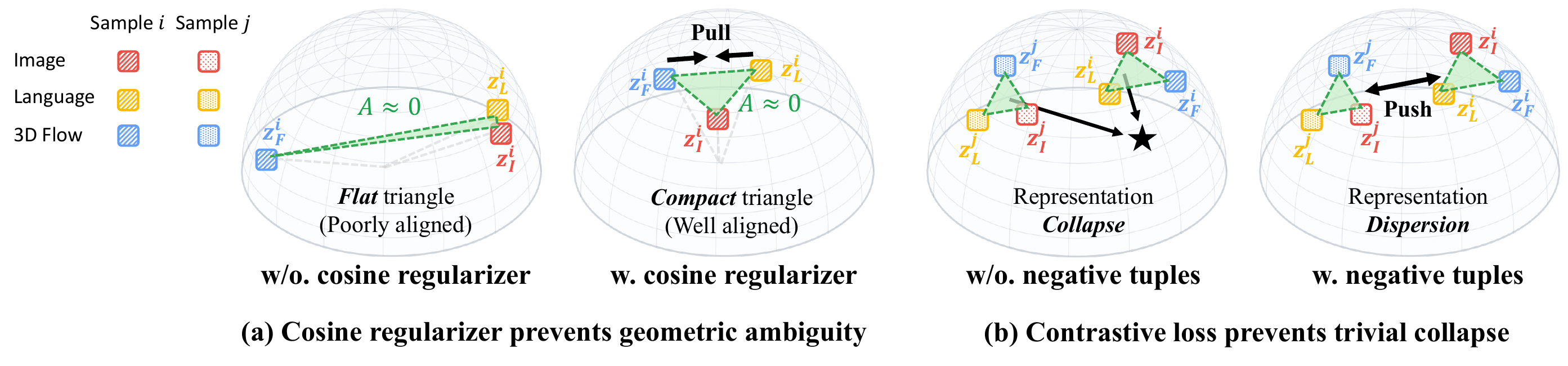}
\caption{
    \textbf{Two optimization pitfalls of na\"ive simplex-volume minimization.} 
    \textbf{(a) Geometric ambiguity.} A flat triangle has near-zero area even when one modality remains far from the other two. 
    The cosine regularizer pulls selected modality pairs together, yielding a desired alignment (see Eq.~\eqref{eq:three_modal_energy}). 
    \textbf{(b) Trivial collapse.} Without negative tuples, all modality embeddings collapse to a single point. Negative tuples in our contrastive framework push apart mismatched configurations, preventing collapse (see Eq.~\eqref{eq:alignment_loss}).}
\label{fig:optimization_pitfalls}
\vspace{-0.5em}
\end{figure*}

% Challenge 1 + cosine regularizer
\noindent\textbf{Cosine regularization.}
However, naive triangle-area minimization suffers from geometric ambiguity: the triangle area can shrink to zero even when one modality remains far from the other two---for example, when all three embeddings lie nearly on a single line, the triangle collapses to a flat shape with near-zero area despite poor mutual alignment (Figure~\ref{fig:optimization_pitfalls} left). To prevent such configurations, we augment the triangle area with a cosine regularizer between language and 3D flow embeddings, defining the joint alignment energy as
\begin{equation}
    E(z_L, z_I, z_F) = A(z_L, z_I, z_F) - \alpha \langle z_L, z_F \rangle,
    \label{eq:three_modal_energy}
\end{equation}
where $\alpha \ge 0$ balances triangle-area minimization and 
pairwise cosine alignment. The cosine term explicitly pulls $z_L$ and $z_F$ together, penalizing flat configurations where these modalities remain far apart even though the triangle area is small. Combined with the triangle area's joint constraint, the resulting energy encourages that low values reflect genuine alignment among all three modalities. Appendix~\ref{app:volume_only_pitfalls} provides a formal analysis of the issues underlying triangle-area minimization alone, and Appendix~\ref{app:cosine_regularization} shows how the cosine regularizer mitigates them.

% Challenge 2 + contrastive framework
\noindent\textbf{Contrastive framework.}
Yet directly minimizing $E$ admits another degeneracy: trivial collapse, where all three embeddings reduce to a single point and $E$ vanishes (Figure~\ref{fig:optimization_pitfalls} right). To prevent this, we embed the joint alignment energy into an InfoNCE-style contrastive objective~\cite{oord2018representation}. For each sample $i$ in a batch $\mathcal{B}$, we construct a set of negative tuples $\mathcal{N}(i)$ by mismatching one or more modality embeddings across the batch, and define the alignment loss as 
\begin{equation}
    \mathcal{L}_{\mathrm{align}} = -\sum_{i \in \mathcal{B}} \log
    \frac{\exp(-E(z_L^i, z_I^i, z_F^i)/\tau)}
    {\exp(-E(z_L^i, z_I^i, z_F^i)/\tau) + 
    \sum_{\tilde{\mathbf z} \in \mathcal{N}(i)} \exp(-E(\tilde{\mathbf z})/\tau)},
    \label{eq:alignment_loss}
\end{equation}
where $\tau > 0$ is the temperature parameter. By forcing matched tuples to achieve lower energy than mismatched ones, the contrastive loss prevents the collapse mode in which all samples share the same embedding and attain low energy simultaneously.

% Encoder architecture
\noindent\textbf{Encoder architecture.}
We instantiate the three encoders as follows. Given an image observation $I_t$, a future observation $I_{t+H}$ separated by temporal offset $H$, a language instruction $L$, and a 3D flow trajectory $F_{t:t+K}$ over a temporal window of length $K$, we encode the three modalities as
\begin{equation}
    z_I^{(t)} = \Pi\bigl(f_\phi(I_{t+H}) - f_\phi(I_t)\bigr), \quad
    z_L = \Pi\bigl(h_\theta(L)\bigr), \quad
    z_F^{(t)} = \Pi\bigl(g_\psi(F_{t:t+K};\, \mathrm{sg}(f_\phi(I_t)))\bigr),
    \label{eq:modality_embeddings}
\end{equation}
where $\Pi(v) = v/\|v\|_2$ projects features onto the unit sphere, and $f_\phi$, $h_\theta$, and $g_\psi$ denote the image, language, and 3D flow encoders, respectively. The image transition embedding $z_I^{(t)}$ is defined as the normalized feature difference between $I_t$ and $I_{t+H}$, forcing the embedding to capture visual state change rather than static appearance. The 3D flow embedding $z_F^{(t)}$ conditions on the current image feature with stop-gradient ($\mathrm{sg}$) to preserve semantic grounding while blocking trivial shortcut solutions through the image branch.
\subsection{Auxiliary Objectives for Dynamics-aware Representation}
\label{sec:full_pretraining_objective}

The alignment objective $\mathcal{L}_{\mathrm{align}}$ captures dynamics within each transition window, but it does not provide a signal about how representations should relate across longer temporal horizons. To encode trajectory-level temporal structure, we adopt a temporal contrastive loss~\cite{nair2022r3m, jiang2024robots}, which pulls embeddings of nearby frames closer than distant frames within the same trajectory. Given a triplet $(I_{t_1}, I_{t_2}, I_{t_3})$ from the same video with $t_1 < t_2 < t_3$, let $z_{t_1}^i, z_{t_2}^i, z_{t_3}^i$ denote their embeddings, and let $z_{t_1}^{\neq i}$ denote a negative embedding from a different video in the batch. We define
\begin{equation}
\mathcal{L}_{\mathrm{tcn}} = -\sum_{i \in \mathcal{B}} \log
\frac{\exp(\mathcal{S}(z_{t_1}^i, z_{t_2}^i))}
{\exp(\mathcal{S}(z_{t_1}^i, z_{t_2}^i)) + 
\exp(\mathcal{S}(z_{t_1}^i, z_{t_3}^i)) + 
\exp(\mathcal{S}(z_{t_1}^i, z_{t_1}^{\neq i}))},
\end{equation}
where $\mathcal{S}(\cdot, \cdot)$ is the negative $\ell_2$ distance, so that closer embeddings receive higher similarity scores.

To further reinforce the dynamics-aware representations, we introduce an auxiliary actor loss via a single-step 3D flow prediction objective in the spirit of behavior cloning~\cite{qi2024control}. This objective requires the image encoder to predict motion explicitly from a single frame, thereby encouraging the representation to encode manipulation dynamics more directly. Given the image feature $f_\phi(I_t)$, a 3D flow prediction head outputs $\hat{F}_t$, and we minimize the mean squared error to the ground-truth flow:
\begin{equation}
    \mathcal{L}_{\mathrm{act}} = \sum_{i \in \mathcal{B}} \| \hat{F}_t^{(i)} - F_t^{(i)} \|_2^2.
\end{equation}
Combining the three objectives yields the full pre-training objective
\begin{equation}
    \mathcal{L}_{\text{\method}} = \mathcal{L}_{\mathrm{align}} + 
    \lambda_{\mathrm{tcn}} \mathcal{L}_{\mathrm{tcn}} + 
    \lambda_{\mathrm{act}} \mathcal{L}_{\mathrm{act}},
\end{equation}
where $\lambda_{\mathrm{tcn}}$ and $\lambda_{\mathrm{act}}$ control the relative importance of the two auxiliary objectives.

\subsection{Dataset Construction}
\label{sec:dataset_construction}

Our pre-training framework relies only on RGB videos. Although the training objective uses image--language--3D flow triplets, all three signals can be derived from video alone: image transitions are obtained by sampling frames, 3D flow trajectories are estimated through point tracking and depth estimation while compensating for camera motion, and language instructions are generated by a vision-language model. This video-only requirement enables pre-training to scale across both human and robot videos. Building on the unified data generation pipeline of~\cite{lee2025tracegen} with several modifications tailored to our setting, we construct a large-scale dataset comprising 260K trajectories, each paired with image--language--3D flow triplets. The dataset is built from heterogeneous human and robot video sources~\cite{bu2025agibot, goyal2017something, grauman2022ego4d, khazatsky2024droid, o2024open, walke2023bridgedata, brohan2022rt, kalashnikov2018scalable}, providing broad diversity in objects, environments, and interaction patterns. Additional details on data sources, statistics, and generation procedures are provided in Appendix~\ref{app:dataset_construction}.

\section{Experiments}
\label{sec:experiments}

In this section, we evaluate \method through extensive experiments in both simulation and the real world.
Through these experiments, we aim to answer the following questions:

\begin{enumerate}[label=\textbf{Q\arabic*:}, leftmargin=3.2em, labelsep=0.6em, align=left, nosep]
    \item Does \method learn dynamics-aware representations that preserve control-relevant information for manipulation?
    \item Do dynamics-aware representations improve downstream policy learning compared to strong baselines?
    \item Can \method improve real-world manipulation under both in-distribution and out-of-distribution settings?
    \item Which design choices in \method are most critical to its performance? 
\end{enumerate}
\subsection{Benchmarks and Baselines}

\begin{figure*}[t!]
\centering
\includegraphics[width=1.0\textwidth]{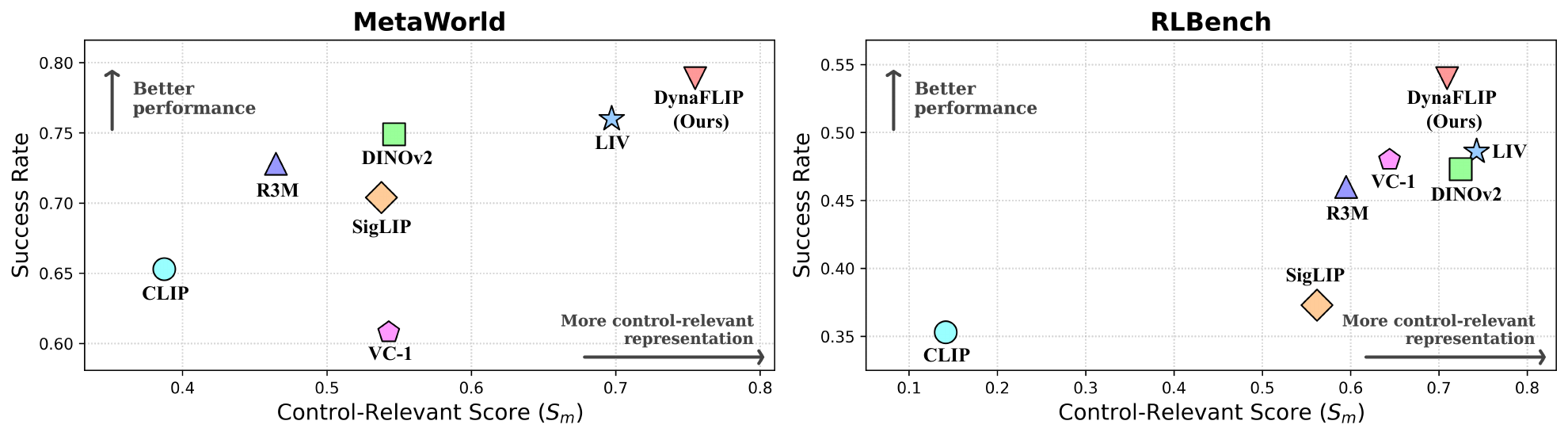}
\vspace{-1em}
\caption{\textbf{Control-relevant score versus downstream success rate \textnormal{(MLP policy)}.} The control-relevant score $S_m$~\cite{dong2026capturing} (x-axis) measures how well a frozen image encoder preserves state information relevant to control, and the y-axis reports policy success rate on MetaWorld~\cite{yu2020meta} (left) and RLBench~\cite{james2020rlbench} (right). 
\method appears in the top-right region of both plots, indicating its dynamics-aware representations preserve control-relevant information and improve manipulation performance.}
\label{fig:metaworld_rlbench_state_results}
\end{figure*}

\begin{figure*}[t!]
\centering
\begin{subfigure}[t]{0.49\textwidth}
\centering
\includegraphics[width=\textwidth]{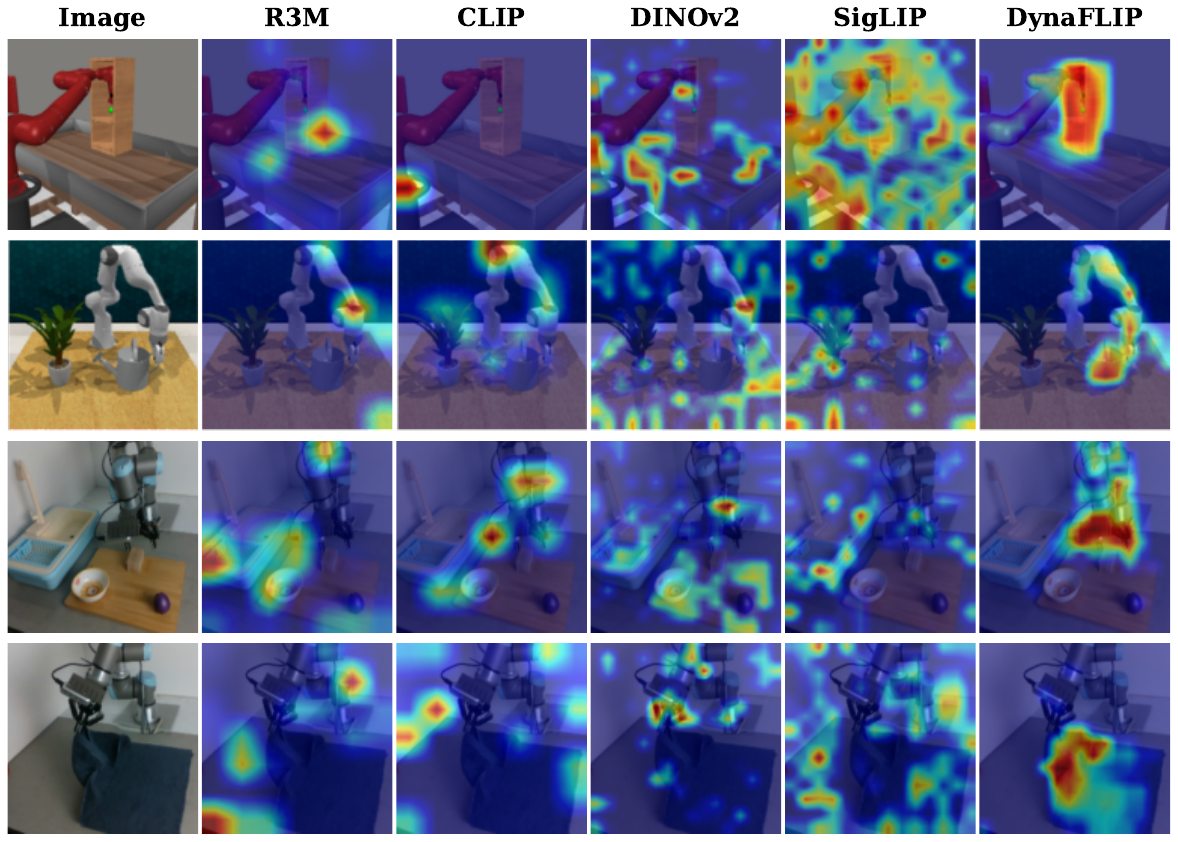}
\vspace{-1.5em}
\caption{Grad-CAM heatmaps over action prediction.}
\end{subfigure}
\hfill
\begin{subfigure}[t]{0.49\textwidth}
\centering
\includegraphics[width=\textwidth]{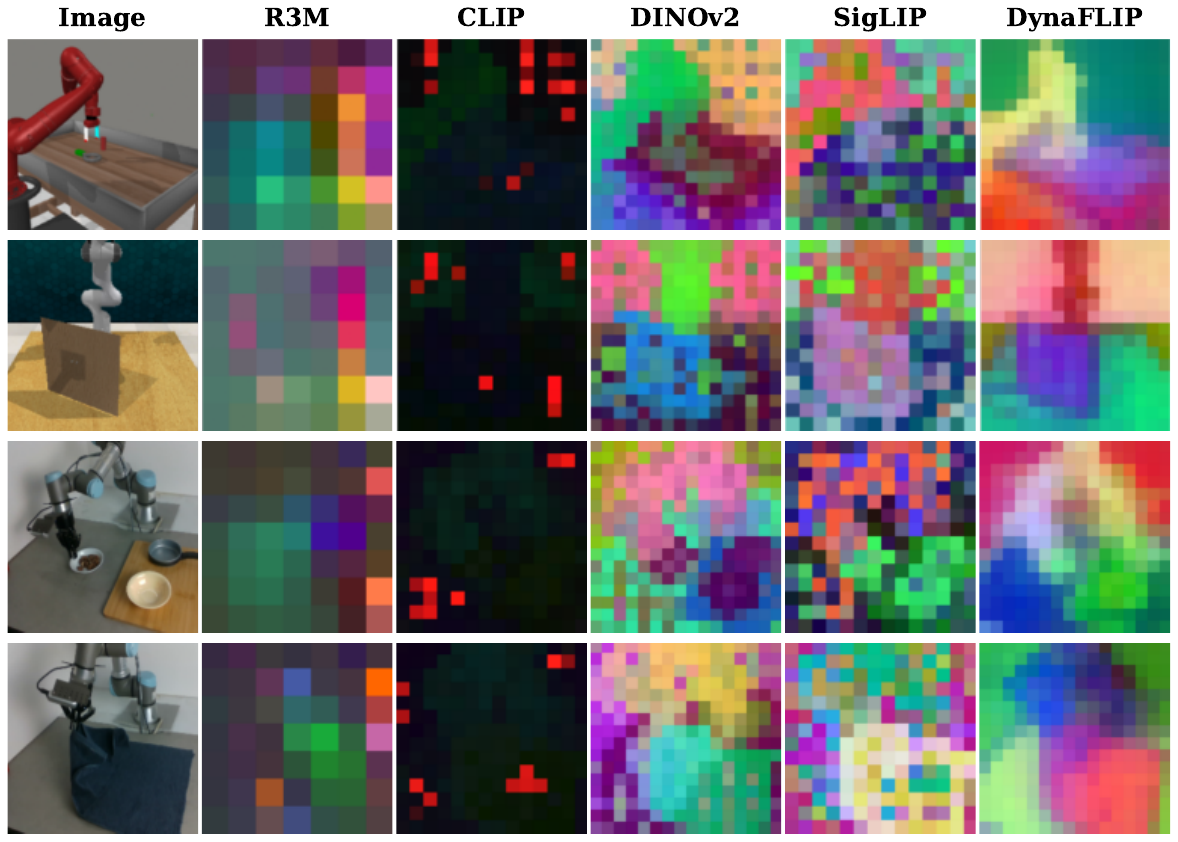}
\vspace{-1.5em}
\caption{Feature visualization with PCA.}
\end{subfigure}
\vspace{-0.5em}
\caption{\textbf{Grad-CAM and PCA visualizations \textnormal{(MLP policy)}.}
\textbf{(a)} Grad-CAM heatmaps show that \method attends to manipulated objects and interaction regions, whereas baselines often focus on task-irrelevant areas. 
\textbf{(b)} PCA visualizations show that \method yields more spatially coherent, object-level feature structures than the baselines. 
Additional visualizations are provided in Appendix~\ref{app:additional_gradcam} and Appendix~\ref{app:additional_pca}.}
\label{fig:gradcam_pca_visualizations}
\vspace{-1.0em}
\end{figure*}

\noindent\textbf{Benchmarks.}
We evaluate \method on three simulation benchmarks and three real-world manipulation tasks. 
\textbf{MetaWorld}~\cite{yu2020meta} uses a Sawyer arm with a two-finger gripper. We evaluate 15 tasks spanning varying difficulty levels~\cite{seo2023masked} with 25 demonstrations per task. 
\textbf{RLBench}~\cite{james2020rlbench} employs a Franka Panda arm. We evaluate 6 tasks from front-view observations with 100 demonstrations per task collected via the Open Motion Planning Library~\cite{sucan2012open}.
\textbf{LIBERO}~\cite{liu2023libero} is a multi-task, language-conditioned manipulation benchmark. We evaluate on LIBERO-90, LIBERO-Goal, LIBERO-Object, LIBERO-Spatial, and LIBERO-Long, where LIBERO-90 contains 90 tasks and each remaining suite contains 10 tasks with 50 demonstrations per task. 
\textbf{Real-World Manipulation} experiments use a UR3 robot arm equipped with a two-finger gripper. We consider two multi-instruction tasks, \textit{Pick <object> into Sink} and \textit{Pour almonds into <object>}, together with an \textit{Unfold Towel} task.

\noindent\textbf{Baselines.}
We compare \method with strong pre-trained representation baselines from three categories: robotic visual representations, self-supervised visual encoders, and vision-language pre-training models.
Among robotic visual representations, 
\textbf{R3M}~\cite{nair2022r3m} trains a ResNet~\cite{he2016deep} on human videos via time-contrastive learning and video-language alignment.
\textbf{VC-1}~\cite{majumdar2023we} pre-trains a ViT~\cite{dosovitskiy2020image} with Masked Auto-Encoding~\cite{he2022masked} on navigation and ImageNet~\cite{deng2009imagenet} data.
\textbf{LIV}~\cite{ma2023liv} trains a ResNet on human videos by aligning goal images with language and modeling rewards relative to goal states.
As a self-supervised visual encoder,
\textbf{DINOv2}~\cite{oquab2024dinov2} combines self-distillation with masked image modeling on large-scale curated image data.
Among vision-language models,
\textbf{CLIP}~\cite{radford2021learning} and \textbf{SigLIP}~\cite{zhai2023sigmoid} learn image-text alignment on large-scale paired data, with SigLIP replacing CLIP's multinomial cross-entropy objective with a pairwise sigmoid loss.

\subsection{Q1: Does \method{} learn dynamics-aware and control-relevant representations?}
\label{sec:analysis}

\noindent\textbf{Experiment setup.}
We first verify our central claim that \method's pre-training yields dynamics-aware representations that preserve control-relevant information. We analyze pre-trained image encoders on MetaWorld and RLBench: each encoder remains frozen, and only a \textbf{lightweight three-layer MLP policy} is trained on top, ensuring that downstream performance reflects representation quality rather than policy capacity. Appendix~\ref{app:experimental_details_metaworld_rlbench} describes the training and evaluation protocols for MetaWorld and RLBench.

\noindent\textbf{Quantitative analysis.}
We measure how well each encoder preserves control-relevant information using the control-relevant score ($S_m$) proposed in~\cite{dong2026capturing}, which quantifies how well a visual representation captures information needed for control. This score is computed by training a lightweight probe on top of the frozen image encoder to predict robot joint angles, end-effector pose, and the 6D pose and shape of task-relevant objects; Appendix~\ref{app:control_relevant_metric} provides the formal definition and evaluation protocol. 
Figure~\ref{fig:metaworld_rlbench_state_results} plots the control-relevant score ($S_m$) against downstream success rate on MetaWorld and RLBench. 
\method lies in the top-right region of both plots, achieving \textbf{the highest downstream success rate with high control-relevant scores}. This result indicates that \method preserves control-relevant information more faithfully, leading to higher downstream success rates.

\noindent\textbf{Qualitative analysis.}
We further inspect the learned representations through two visualizations. \textbf{(1) Grad-CAM}~\cite{selvaraju2017grad}, applied to the trained MLP policy with negative action-prediction error as the target, highlights the visual regions most influential for action prediction. \textbf{(2) PCA} on patch features examines the overall structure of the learned feature space. 
Figure~\ref{fig:gradcam_pca_visualizations} shows that \method concentrates attention on task-relevant objects and interaction regions, whereas baselines distribute attention over less relevant areas such as the background or irrelevant objects. PCA visualizations further show that \method produces more spatially coherent and object-aware feature structures than the baselines.

\noindent\textbf{Multimodal alignment.}
Finally, we measure multimodal alignment directly, without any policy learning in the loop, to confirm that the three modalities are jointly aligned as intended. On a held-out validation split of the pre-training data, we group episodes by instruction into three task categories (pick, open, pour) and sample 150 episodes per category. For each triplet, we compute the triangle area $A$ (Eq.~\eqref{eq:triangle_area}) spanned by the three embeddings---the exact quantity our objective minimizes---and report $1-A$, where $1$ indicates that the three embeddings coincide. Table~\ref{tab:alignment} compares matched triplets (diagonal) against triplets in which exactly one modality is swapped with that of a different task category (off-diagonal). Matched triplets score $0.64$--$0.69$, while every mismatched triplet falls well below, indicating that the three modalities are well aligned for matched transitions but not for mismatched ones.

Together, the quantitative, qualitative, and alignment results show that \method learns dynamics-aware representations that preserve control-relevant information and focus on regions critical for manipulation.

\begin{table*}[t]
\centering
\vspace{-0.3em}
\caption{\textbf{Multimodal alignment on held-out episodes.} We report $1-A$ (higher indicates stronger alignment), where $A$ is the triangle area in Eq.~\eqref{eq:triangle_area}, averaged over 150 episodes per task category. In each subtable, one modality (columns) is swapped across task categories while the other two (rows) remain matched; the diagonal corresponds to fully matched triplets.}
\label{tab:alignment}
\vspace{-0.5em}
\footnotesize
\setlength{\tabcolsep}{4pt}
\renewcommand{\arraystretch}{1.08}
\subfloat[
\textbf{Image swapped.}
\label{subtab:alignment_img}
]{
\begin{minipage}{0.30\linewidth}
\centering
\begin{tabular}{l ccc}
\toprule
$(1-A)\uparrow$ & \multicolumn{3}{c}{\textbf{Image}} \\
\cmidrule(lr){2-4}
\textbf{(Lang, Flow)} & Pick & Open & Pour \\
\midrule
Pick & \textbf{0.67} & 0.36 & 0.35 \\
Open & 0.34 & \textbf{0.64} & 0.34 \\
Pour & 0.34 & 0.34 & \textbf{0.69} \\
\bottomrule
\end{tabular}
\end{minipage}
}
\hfill
\subfloat[
\textbf{Language swapped.}
\label{subtab:alignment_lang}
]{
\begin{minipage}{0.30\linewidth}
\centering
\begin{tabular}{l ccc}
\toprule
$(1-A)\uparrow$ & \multicolumn{3}{c}{\textbf{Language}} \\
\cmidrule(lr){2-4}
\textbf{(Img, Flow)} & Pick & Open & Pour \\
\midrule
Pick & \textbf{0.67} & 0.14 & 0.12 \\
Open & 0.12 & \textbf{0.64} & 0.12 \\
Pour & 0.11 & 0.12 & \textbf{0.69} \\
\bottomrule
\end{tabular}
\end{minipage}
}
\hfill
\subfloat[
\textbf{3D flow swapped.}
\label{subtab:alignment_flow}
]{
\begin{minipage}{0.30\linewidth}
\centering
\begin{tabular}{l ccc}
\toprule
$(1-A)\uparrow$ & \multicolumn{3}{c}{\textbf{3D flow}} \\
\cmidrule(lr){2-4}
\textbf{(Lang, Img)} & Pick & Open & Pour \\
\midrule
Pick & \textbf{0.67} & 0.47 & 0.48 \\
Open & 0.44 & \textbf{0.64} & 0.46 \\
Pour & 0.47 & 0.46 & \textbf{0.69} \\
\bottomrule
\end{tabular}
\end{minipage}
}
\end{table*}
\subsection{Q2: Do \method{}'s representations improve downstream policy learning?}
\label{sec:libero}

\begin{table*}[t] 
    \centering
    \caption{
    \textbf{LIBERO benchmark results \textnormal{(Diffusion policy)}.} We evaluate various pre-trained encoders under two settings: \emph{Frozen} keeps both image and language encoders frozen, while \emph{LoRA Fine-tuned} adapts both encoders jointly with the diffusion policy as an additional comparison. 
    The evaluation metric is success rate (\%). \textbf{Bold} and \underline{underline} numbers indicate the best and second-best results in each column, respectively.}
    \label{tab:libero_results}
    \setlength{\tabcolsep}{4pt} 
    \small 
    \resizebox{\textwidth}{!}{
    \begin{tabular}{ll ccccc>{\columncolor{meangray}}c ccccc>{\columncolor{meangray}}c}
        \toprule
        \multirow{2}{*}{\begin{tabular}[c]{@{}l@{}}\textbf{Image}\\ \textbf{Encoder}\end{tabular}} & \multirow{2}{*}{\begin{tabular}[c]{@{}l@{}}\textbf{Language}\\ \textbf{Encoder}\end{tabular}} & \multicolumn{6}{c}{\textbf{Frozen}} & \multicolumn{6}{c}{\textbf{LoRA Fine-tuned}} \\
        \cmidrule(lr){3-8} \cmidrule(lr){9-14}
         & & \textbf{90} & \textbf{Goal} & \textbf{Object} & \textbf{Spatial} & \textbf{Long} & \textbf{Mean} & \textbf{90} & \textbf{Goal} & \textbf{Object} & \textbf{Spatial} & \textbf{Long} & \textbf{Mean} \\
        \midrule
        R3M~\cite{nair2022r3m} & CLIP~\cite{radford2021learning}   & 24.4 & 45.0 & 0.5  & \textbf{53.0} & 13.5 & 27.3 & 38.5 & 67.0 & 2.5    & 56.5    & 37.5    & 40.4 \\
        VC-1~\cite{majumdar2023we} & CLIP~\cite{radford2021learning}   & 12.8    & 52.5    & 11.5    & \underline{52.0} & 12.5    & 28.3 & 72.4    & \underline{83.0}    & \textbf{83.5}    & 71.0    & 62.0    & 74.4 \\
        LIV~\cite{ma2023liv} & LIV~\cite{ma2023liv}  & 22.3    & 64.0    & 6.5    & 51.0  & 9.0    & 30.6 & 72.7    & 78.5    & 49.0    & 75.5    & 62.0    & 67.5 \\
        CLIP~\cite{radford2021learning} & CLIP~\cite{radford2021learning} & 13.8 & 38.5 &  1.5 & 50.0 &  9.5 & 22.7 & 78.1 & 79.5 & 79.0    & 75.5    & 68.5    & 76.1 \\
        DINOv2~\cite{oquab2024dinov2} & CLIP~\cite{radford2021learning}  & 14.4 & \textbf{75.0} & \underline{33.5} & 42.5 & \textbf{20.5} & \underline{37.2} & \textbf{83.6}    & 77.5 & \underline{82.0}    & \textbf{81.0}    & 67.5    & 78.3 \\
        SigLIP~\cite{zhai2023sigmoid} & SigLIP~\cite{zhai2023sigmoid} & 24.3 & 54.5 & 13.0 & \underline{52.0} &  8.5 & 30.5 & \underline{82.6} & 80.5 & \underline{82.0}   & 74.0    & \underline{76.5}    & \underline{79.1} \\
        \method (Ours) & \method (Ours)   & \textbf{31.7} & \underline{70.5} & \textbf{37.5} & 51.5 & \underline{16.5} & \textbf{41.5} & 78.1    & \textbf{84.5} & \textbf{83.5} & \underline{78.5} & \textbf{80.5} & \textbf{81.0} \\
        \bottomrule
    \end{tabular}
    }
% \vspace{-0.5em}
\end{table*}

\noindent\textbf{Experiment setup.}
We next ask whether dynamics-aware representations improve downstream policy learning. We evaluate on the LIBERO benchmark (LIBERO-90, Goal, Object, Spatial, and Long) using \textbf{Diffusion Policy}~\cite{chi2025diffusion} as the imitation-learning backbone. Each setup pairs a pre-trained image encoder with a language encoder; for baselines without their own text encoder, we substitute CLIP's text encoder. Our primary setting is \emph{frozen}: both encoders remain fixed, so downstream performance directly reflects the quality and reusability of the pre-trained representations. We additionally report a \emph{fine-tuned} setting, in which LoRA~\cite{hulora} adapters on both encoders are trained jointly with the diffusion policy. Appendix~\ref{app:experimental_details_libero} provides detailed training settings and evaluation protocols.

\noindent\textbf{Results.}
Table~\ref{tab:libero_results} reports the LIBERO results. \method achieves the highest mean success rate in both the frozen and fine-tuned settings, outperforming all baselines. \textbf{(1)} \textbf{The frozen-setting} results show that \method's pre-trained features can be reused effectively without encoder adaptation. \textbf{(2)} \textbf{The fine-tuned setting} further confirms that this advantage persists after task-specific adaptation. We attribute this consistent advantage to differences in pre-training paradigms. Most baselines are trained primarily on \emph{static} visual data and therefore receive limited signal about how scenes evolve under interaction. In contrast, \method explicitly aligns three transition-centric modalities---image transitions, language, and 3D flow trajectories---encouraging the encoder to focus on control-relevant regions rather than background appearance.

\subsection{Q3: Does \method{} improve real-world manipulation under distribution shift?}
\label{sec:real_world}

\begin{figure*}[t!]
\centering
\includegraphics[width=1.0\textwidth]{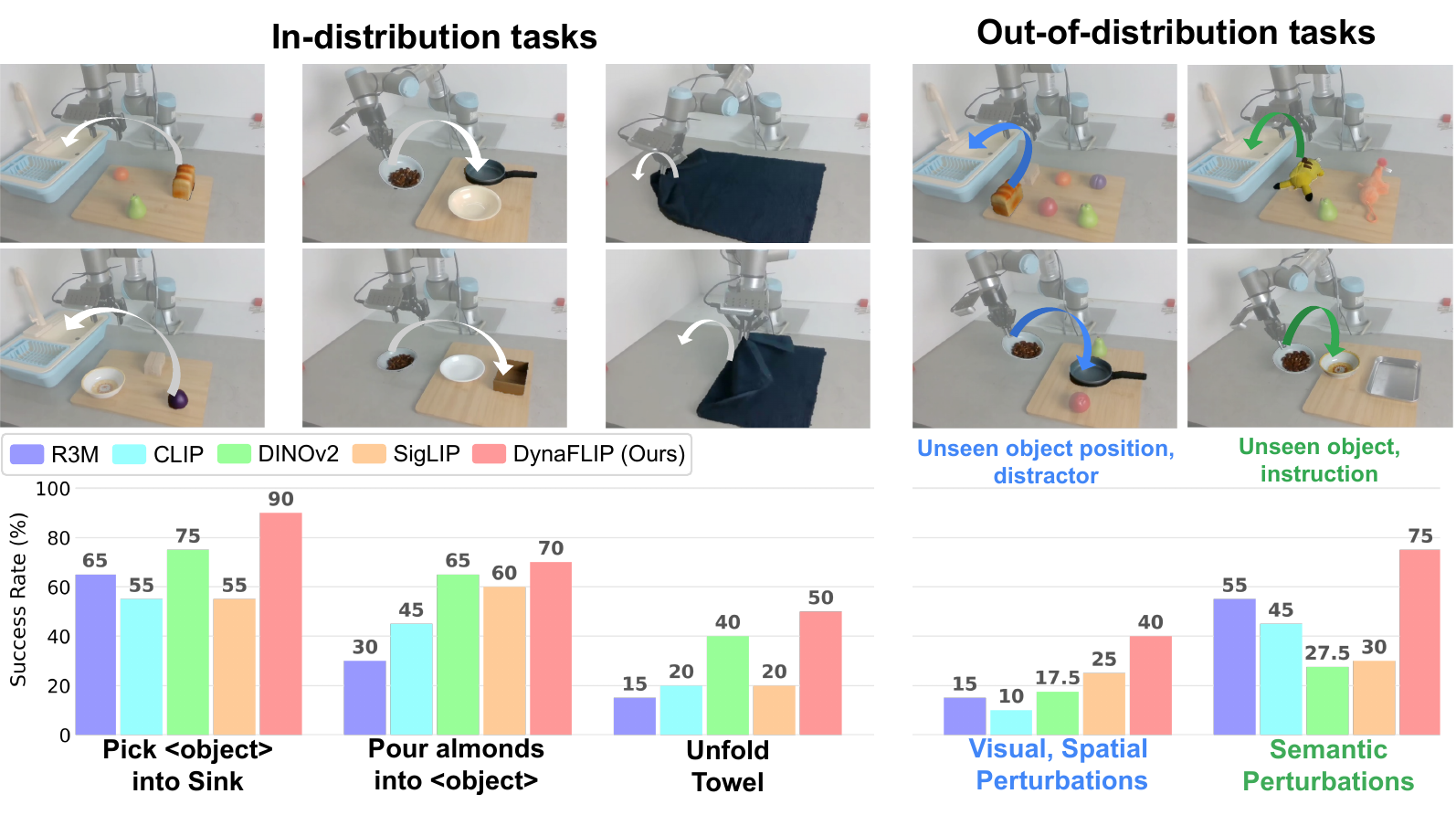}
\vspace{-1em}
\caption{\textbf{Real-world manipulation results \textnormal{(VLA policy)}.} \method performs well not only on the three in-distribution tasks, but also under both out-of-distribution perturbation types. The top row contrasts in-distribution (seen) tasks with out-of-distribution (unseen) evaluation settings, and the bottom row reports success rates (\%) on the three in-distribution tasks together with two out-of-distribution settings.}
\label{fig:real_world_results}
\vspace{-0.5em}
\end{figure*}

\noindent\textbf{Experiment setup.}
We evaluate \method in real-world manipulation by integrating a frozen pre-trained image encoder into \(\pi_{0.5}\)~\cite{intelligence2025pi_}, a vision-language-action (VLA) model. We adopt a lightweight visual-injection design similar to plug-in visual injection (PVI)~\cite{zhang2026pvi}: an additional visual branch encodes features from the pre-trained image encoder, and an injection module projects them into the hidden feature space of the diffusion transformer of $\pi_{0.5}$. The additional visual branch remains frozen and only the lightweight injection module is trained, testing whether \method can be reused inside a VLA without end-to-end visual fine-tuning. We evaluate on a UR3 robot arm with a two-finger gripper across three in-distribution tasks: \textit{Pick <object> into Sink}, \textit{Pour almonds into <object>}, and \textit{Unfold Towel}. For out-of-distribution (OOD) evaluation, we introduce two types of perturbations: \emph{visual and spatial perturbations} (unseen object positions and distractors) and \emph{semantic perturbations} (unseen objects and instructions). Appendix~\ref{app:experimental_details_real} provides additional details on the hardware setup, data collection, model architecture, and training and evaluation protocol.

\noindent\textbf{Results.} \textbf{(1) In-Distribution.}
Figure~\ref{fig:real_world_results} shows that \method achieves the highest success rates across all three in-distribution tasks. Together with the frozen results on MetaWorld, RLBench, and LIBERO, this demonstrates that \method transfers robustly across diverse downstream policies---MLP, diffusion policy, and VLA---without task-specific visual fine-tuning. \textbf{(2) Out-of-Distribution.} The advantage of \method becomes even more pronounced under OOD settings. Under visual and spatial perturbations, CLIP and SigLIP often fail at precise grasping, while \method's focus on control-relevant regions enables it to remain robust to changes in object layouts and the presence of distractors. Under semantic perturbations, DINOv2 frequently interacts with objects irrelevant to the instruction, reflecting its lack of direct language grounding. By contrast, \method incorporates language as one of its pre-training modalities and learns to align visual changes with task-relevant instructions, yielding representations that remain robust under unseen objects and instructions.
\subsection{Q4: Which design choices of \method{} matter most?}

Table~\ref{tab:design_ablation} presents ablations of our system with respect to four aspects: multimodal input, alignment design, optimization-pitfall mitigation, and auxiliary objectives.

\noindent\textbf{All three modalities are necessary for dynamics-aware representation learning.} Removing either 3D flow or language causes a clear drop in performance. 3D flow provides explicit motion cues, while language supplies task-level semantics. Both contribute complementary signals beyond image transition alone.

\noindent\textbf{Alignment design matters more than simply adding modalities.} We replace the simplex-guided alignment with an anchor-based pairwise loss that uses language as the anchor and applies two separate InfoNCE objectives, language--image transition and language--3D flow, so that each non-anchor modality is aligned to language but not to each other. This variant leads to a clear drop in performance. \method's gains thus stem not merely from using multiple modalities, but from how those modalities are aligned through higher-order multimodal geometry rather than pairwise similarity.

\noindent\textbf{Mitigating optimization pitfalls is crucial for stable learning.} Removing the contrastive framework---i.e., directly minimizing the joint alignment energy (Eq.~\eqref{eq:three_modal_energy}) without negative tuples---causes the most severe drop, confirming that the contrastive framework is necessary to prevent trivial collapse. Removing the cosine regularizer also reduces performance, supporting its role in mitigating geometric ambiguity.
Geometric degeneracy is a theoretical possibility rather than a guaranteed failure mode; even when it does not occur, the cosine regularizer still improves performance by stabilizing positive alignment gradients. A detailed analysis is provided in Appendix~\ref{app:cosine_regularization}.

\noindent\textbf{Auxiliary objectives provide additional gains.} Removing either auxiliary loss degrades performance. The larger drop from removing $\mathcal{L}_{\text{tcn}}$ confirms its complementary role: it captures trajectory-level temporal structure beyond the transition window covered by $\mathcal{L}_{\text{align}}$.           

\renewcommand{\thesubtable}{{\alph{subtable}}}
\begin{table*}[t]
\centering
\vspace{-0.3em}
\caption{\textbf{Ablation studies \textnormal{(Diffusion policy)}.} We report mean success rate (\%) averaged over LIBERO-Goal, Object, Spatial, and Long, with both image and language encoders frozen.}
\label{tab:design_ablation}
\vspace{-0.5em}
\centering
\footnotesize
\setlength{\tabcolsep}{2pt}
\renewcommand{\arraystretch}{1.08}
\subfloat[
\textbf{Multimodal input.}
\label{subtab:ablation_multimodal}
]{
\begin{minipage}{0.22\linewidth}
\centering
\begin{tabular}{p{0.62\linewidth} c}
\toprule
\textbf{Variant} & \textbf{Mean} \\
\midrule
w/o. 3D flow & 37.1 \\
w/o. Language & 35.4 \\
\midrule
\method (full) & \textbf{44.0} \\
\bottomrule
\end{tabular}
\end{minipage}
}
\hspace{0.1em}
\subfloat[
\textbf{Alignment design.} \label{subtab:ablation_alignment}
]{
\begin{minipage}{0.22\linewidth}
\centering
\begin{tabular}{p{0.62\linewidth} c}
\toprule
\textbf{Variant} & \textbf{Mean} \\
\midrule
\noalign{\vskip 0.1em}
\makecell[l]{Anchor-based\\alignment} & 31.8 \\
\noalign{\vskip 0.1em}
\midrule
\method (full) & \textbf{44.0} \\
\bottomrule
\end{tabular}
\end{minipage}
}
\hspace{0.1em}
\subfloat[
\textbf{Optimization pitfalls.} \label{subtab:ablation_optimization}
]{
\begin{minipage}{0.27\linewidth}
\centering
\begin{tabular}{p{0.62\linewidth} c}
\toprule
\textbf{Variant} & \textbf{Mean} \\
\midrule
w/o. Negative tuples & 18.1 \\
w/o. Cosine reg. & 39.8 \\
\midrule
\method (full) & \textbf{44.0} \\
\bottomrule
\end{tabular}
\end{minipage}
}
\hspace{0.1em}
\subfloat[
\textbf{Auxiliary objectives.}\label{subtab:ablation_auxiliary}
]{
\begin{minipage}{0.22\linewidth}
\centering
\begin{tabular}{p{0.62\linewidth} c}
\toprule
\textbf{Variant} & \textbf{Mean} \\
\midrule
w/o. $\mathcal{L}_{\text{act}}$ & 43.4 \\
w/o. $\mathcal{L}_{\text{tcn}}$ & 39.6 \\
\midrule
\method (full) & \textbf{44.0} \\
\bottomrule
\end{tabular}
\end{minipage}
}
\vspace{-0.5em}
\end{table*}

\section{Related work}
\noindent\textbf{Visual Representations for Robotic Manipulation.}
Visual foundation models have driven progress in robot policy learning, mainly through two paradigms: self-supervised visual pre-training~\cite{grill2020bootstrap, caron2020unsupervised, caron2021emerging, oquab2024dinov2} and contrastive vision-language pre-training~\cite{joulin2016learning, radford2021learning, cherti2023reproducible, zhai2023sigmoid}. Self-supervised models such as DINOv2~\cite{oquab2024dinov2} learn spatially precise features that capture both global context and local detail, but lack a direct interface to language, limiting their use in open-vocabulary settings and instruction-following robots. Contrastive vision-language models such as CLIP~\cite{radford2021learning} and SigLIP~\cite{zhai2023sigmoid} learn semantically aligned representations from large-scale paired data, supporting strong zero-shot generalization, but lack the fine-grained spatial reasoning needed for manipulation~\cite{jose2025dinov2}. 

Both paradigms, however, learn primarily from static data and therefore lack \emph{dynamics awareness}. This limitation matters for manipulation, where success depends on how scenes change under interaction, not only on object and instruction recognition. \method addresses this gap by aligning three transition-centric modalities---image transitions, language, and 3D flow. These signals allow the encoder to focus on control-relevant regions rather than visually salient but task-irrelevant areas.

\noindent\textbf{Pre-training Objectives for Robotic Representations.}
A separate line of work develops pre-training objectives specifically tailored for robotic representations. Early efforts applied \emph{single-modality} self-supervised objectives over static images: MVP~\cite{xiao2022masked} and VC-1~\cite{majumdar2023we} utilize Masked Autoencoder (MAE) on large-scale human datasets to learn visual features, VIP~\cite{ma2022vip} learns implicit value functions to encode distance-to-goal representations, and HRP~\cite{srirama2024hrp} extracts human affordances from videos. Another line of work introduces \emph{multimodal} supervision: R3M~\cite{nair2022r3m} and LIV~\cite{ma2023liv} align images with language, while MCR~\cite{jiang2024robots} aligns images with robot actions and proprioceptive trajectories; these methods use only two modalities, and MCR additionally requires robot-specific signals that prevent direct use of human videos. LaDA~\cite{weng2026language} extends to three modalities by aligning concatenated image-language features with action embeddings via contrastive learning for VLA training, but this design treats language as an auxiliary input to the image branch rather than aligning all three modalities jointly. In contrast, \method aligns image transitions, language, and 3D flow \emph{jointly} through a simplex-based formulation, enabling mutual alignment among all three modalities while relying only on RGB video.

\section{Conclusion}
\label{sec:conclusion}

We present \method, a dynamics-aware 3D flow-language-image pre-training framework that pushes motion understanding upstream into perception. By jointly aligning image transitions, language, and 3D flow through a simplex-based formulation---augmented with a cosine regularizer and a contrastive framework to resolve optimization pitfalls---\method learns visual representations that focus on control-relevant regions. Across simulated and real-world manipulation, \method transfers strongly as a reusable visual backbone and consistently outperforms baselines, with especially large gains under visual, spatial, and semantic distribution shifts. Our results indicate that robot generalization improves when visual representations are trained to encode not just what is present, but how the world changes under action.

\noindent\textbf{Limitations and future work.}
First, \method is pre-trained on 260K trajectories, which is smaller than the data scales used by several large-scale visual and vision-language baselines~\cite{oquab2024dinov2, zhai2023sigmoid, nair2022r3m}. Scaling \method to larger human and robot video corpora is a promising direction for future work.
Second, our 3D flow is extracted from a uniform $20\times20$ grid of keypoints, which captures all motion in the scene after compensating for camera motion---including task-irrelevant motion. As a result, pre-training videos containing task-irrelevant motion may inject noisy supervision into the representation; future work could explore keypoint sampling focused on the agent and task-relevant objects to mitigate this issue.

\newpage
\begin{ack}
This work was supported by Samsung Research Funding \& Incubation Center of Samsung Electronics under Project Number SRFC-IT2402-17. Lee and Huang are supported by DARPA HR001124S0029-AIQ-FP-019, National Science Foundation TRAILS Institute (2229885). Private support was provided by Open Philanthropy and Apple.
\end{ack}

\bibliographystyle{plainnat}
\bibliography{ref}

%%%%%%%%%%%%%%%%%%%%%%%%%%%%%%%%%%%%%%%%%%%%%%%%%%%%%%%%%%%%
\newpage
\appendix
\onecolumn

% ToC of Appendix
\let\addcontentsline\oldaddcontentsline
\renewcommand{\contentsname}{\Large{Appendix}}
\setcounter{tocdepth}{4}
\tableofcontents
\clearpage
\section{Mathematical Proofs and Theoretical Details}
\label{app:mathematical_proofs}

This section provides the theoretical details of the proposed simplex-guided contrastive objective in \method. The main paper identifies two optimization pitfalls of naive simplex-volume minimization: \emph{geometric ambiguity} and \emph{trivial collapse}, addressed through a cosine regularizer and a contrastive framework, respectively. In this section, we focus on the analysis underlying the cosine regularizer: we show that, beyond geometric ambiguity, naive volume minimization suffers from an additional issue---\emph{conflicting alignment gradients}---and that the cosine regularizer mitigates both. We organize the analysis into four parts.

Section~\ref{app:simplex_volume_definition} introduces the generalized simplex volume, which reduces to the triangle area in the three-modal setting used by \method. Section~\ref{app:contrastive_form} shows that our objective retains the standard energy-based contrastive learning structure and only modifies the geometry of the positive alignment energy.
Section~\ref{app:volume_only_pitfalls} then analyzes the volume-only objective and identifies the two optimization pitfalls: geometric ambiguity and conflicting alignment gradients. Finally, Section~\ref{app:cosine_regularization} explains how the cosine regularizer mitigates both pitfalls by introducing explicit pairwise attraction between selected modality embeddings.

\subsection{Generalized Simplex Volume}
\label{app:simplex_volume_definition}

For an $m$-modal tuple, let $z_1, \dots, z_m \in \mathbb{R}^d$ be 
$\ell_2$-normalized modality embeddings, and define
\[
U = [z_2 - z_1, \dots, z_m - z_1] \in \mathbb{R}^{d \times (m-1)}.
\]
Let $G = U^\top U \in \mathbb{R}^{(m-1) \times (m-1)}$ be the Gram 
matrix of the simplex edge vectors. The generalized simplex volume 
is defined as
\begin{equation}
\label{eq:simplex_volume}
\mathcal{V}_m(z_1, \dots, z_m) = \frac{1}{(m-1)!}\sqrt{\det(G)}.
\end{equation}
A smaller value of $\mathcal{V}_m$ indicates that the modality 
embeddings form a lower-volume configuration in the shared latent 
space, reflecting stronger joint alignment across modalities.
In the three-modal setting used by \method, this quantity reduces to the triangle area defined in Eq.~\eqref{eq:triangle_area}, and we focus on this case in the analyses that follow.

\subsection{Contrastive Learning with Simplex-Guided Energy}
\label{app:contrastive_form}
We recall the joint alignment energy function in Eq.~\eqref{eq:three_modal_energy}:
\begin{equation*}
    E(z_L, z_I, z_F) = A(z_L, z_I, z_F) - \alpha \langle z_L, z_F \rangle,
\end{equation*} 
where $A$ denotes the triangle area defined in Eq.~\eqref{eq:triangle_area} and $\alpha \ge 0$ balances triangle-area minimization and pairwise cosine alignment.

For each matched tuple $(z_L^i, z_I^i, z_F^i)$, let $\mathcal{N}(i)$ denote a set of mismatched negative tuples constructed by mismatching one or more modality embeddings across the batch. We define $E_i^+ = E(z_L^i, z_I^i, z_F^i)$ for the matched tuple and $E_{i\ell}^- = E(\tilde{z}_L^i, \tilde{z}_I^i, \tilde{z}_F^i)$ for each negative tuple $\ell$, and incorporate the energy into an 
InfoNCE-style contrastive objective~\cite{oord2018representation}:
\begin{equation}
    \mathcal{L}_i = -\log \frac{\exp(-E_i^+/\tau)}{\exp(-E_i^+/\tau) + 
    \sum_{\ell} \exp(-E_{i\ell}^-/\tau)}.
    \label{eq:app_infonce}
\end{equation}

Let $p_i^+$ and $p_{i\ell}^-$ denote the corresponding softmax probabilities.  
Differentiating the loss yields:
\begin{equation}
    \nabla \mathcal{L}_i = \underbrace{\frac{1 - p_i^+}{\tau} \nabla 
E_i^+}_{\text{Alignment Term}} - \underbrace{\sum_{\ell} 
    \frac{p_{i\ell}^-}{\tau} \nabla E_{i\ell}^-}_{\text{Uniformity 
    Term}}.
    \label{eq:app_grad_loss}
\end{equation}

The alignment term decreases the energy of the matched tuple, while the uniformity term increases the energy of mismatched tuples.
In this work, we focus on the alignment term, which directly captures the effect of the proposed energy on matched multimodal tuples.
The uniformity term follows the standard contrastive repulsion mechanism, and we refer interested readers to prior analyses~\cite{wang2020understanding, yin2026towards, betser2026infonce}.

Substituting $E$ into the alignment term gives:
\begin{equation}
    \nabla E_i^+ = \nabla A(z_L^i, z_I^i, z_F^i) - \alpha \nabla 
    \langle z_L^i, z_F^i \rangle.
\end{equation}
Therefore, the alignment gradient is a linear combination of a triangle-area term and a cosine-based pairwise attraction term, capturing higher-order geometry and directional consistency.
While this decomposition reveals the structure of the alignment gradient, it also suggests that the volume term alone may not provide a reliable alignment signal. 
We analyze this issue in the following section.

\subsection{Why Simplex-Volume Alone is Insufficient}
% \subsection{Optimization Pitfalls of Simplex-Volume-Only Alignment}
\label{app:volume_only_pitfalls}

We analyze the simplex-volume alignment in the three-modal case, where the objective reduces to the triangle area defined in Eq.~\eqref{eq:triangle_area}.
This setting allows us to characterize the alignment gradient and reveal two key limitations: ambiguity of low-volume configurations and 
conflicting alignment gradients.

\subsubsection{Ambiguity of Low-Volume Configurations}
\label{app:low_volume_ambiguity}

Low simplex volume does not necessarily imply pairwise alignment among all modalities. The simplex volume can vanish even when some modality pairs remain far apart---for example, when a subset of embeddings collapses together, or when all embeddings become nearly collinear.

For example, we illustrate this in the three-modal case. 
Consider unit vectors in $\mathbb R^2$:
\begin{equation}
    x=e_1,    \qquad    y=-e_1,    \qquad    z=\cos\theta\, e_1+\sin\theta\, e_2,
    \label{eq:app_flat_triangle_example}
\end{equation}
As $\theta\to0$, we have $z\to x$, so the pair $(x,z)$ collapses and the triangle area satisfies $A(x,y,z)=|\sin\theta|\to0$.
However, the pair $(x,y)$ remains maximally misaligned, with $\langle x, y\rangle=-1, \|x-y\|=2$.
This example shows that the simplex volume can be minimized by collapsing only a subset of modalities. 
Therefore, low volume does not guarantee pairwise alignment among all modalities.

\subsubsection{Conflicting Alignment Directions in Volume-Induced Gradients}
\label{app:conflicting_directions}

The volume-induced alignment does not define a single direction. 
Instead, it decomposes into multiple edge-wise pulls that can conflict with each other.
Let $x,y,z\in\mathbb S^{d-1}$ denote unit-normalized embeddings from three modalities, and define:
\[
    a=\langle x,y\rangle,\qquad
    b=\langle x,z\rangle,\qquad
    c=\langle y,z\rangle .
\]
The triangle area $A(x,y,z)$ can be expressed as a function of these pairwise inner products. 
For a non-degenerate triangle, the gradient with respect to modality $x$ decomposes as:
\begin{equation}
    \nabla_x A
    =
    \frac{\partial A}{\partial a}\nabla_x a
    +
    \frac{\partial A}{\partial b}\nabla_x b ,
    \label{eq:app_triangle_chain_rule}
\end{equation}
Since the embeddings lie on the unit sphere, the gradients are taken in the tangent space:
\begin{equation}
    \nabla_x a
    =
    \nabla_x\langle x,y\rangle
    =
    P_x^\perp y,
    \qquad
    \nabla_x b
    =
    \nabla_x\langle x,z\rangle
    =
    P_x^\perp z,
    \label{eq:app_tangent_gradients}
\end{equation}
where $P_x^\perp v = v-\langle x,v\rangle x$. Thus, the gradient becomes:
\begin{equation}
    \nabla_x A
    =
    \omega_{xy}P_x^\perp y
    +
    \omega_{xz}P_x^\perp z,
    \qquad
    \omega_{xy}=\frac{\partial A}{\partial a},
    \qquad
    \omega_{xz}=\frac{\partial A}{\partial b}.
    \label{eq:app_triangle_grad_x}
\end{equation}

Under gradient descent, the update direction is $-\nabla_x A$, which decomposes into two edge-wise positive pulls:
\begin{equation}
    u_{xy}
    =
    -\omega_{xy}P_x^\perp y,
    \qquad
    u_{xz}
    =
    -\omega_{xz}P_x^\perp z .
    \label{eq:app_edgewise_pulls}
\end{equation}
The total volume-induced alignment pull on modality $x$ then becomes:
\begin{equation}
    u_x^{\mathrm{vol}}
    =
    u_{xy}+u_{xz}.
    \label{eq:app_total_volume_pull}
\end{equation}
This shows that the volume-induced alignment is not governed by a single direction but by the sum of multiple edge-wise pulls.
When these pulls are aligned, they reinforce each other.
However, when they point in different directions, they partially cancel, leading to a weaker effective update.

\subsection{Mitigating Volume-Only Pitfalls with Cosine Regularization}
\label{app:cosine_regularization}

The cosine regularizer complements the simplex-volume term by adding explicit pairwise alignment constraints. This additional pairwise signal mitigates the two pitfalls identified in Section~\ref{app:volume_only_pitfalls}: geometrically ambiguous low-volume configurations and conflicting alignment gradients.

\noindent\textbf{Reducing low-volume ambiguity.}
The cosine regularizer also reduces low-volume ambiguity by introducing an explicit distance-based penalty between selected modality pairs.
For unit-normalized embeddings $x$ and $y$, we have:
\begin{equation}
    1-\langle x,y\rangle=\frac{1}{2}\|x-y\|^2.
    \label{eq:app_cosine_distance_identity}
\end{equation}
Thus, the cosine term directly penalizes large distances between modality embeddings. 
Under the combined objective
\begin{equation}
    A(x,y,z)
    +
    \alpha\bigl(1-\langle x,y\rangle\bigr)
    =
    A(x,y,z)
    +
    \frac{\alpha}{2}\|x-y\|^2,
    \label{eq:app_triangle_shifted_energy}
\end{equation}
the area term encourages a low-volume configuration, while the cosine term discourages distant modality pairs.
As a result, configurations with low volume but large pairwise distances become less favorable.

\noindent\textbf{Reducing conflict in volume-based gradients.}
We consider the three-modal setting and recall the cosine-regularized energy:
\begin{equation}
    E(x,y,z)=A(x,y,z)-\alpha\langle x, y \rangle,
\end{equation}
where $A(x,y,z)$ denotes the triangle-area term.
For the anchor modality $x$, the volume-only update direction is given by $-\nabla_x A=u_{xy}+u_{xz}$, which decomposes into two edge-wise pulls that may partially cancel.
With cosine regularization, the update becomes:
\begin{equation}
    -\nabla_x E=-\nabla_x A+\alpha P_x^\perp y=u_{xy}+u_{xz}+\alpha P_x^\perp y,
\end{equation}
where $P_x^\perp y=y-\langle x, y\rangle x$.
The additional term $\alpha P_x^\perp y$ introduces an explicit pairwise alignment direction.
To see this, consider a small step $\delta x = \eta P_x^\perp y$ with $\eta>0$.
Then 
\begin{equation}
    \frac{d}{d\eta}\langle x + \eta P_x^\perp y, y \rangle = \langle P_x^\perp y, y\rangle = 1-\langle x,y\rangle^2 \ge 0.
\end{equation}
Thus, $P_x^\perp y$ is an ascent direction for $\langle x, y \rangle$, meaning that the cosine term directly increases the similarity between the selected pair.
As a result, even when the volume-induced edge-wise pulls partially cancel, the cosine regularizer preserves a non-vanishing pairwise alignment signal for the selected modalities.

In summary, the cosine regularizer complements the simplex-volume term by resolving its inherent ambiguities.
While the volume term captures higher-order geometric structure, the cosine term introduces explicit pairwise constraints that prevent degenerate low-volume configurations and maintain a meaningful alignment signal.

\section{Dataset Construction}
\label{app:dataset_construction}

In this section, we provide additional details on the construction of our image--language--3D flow dataset. We first describe the dataset composition across heterogeneous human and robot video sources. We then present the generation pipeline that converts raw videos into image--language--3D flow triplets.

\subsection{Dataset Composition}
\begin{figure*}[ht]
\centering
\includegraphics[width=0.5\textwidth]{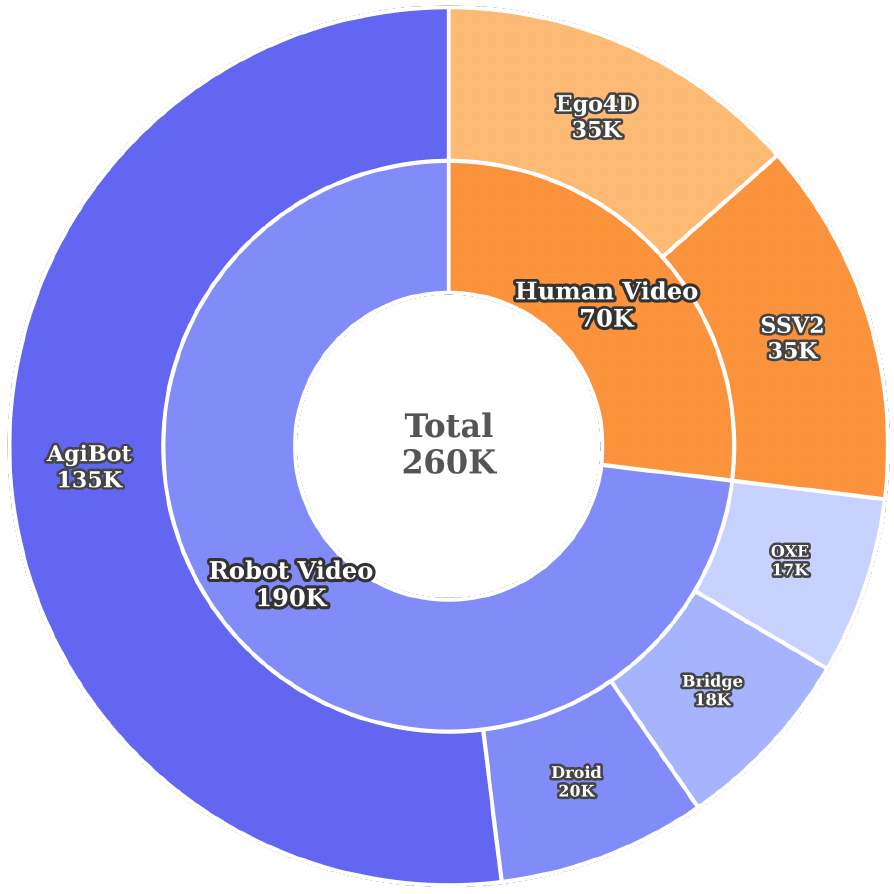}
\caption{\textbf{Composition of pre-training dataset.} The dataset contains 260K image--language--3D flow triplets in total, combining 190K trajectories from robot videos and 70K from human videos.}
\label{fig:dataset_composition}
\end{figure*}

Our dataset is constructed from heterogeneous human and robot video sources in order to cover a broad range of objects, environments, camera viewpoints, and manipulation styles. As summarized in Figure~\ref{fig:dataset_composition}, the final dataset contains 260K trajectories in total: 190K from robot demonstrations and 70K from human videos. The robot portion combines AgiBot~\cite{bu2025agibot} (135K), Droid~\cite{khazatsky2024droid} (20K), Open X-Embodiment~\cite{o2024open} (17K), and BridgeData V2~\cite{walke2023bridgedata} (18K). The human portion combines Ego4D~\cite{grauman2022ego4d} (35K) and Something-Something V2~\cite{goyal2017something} (35K).

\subsection{Dataset Generation Pipeline}
\begin{figure*}[ht]
\centering
\includegraphics[width=1.0\textwidth]{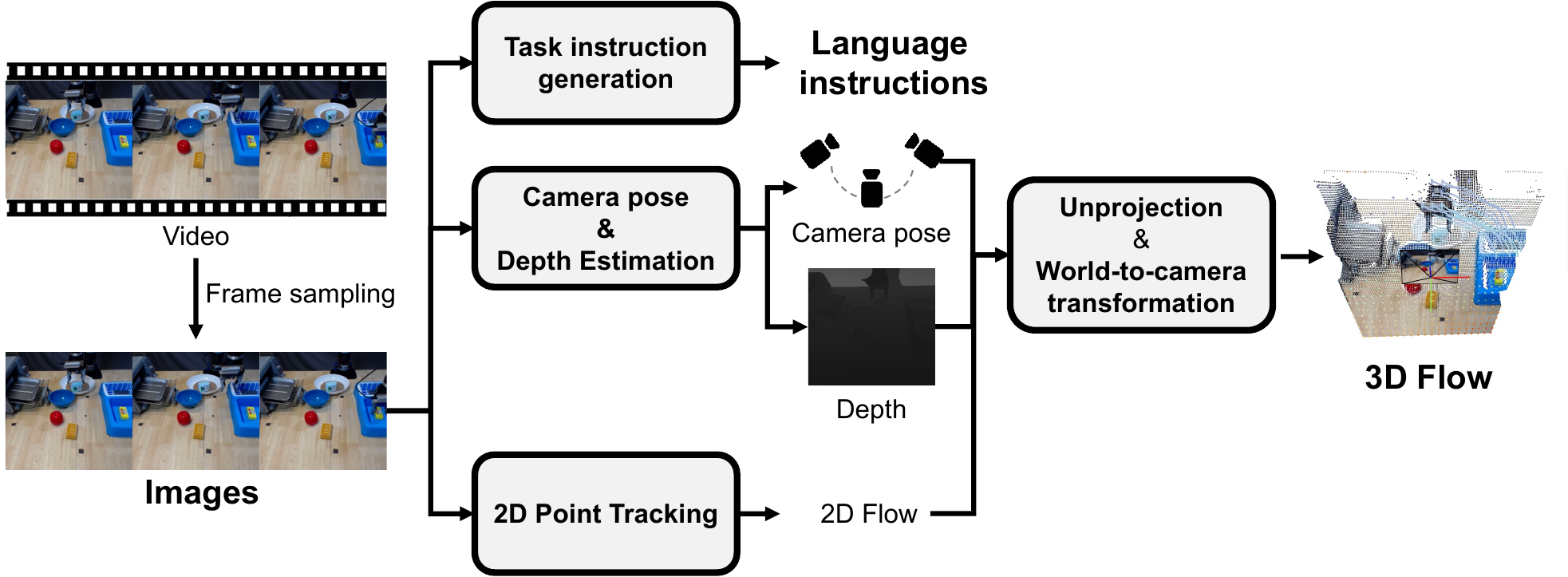}
\caption{\textbf{Dataset generation pipeline.} Each raw video is first frame-sampled to obtain image observations. Three parallel branches then process these images: (i) a VLM generates language instructions describing the manipulation intent, (ii) per-frame camera pose and depth are estimated using SpatialTrackerV2~\cite{xiao2025spatialtrackerv2}, and (iii) 2D points are tracked across frames using CoTracker3~\cite{karaev2025cotracker3}. Tracked 2D points are unprojected with the estimated depth and transformed into the reference camera coordinate frame to produce 3D flow trajectories that are invariant to camera motion.}
\label{fig:dataset_generation}
\end{figure*}

We follow the unified data generation pipeline of TraceForge~\cite{lee2025tracegen} with several modifications tailored to our setting. As illustrated in Figure~\ref{fig:dataset_generation}, our pipeline converts raw videos into aligned image--language--3D flow triplets. Compared with the original pipeline, we omit event chunking and speed retargeting, and instead directly sample frames from each video so that the effective temporal resolution is approximately matched across datasets collected at different frame rates while preserving the original motion timing.

\noindent\textbf{Language instruction generation.}
For each sampled trajectory, we generate language instructions that describe the underlying manipulation intent using a vision-language model (VLM).  The VLM takes as input a small set of representative frames sampled from the trajectory together with a prompt asking it to describe the task in three forms: a short imperative instruction, a detailed natural-language description, and a multi-step instruction that decomposes the task into sequential subgoals.

\noindent\textbf{3D flow generation.}
For each sampled trajectory, we construct a 3D flow that remains consistent under moving camera viewpoints. We select the reference frame from the early part of the trajectory, since the first frame may not always contain the robot or human demonstrator. On this reference frame, we place a uniform \(20 \times 20\) grid of keypoints and track them throughout the trajectory. Rather than representing motion in full camera coordinates, we represent each tracked point as \((x, y, z)\), where \((x, y)\) denotes the image-plane coordinates and \(z\) denotes the corresponding depth. This representation preserves spatial alignment with the original image while retaining physically meaningful motion in 3D space. 

To obtain the required 3D information, we estimate camera pose, depth, and point trajectories for every frame in the sampled trajectory. We use TAPIP3D~\cite{zhang2025tapip3d} for 3D flow construction, CoTracker3~\cite{karaev2025cotracker3} for point tracking, and a fine-tuned VGGT~\cite{wang2025vggt} model from SpatialTrackerV2~\cite{xiao2025spatialtrackerv2} for efficient depth and camera-pose prediction. Given a trajectory, these models produce per-frame depth maps, camera poses, and tracked 2D keypoint trajectories. We then unproject the tracked points with the predicted depth to reconstruct their 3D trajectories over time. 

To compensate for camera motion, we express all reconstructed 3D flow in the coordinate system of the reference camera frame. Specifically, we first transform the 3D points from world coordinates into the reference camera coordinates using the estimated camera extrinsics. We then project them back to the image plane using the camera intrinsics. The final 3D flow is stored as a screen-aligned sequence \(F_{t:t+K} = [x_i, y_i, z_i]_{i=t}^{t+K}\), where \(z_i\) denotes the depth value in the reference camera frame. This formulation compensates for camera motion and isolates true scene motion, rather than mixing it with viewpoint-dependent image-plane displacement. 
\section{Experiment Details}
\label{app:experimental_details}

\subsection{Pre-training \method}
\label{app:pretraining}
\noindent\textbf{Model architecture.}
As shown in Figure~\ref{fig:architecture}, \method consists of 
three modality encoders: image, language, and 3D flow. We 
describe each modality encoder below.

\begin{itemize}[leftmargin=1.6em, itemsep=0.7em]
    \item \textbf{Image encoder.}
    We initialize the image encoder with a pre-trained 
    DINOv2-Base (ViT-B/14) backbone and keep the entire backbone 
    trainable. Given an input image $I_t$, the backbone produces a 
    $\mathrm{[CLS]}$ token and a sequence of patch tokens, each 
    of dimension $768$. We form the per-frame embedding by 
    concatenating the $\mathrm{[CLS]}$ token with the 
    average-pooled patch tokens:
    \[
        d_t = \mathrm{CLS}(I_t) \oplus \sigma\big(\mathrm{Patch}(I_t)\big) 
        \in \mathbb{R}^{1536},
    \]
    where $\sigma(\cdot)$ denotes average pooling over patch 
    tokens. We apply the same procedure to every sampled frame 
    in the clip. An MLP fusion block then combines the 
    embeddings from each adjacent sampled frame pair to produce 
    the image-transition embedding $z_I$.

    \item \textbf{Language encoder.}
    We use a frozen T5-Base encoder with a learnable adapter on 
    top. Task instructions are tokenized with a maximum length 
    of $77$ tokens. The encoder produces a sequence of 
    $768$-dimensional token embeddings, from which we extract 
    the sentence-level representation via EOS-token pooling. The 
    pooled representation is then projected through an adapter 
    to obtain the language embedding $z_L$.

    \item \textbf{3D flow encoder.}
    The 3D flow encoder receives a sequence of $K$ timesteps of 
    $20 \times 20 \times 3$ flow data, representing 3D 
    displacement vectors at $20 \times 20$ keypoints. The 
    encoder consists of two stages: a 3D motion encoder and a 
    temporal motion transformer.
    \begin{itemize}[leftmargin=1.6em, itemsep=0.3em]
        \item \textbf{3D motion encoder.} A 4-layer CNN encodes 
        each timestep independently into a per-timestep feature.
        \item \textbf{Temporal motion transformer.} A 4-layer 
        transformer encoder aggregates information across the 
        temporal window. To incorporate visual context, we 
        prepend the current-frame image embedding $d_t$ as a 
        conditioning token. A learnable temporal $\mathrm{[CLS]}$ 
        token and positional embeddings are added to the 
        sequence. The temporal $\mathrm{[CLS]}$ output is then 
        projected through a linear layer to produce the 3D flow 
        embedding $z_F$.
    \end{itemize}
\end{itemize}

\noindent\textbf{Training protocol.}
Following R3M~\cite{nair2022r3m}, we sample five frames from each video clip during pre-training: an initial frame, a final frame, and three intermediate frames. The initial and final frames are sampled from the first 10\% and the last 10\% of the clip, respectively. The three intermediate frames are sampled from the remaining portion of the clip in temporal order. This sampling strategy yields an ordered frame sequence, from which we construct sequential transition pairs instead of a single pair such as $(I_t, I_{t+H})$. The pre-training hyperparameters are summarized in Table~\ref{tab:appendix_train_hparams}, and pre-training takes approximately 4 days on 4 NVIDIA L40S.

\begin{table}[h]
    \centering
    \captionsetup{skip=4pt}
    \caption{\textbf{Pre-training hyperparameters.} Loss weights, optimization settings, and augmentation parameters used to pre-train \method.}
    \label{tab:appendix_train_hparams}
    \footnotesize
    \setlength{\tabcolsep}{6pt}
    \begin{tabular}{l l c}
        \toprule
        \textbf{Category} & \textbf{Hyperparameter} & \textbf{Value} \\
        \midrule
        \multirow{5}{*}{Loss}
            & $\lambda_{\text{tcn}}$ & 1.0 \\
            & $\lambda_{\text{act}}$ & 1.0 \\
            & Contrastive temperature $\tau$ & 0.07 \\
            & Cosine regularization $\alpha$ & 1.0 \\
            & 3D flow temporal window of length $K$ & 7 \\
        \midrule
        \multirow{4}{*}{Optimization}
            & Optimizer & AdamW \\
            & Learning rate & $10^{-4}$ \\
            & Weight decay & $10^{-2}$ \\
            & Batch size & 32 \\
        \midrule
        \multirow{3}{*}{Augmentation}
            & Image resolution & $224 \times 224$ \\
            & Brightness / contrast jitter & 0.1 / 0.1 \\
            & Saturation / hue jitter & 0.05 / 0.02 \\
        \bottomrule
    \end{tabular}
\end{table}

\subsection{MetaWorld and RLBench}
\label{app:experimental_details_metaworld_rlbench}

We provide additional details for the MetaWorld and RLBench experiments used to evaluate downstream performance and control-relevant representations. These experiments follow a frozen-representation protocol: the image encoder remains fixed throughout downstream training, and only a lightweight three-layer MLP policy is optimized on top. Each policy receives a visual feature extracted from a \(224 \times 224\) third-person RGB observation, concatenated with the proprioceptive robot state.

\noindent\textbf{MetaWorld.}
MetaWorld evaluates single-task manipulation with a Sawyer arm and a two-finger gripper. We select 15 tasks that span multiple difficulty levels, following the task grouping used in prior work~\cite{seo2023masked}. The easy tasks are \textit{button-press}, \textit{drawer-open}, \textit{reach}, \textit{handle-pull}, \textit{peg-unplug-side}, \textit{lever-pull}, and \textit{dial-turn}. The medium tasks are \textit{hammer}, \textit{sweep-into}, \textit{bin-picking}, \textit{push-wall}, and \textit{box-close}. The hard and very hard tasks are \textit{assembly}, \textit{hand-insert}, and \textit{shelf-place}. For each task, we collect 25 demonstrations using the official scripted policy and use only the corner-view camera as visual input. Table~\ref{tab:metaworld_results} reports the detailed MetaWorld success rates grouped by task difficulty.

\begin{table}[ht]
    \centering
    \captionsetup{skip=4pt}
    \caption{\textbf{MetaWorld success rates.} Detailed success rates (\%) grouped by task difficulty. \textbf{Bold} and \underline{underlined} numbers indicate the best and second-best results in each column, respectively.}
    \label{tab:metaworld_results}
    \setlength{\tabcolsep}{4pt}
    \footnotesize
    \begin{tabular}{@{}l c c c c@{}}
        \toprule
        Algorithm & Easy (7) & Medium (5) & Hard \& Very Hard (3) & \textbf{Mean} \\
        \midrule
        R3M~\cite{nair2022r3m}       & 78.3          & 68.0          & \underline{68.0}          & 72.8 \\
        VC-1~\cite{majumdar2023we}   & 62.6          & 71.6          & 38.7                      & 60.8 \\
        LIV~\cite{ma2023liv}         & \underline{79.4} & 76.8       & 66.7                      & \underline{76.0} \\
        CLIP~\cite{radford2021learning} & 72.9       & 68.8          & 42.0                      & 65.3 \\
        DINOv2~\cite{oquab2024dinov2} & 77.7         & \underline{77.6} & 64.0                   & 74.9 \\
        SigLIP~\cite{zhai2023sigmoid} & 74.3         & 72.8          & 56.7                      & 70.4 \\
        \method (Ours) & \textbf{81.1} & \textbf{81.6} & \textbf{69.3} & \textbf{78.9} \\
        \bottomrule
    \end{tabular}
\end{table}

\noindent\textbf{RLBench.}
RLBench evaluates visuomotor manipulation with a Franka Panda arm. We evaluate six tasks: \textit{close box}, \textit{put rubbish in bin}, \textit{close laptop lid}, \textit{water plants}, \textit{unplug charger}, and \textit{toilet seat down}. For each task, we collect 100 demonstration trajectories using the Open Motion Planning Library (OMPL)~\cite{sucan2012open} and use only the front-view camera as visual input. Table~\ref{tab:rlbench_success_rates} reports the task-wise RLBench success rates for each encoder.

\begin{table}[ht]
    \centering
    \captionsetup{skip=4pt}
    \caption{\textbf{RLBench success rates.} Detailed task-wise success rates (\%). \textbf{Bold} and \underline{underlined} numbers indicate the best and second-best results in each column, respectively.}
    \label{tab:rlbench_success_rates}
    \setlength{\tabcolsep}{4pt}
    \footnotesize
    \begin{tabular}{@{}l c c c c c c c@{}}
        \toprule
        Algorithm & \makecell[c]{close\\ box} & \makecell[c]{put rubbish\\ in bin} & \makecell[c]{close laptop\\ lid} & \makecell[c]{water\\ plants} & \makecell[c]{unplug\\ charger} & \makecell[c]{toilet seat\\ down} & \textbf{Mean} \\ \midrule
        R3M~\cite{nair2022r3m}       & \textbf{96}     & 4      & 56     & \underline{8}     & 20      & \underline{92}      & 46.0      \\
        VC-1~\cite{majumdar2023we}      & 84     & \textbf{12}     & \underline{72}     & \underline{8}      & 16     & \textbf{96}     & 48.0   \\
        LIV~\cite{ma2023liv}       & \underline{92}      & \underline{8}      & \textbf{76}      & 4      & 20      & \underline{92}      & \underline{48.6}      \\ 
        CLIP~\cite{radford2021learning}      & 60     & 0      & 56     & 4     & 12     & 80     & 35.3      \\
        DINOv2~\cite{oquab2024dinov2}    & 84     & \textbf{12}     & \textbf{76}     & 4      & \underline{24}     & 84     & 47.3   \\
        SigLIP~\cite{zhai2023sigmoid}    & 80     & 4      & 52     & 0      & 12     & 76     & 37.3      \\ 
        \method (Ours)      & 88     & \underline{8}      & \textbf{76}      & \textbf{20}      & \textbf{36}      & \textbf{96}      & \textbf{54.0}    \\ \bottomrule
    \end{tabular}
\end{table}

\noindent\textbf{Training and evaluation protocol.}
For both benchmarks, we train each method for 100 epochs with the visual encoder kept frozen throughout downstream training. Every 10 epochs, we evaluate the policy using 25 rollouts. We then select the best-performing checkpoint across training and report its average rollout success rate.

\subsection{LIBERO}
\label{app:experimental_details_libero}

We evaluate \method on five LIBERO suites: LIBERO-90, LIBERO-Goal, LIBERO-Object, LIBERO-Spatial, and LIBERO-Long. LIBERO-90 contains 90 tasks, while each of the other four suites contains 10 tasks with 50 demonstrations per task.

\noindent\textbf{Model architecture.}
We adopt Diffusion Policy~\cite{chi2025diffusion} as the downstream imitation-learning policy, using a U-Net backbone with channel dimensions \([256, 512, 1024]\). We use DDIM~\cite{song2020denoising} for diffusion-based action generation, with 100 forward diffusion steps and 10 denoising steps during inference. We set the prediction horizon to 32, the execution horizon to 16, and the observation history to 1.

For visual input, we use only third-person RGB observations and exclude gripper-view images. The image encoder output serves as the visual conditioning vector for the diffusion policy. For CNN-based encoders, we obtain the global image feature by applying global average pooling to the final feature map of the ResNet backbone. For ViT-based encoders, we concatenate the $\mathrm{[CLS]}$ token with the average-pooled patch tokens to form the image feature.

Language instructions are encoded using the corresponding text encoder when available. For R3M, VC-1, and DINOv2, which do not provide native text encoders, we use the CLIP text encoder. CLIP, LIV, and \method use the $\mathrm{[EOS]}$ token representation as the sentence-level language feature, whereas SigLIP uses mean pooling over all token embeddings.

\noindent\textbf{Training and evaluation protocol.}
Our primary LIBERO setting follows a reusable-encoder protocol: both the image and language encoders remain frozen, and only the diffusion policy is trained. This setting directly evaluates whether each pre-trained representation can transfer to downstream policy learning without task-specific encoder adaptation. As an additional comparison, we also report a LoRA setting that adapts both encoders jointly with the diffusion policy.

For each LIBERO suite, we train a separate diffusion policy using demonstrations from that suite and evaluate it on the corresponding suite. Each method is trained for 200 epochs. Every 20 epochs, we evaluate the policy using 20 rollouts per task. We then select the best-performing checkpoint across training and report its average rollout success rate.

\subsection{Real-world Robot}
\label{app:experimental_details_real}

\begin{figure*}[h]
\centering
\includegraphics[width=0.5\textwidth]{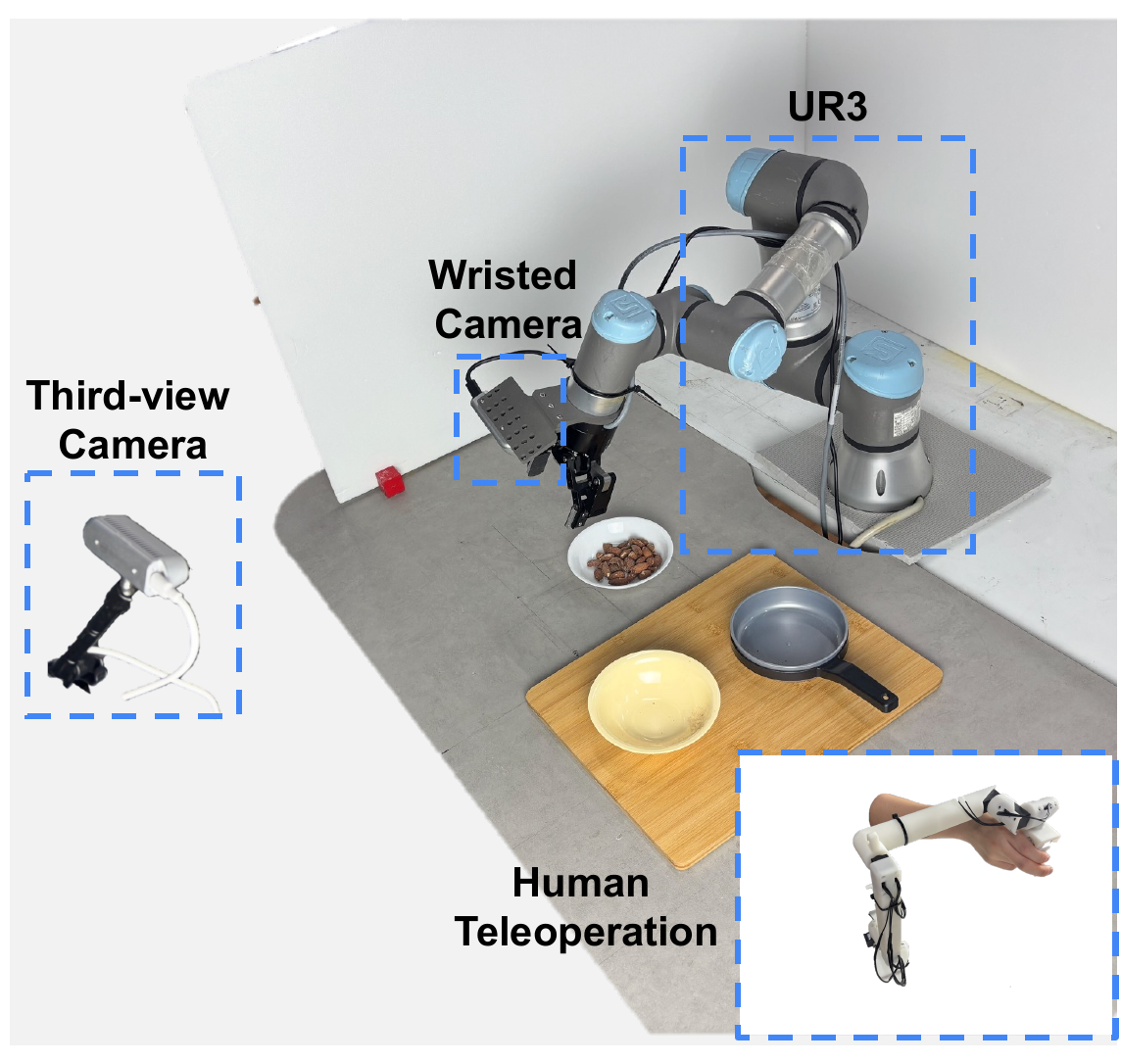}
\caption{\textbf{UR3 hardware setup.} Real-robot setup used for demonstration collection and policy evaluation.}
\label{fig:real_robot_setup}
\end{figure*}

\noindent\textbf{Hardware setup.}
Figure~\ref{fig:real_robot_setup} shows the real-robot setup used for demonstration collection and policy evaluation. A fixed-base UR3 manipulator equipped with a two-finger gripper performs all manipulation tasks. Two RGB cameras, one third-person camera and one wrist-mounted camera, provide $224 \times 224$ visual observations. The policy also receives a 7D proprioceptive state consisting of the 6D end-effector pose and the gripper state. During demonstration collection, a human teleoperator controls the end-effector pose and gripper command through a custom teleoperation interface.

\begin{figure*}[p]
\centering
\includegraphics[width=0.84\textwidth,height=0.95\textheight,keepaspectratio]{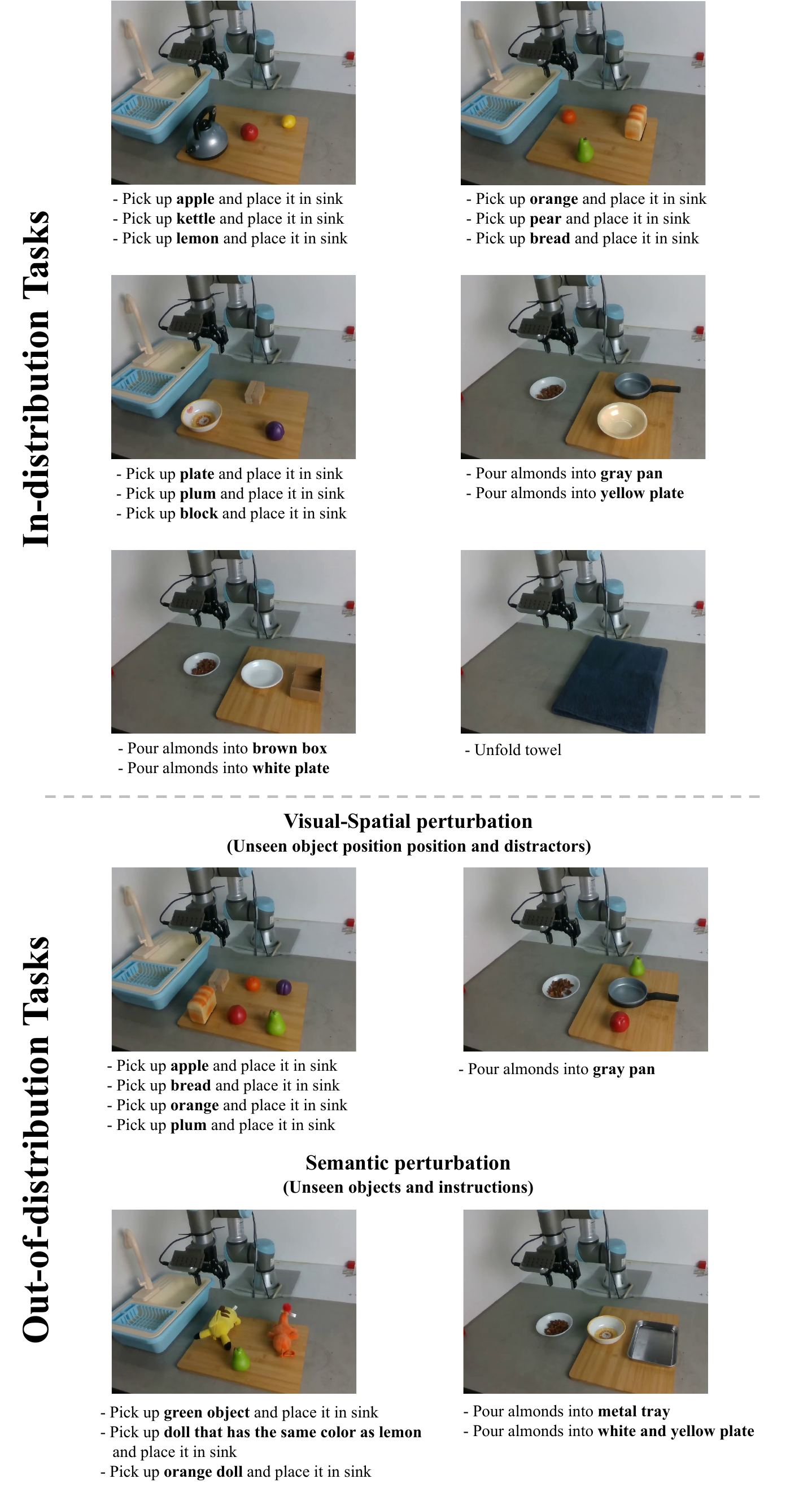}
\caption{\textbf{Real-world evaluation tasks.} Illustration of the in-distribution and out-of-distribution evaluation settings for the three real-world tasks, including the exact task instructions.}
\label{fig:evaluation_tasks}
\end{figure*}

\noindent\textbf{Task and data collection.}
We evaluate \method on three representative real-world manipulation tasks: \textit{Pick <object> into Sink}, \textit{Pour almonds into <object>}, and \textit{Unfold Towel} (see Figure~\ref{fig:evaluation_tasks} for in-distribution examples). These tasks cover both rigid-object manipulation and deformable-object interaction.

For \textit{Pick <object> into Sink}, the robot picks up the instructed object and places it in the sink. The object set contains nine objects: apple, block, bread, kettle, lemon, orange, pear, plate, and plum. We collect 10 demonstrations per object, yielding 90 trajectories in total. To control scene variation during training, we divide the nine objects into three groups of three and collect demonstrations for each group under a fixed scene layout. The task prompt is ``pick up <object> and place it in sink.''

For \textit{Pour almonds into <object>}, the robot grasps a flat plate containing almonds and pours them into the instructed target object. The target set contains four objects: brown box, gray pan, white plate, and yellow plate. We collect 20 demonstrations per target object, yielding 80 trajectories in total. Across demonstrations, the source object remains fixed and only the target object changes. The task prompt is ``pour almonds into <object>.''

For \textit{Unfold Towel}, the robot unfolds a towel initially folded in half. We collect 50 trajectories for this task. The task requires multi-stage deformable-object manipulation: the robot first opens the folded towel by grasping its middle region and then unfolds the two side edges. The task prompt is ``unfold towel.''

\noindent\textbf{Model architecture.}
We integrate the frozen pre-trained image encoder into the pre-trained $\pi_{0.5}$ through a lightweight visual-injection design inspired by plug-in visual injection (PVI)~\cite{zhang2026pvi}. Our design enables parameter-efficient fine-tuning by freezing the $\pi_{0.5}$ backbone and injecting auxiliary visual features into the action expert through a ControlNet-style~\cite{zhang2023adding} side branch.

We augment the original $\pi_{0.5}$ pathway with three lightweight components: (i) an auxiliary visual encoder that processes each camera view into a sequence of patch tokens, (ii) a projection layer that maps these features to the VLA's hidden dimension, and (iii) a trainable copy of the action expert that conditions on these auxiliary features. At each layer of the action expert, the trainable copy produces a residual signal that is added to the hidden state of the frozen main path, and the final action is predicted from the modified hidden state.

During fine-tuning, we freeze both the $\pi_{0.5}$ backbone and the auxiliary visual encoder, and optimize only the lightweight injection modules (projection layer, trainable copy of the action expert, and per-layer injectors). The trainable copy is initialized from the pre-trained action expert, and the projection and injection modules are initialized to zero. This makes the initial policy equivalent to the pre-trained VLA and allows the injected visual signal to become active gradually during training. For all real-robot comparisons, we keep the fine-tuning protocol fixed and change only the auxiliary visual encoder. This protocol isolates the effect of the visual representation from the effect of the VLA fine-tuning strategy.

\noindent\textbf{Feature extraction details.}
We extract patch-level features differently for CNN-based and ViT-based visual encoders because the two architectures produce spatial features in different forms. The resulting features are fed into the projection layer described above.

\begin{itemize}[leftmargin=1.6em, labelsep=0.8em, itemsep=0.7em, 
topsep=0.4em]
    \item \textbf{CNN-based encoders.} We use the output of the final convolution block before spatial pooling, producing a feature map with shape $B \times C \times H \times W$. We flatten the spatial dimensions to obtain a sequence of $H \cdot W$ patch tokens with shape $B \times (H \cdot W) \times C$.

    \item \textbf{ViT-based encoders.} We use the patch tokens (excluding the $\mathrm{[CLS]}$ token) directly, producing $N$ patch tokens arranged in a $\sqrt{N} \times \sqrt{N}$ spatial grid, with sequence shape $B \times N \times C$.
\end{itemize}

\begin{table*}[t!]
\centering
\footnotesize
\caption{\textbf{Real-robot fine-tuning hyperparameters.} Task-specific training hyperparameter settings for fine-tuned $\pi_{0.5}$ policies. Each subtable corresponds to one real-world task.}
\label{tab:real_robot_training_hyperparameters}
\begin{minipage}[t]{0.31\textwidth}
\centering
\textbf{Pick <object> into Sink}

\vspace{0.25em}
\begin{tabular}{@{}lc@{}}
\toprule
Hyperparameter & Value \\
\midrule
Action dimension & 32 \\
Action horizon & 50 \\
Batch size & 32 \\
Optimizer & AdamW \\
Peak learning rate & $1.5 \times 10^{-5}$ \\
Final learning rate & $1.5 \times 10^{-6}$ \\
Warmup steps & 500 \\
Decay steps & 5{,}000 \\
Training steps & 5{,}000 \\
\bottomrule
\end{tabular}
\end{minipage}
\hfill
\begin{minipage}[t]{0.31\textwidth}
\centering
\textbf{Pour almonds into <object>}

\vspace{0.25em}
\begin{tabular}{@{}lc@{}}
\toprule
Hyperparameter & Value \\
\midrule
Action dimension & 32 \\
Action horizon & 50 \\
Batch size & 32 \\
Optimizer & AdamW \\
Peak learning rate & $1.5 \times 10^{-5}$ \\
Final learning rate & $1.5 \times 10^{-6}$ \\
Warmup steps & 700 \\
Decay steps & 7{,}000 \\
Training steps & 7{,}000 \\
\bottomrule
\end{tabular}
\end{minipage}
\hfill
\begin{minipage}[t]{0.31\textwidth}
\centering
\textbf{Unfold Towel}

\vspace{0.25em}
\begin{tabular}{@{}lc@{}}
\toprule
Hyperparameter & Value \\
\midrule
Action dimension & 32 \\
Action horizon & 50 \\
Batch size & 32 \\
Optimizer & AdamW \\
Peak learning rate & $1.5 \times 10^{-5}$ \\
Final learning rate & $1.5 \times 10^{-6}$ \\
Warmup steps & 1{,}000 \\
Decay steps & 10{,}000 \\
Training steps & 10{,}000 \\
\bottomrule
\end{tabular}
\end{minipage}
\end{table*}

\noindent\textbf{Training and evaluation protocol.}
For each real-world task, we fine-tune a separate $\pi_{0.5}$ policy from the same pre-trained base checkpoint, following the visual-injection setup described above. 
Table~\ref{tab:real_robot_training_hyperparameters} summarizes the task-specific hyperparameters.

After training, we evaluate each policy through closed-loop real-robot rollouts (20 rollouts per setting). The evaluation includes in-distribution trials for all three tasks. For \textit{Pick <object> into Sink} and \textit{Pour almonds into <object>}, we additionally evaluate two out-of-distribution perturbation types: visual-spatial perturbations and semantic perturbations. For \textit{Unfold Towel}, we evaluate only the in-distribution setting. Figure~\ref{fig:evaluation_tasks} summarizes the overall real-world evaluation settings and the exact task instructions.

A rollout is successful when the robot completes the instructed task within the episode horizon. For \textit{Pick <object> into Sink}, success requires placing the instructed object inside the sink. For \textit{Pour almonds into <object>}, success requires pouring the almonds into the instructed target object. For \textit{Unfold Towel}, success requires unfolding both folded edges by the end of the episode.

\begin{figure*}[p]
\centering
\includegraphics[width=\textwidth]{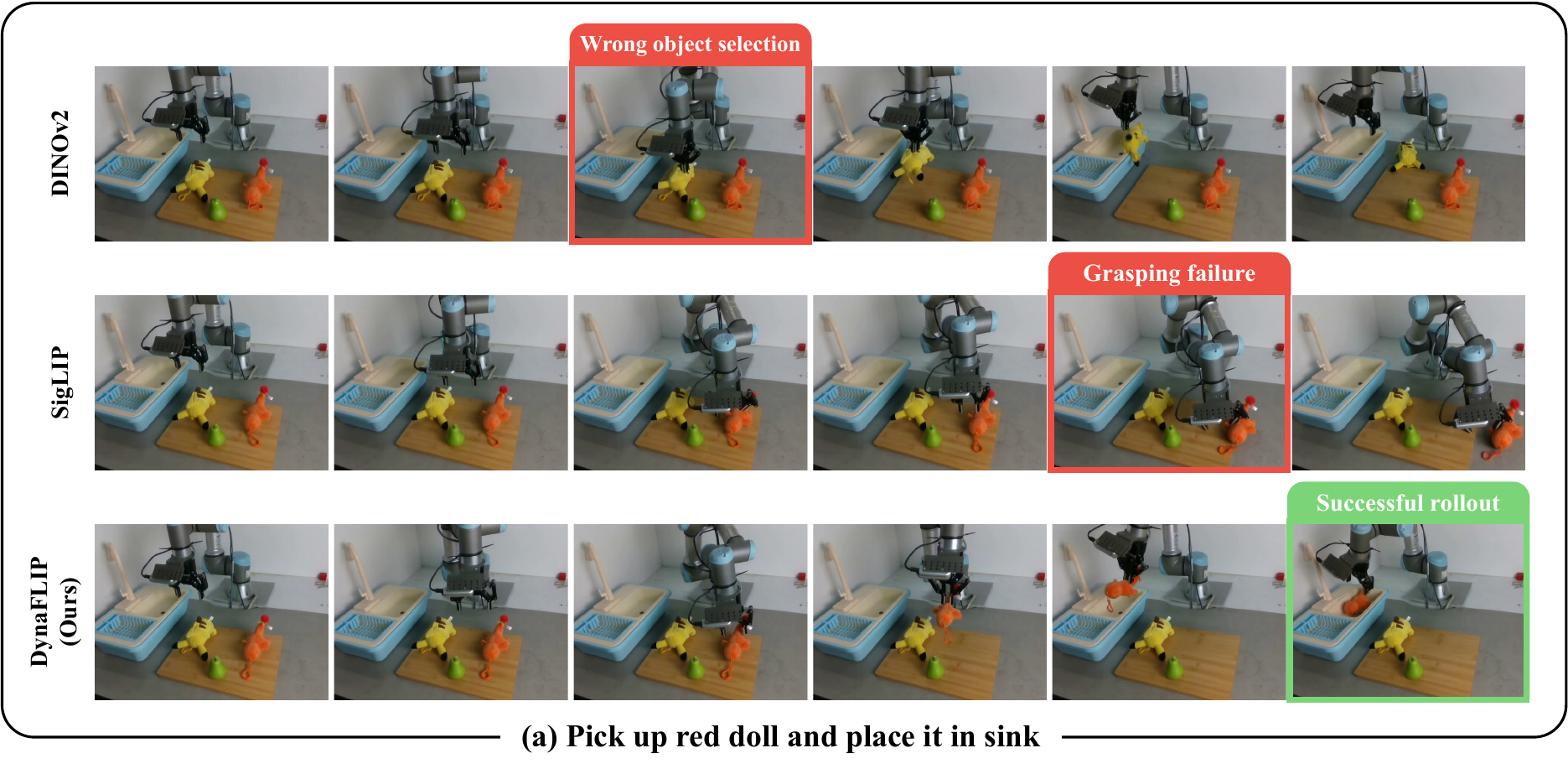}

\vspace{0.5em}

\includegraphics[width=\textwidth]{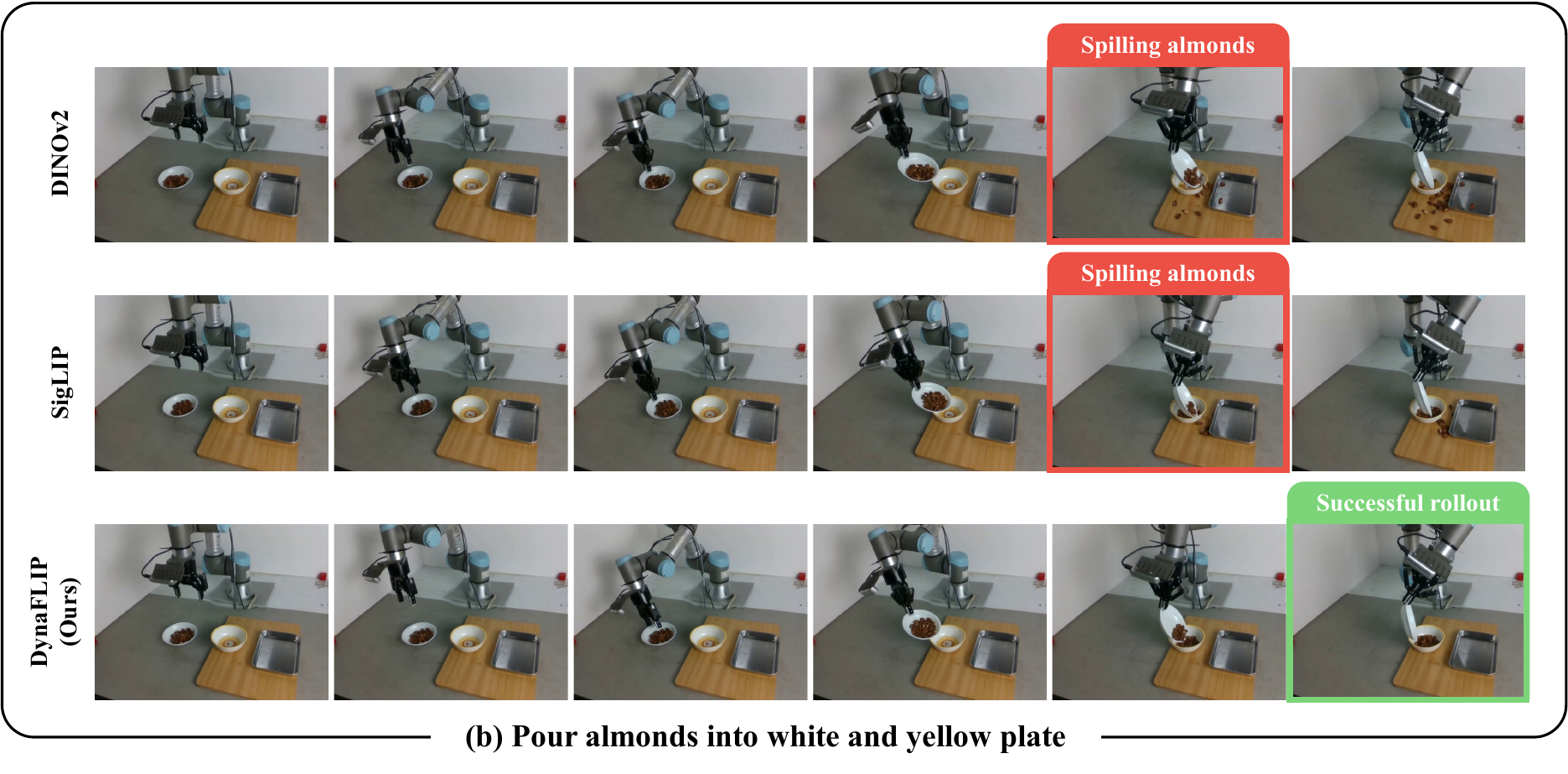}

\vspace{0.5em}

\includegraphics[width=\textwidth]{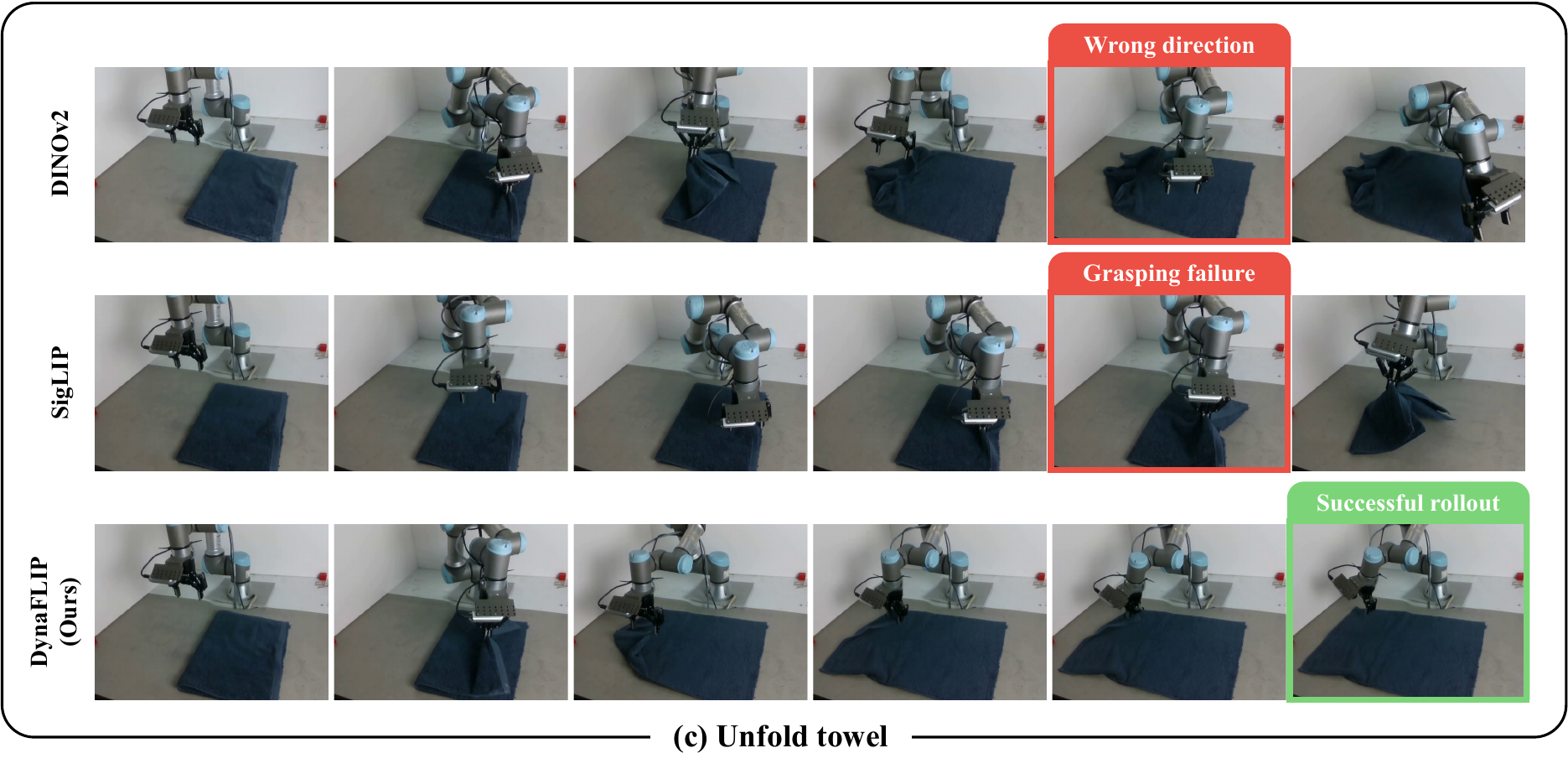}
\caption{\textbf{Representative rollout examples on three real-world tasks.} We compare \method with DINOv2 and SigLIP on 
\textbf{(a)} \textit{Pick up red doll and place it in sink} 
(OOD), \textbf{(b)} \textit{Pour almonds into white and yellow 
plate} (OOD), and \textbf{(c)} \textit{Unfold towel} 
(in-distribution). Baselines exhibit distinct failure modes (wrong object selection, grasping failure, spilling, wrong direction), while \method completes all three tasks successfully.}
\label{fig:real_world_rollout_examples}
\end{figure*} 

\subsection{Control-Relevant Metric}
\label{app:control_relevant_metric}

We adopt the simulator-grounded state prediction metric proposed in~\cite{dong2026capturing} as a quantitative proxy for representation quality. This metric measures how well a visual representation preserves state information relevant to downstream control. We train lightweight probes to predict simulator state from visual features, and computes a normalized score \(S_m\) from the prediction errors.

\noindent\textbf{Simulator state.}
For a scene with $N_o$ objects, the simulator state combines object-level and scene-level information:
\begin{itemize}[leftmargin=1.6em, labelsep=0.8em, itemsep=0.3em, 
topsep=0.4em]
    \item \textbf{Object-level state} for each object $i$: 
    position $p^i_{\mathrm{pose}} \in \mathbb{R}^3$, orientation 
    $q^i_{\mathrm{pose}} \in \mathbb{R}^4$, and bounding-box 
    shape $s^i_{\mathrm{shape}} \in \mathbb{R}^3$.
    \item \textbf{Scene-level state}: robot joint configuration 
    $q^J \in \mathbb{R}^{N_j}$ and end-effector pose $p^{ee} 
    \in \mathbb{R}^{N_{ee}}$.
\end{itemize}
The full simulator state $s$ concatenates all object-level 
states and the scene-level state.

\noindent\textbf{State prediction probe.}
Given an input image $I$, we extract a spatial feature map and a global feature (see Feature extraction details below). Two probes predict the simulator state:
\begin{itemize}[leftmargin=1.6em, labelsep=0.8em, itemsep=0.3em, 
topsep=0.4em]
    \item \textbf{Object-level probe} uses the \emph{feature map}: it predicts each object's state from RoI-pooled features inside its bounding box.
    \item \textbf{Scene-level probe} uses the \emph{global feature}: it predicts the robot and end-effector states.
\end{itemize}
Both probes are lightweight regressors trained on top of the frozen visual features.

\noindent\textbf{Control-relevant score.}
For each state dimension $a$ and model $m$, we compute the raw prediction score $r_{m,a}$ as the negative mean squared error between predicted and ground-truth values across all examples.
To compare models on a unified scale, we min-max normalize $r_{m,a}$ across models within each state dimension and average the normalized scores:
\[
S_m = \frac{1}{|A|} \sum_{a \in A} 
\frac{r_{m,a} - \min_{\tilde m} r_{\tilde m,a}}
{\max_{\tilde m} r_{\tilde m,a} - \min_{\tilde m} r_{\tilde m,a}},
\]
where $A$ denotes the set of evaluated state dimensions. A larger $S_m$ indicates that the representation preserves more control-relevant information.

\noindent\textbf{Feature extraction details.}
We extract feature maps and global features differently for CNN-based and ViT-based visual encoders because the two architectures produce spatial features in different forms.

\begin{itemize}[leftmargin=1.6em, labelsep=0.8em, itemsep=0.7em, 
topsep=0.4em]
    \item \textbf{CNN-based encoders.}
    \begin{itemize}[leftmargin=1.6em, labelsep=0.5em, 
    itemsep=0.3em, topsep=0.3em]
        \item \textit{Feature map:} the output of the final 
        convolution block before spatial pooling, with shape $B 
        \times C \times H \times W$.
        \item \textit{Global feature:} the feature vector 
        obtained by global average pooling over spatial 
        dimensions, with shape $B \times C$.
    \end{itemize}
    
    \item \textbf{ViT-based encoders.}
    \begin{itemize}[leftmargin=1.6em, labelsep=0.5em, 
    itemsep=0.3em, topsep=0.3em]
        \item \textit{Feature map:} patch tokens (excluding the 
        $\mathrm{[CLS]}$ token), reshaped into a 2D spatial grid 
        with shape $B \times C \times \sqrt{N} \times \sqrt{N}$.
        \item \textit{Global feature:} the concatenation of the 
        $\mathrm{[CLS]}$ token and the average-pooled patch 
        tokens, with shape $B \times 2C$.
    \end{itemize}
\end{itemize}

\section{Additional Experimental Results}
\label{sec:additional_experimental_results}

\subsection{LIBERO Results with Paired Image Encoders}
\label{app:libero_paired_encoders}

Recent vision-language-action models commonly pair multiple image encoders to combine complementary visual features~\cite{kim2025openvla, bu2025univla, zheng2024tracevla, 
lee2025tracegen, kim2025fine}: DINOv2 provides fine-grained, low-level spatial features, while language-aligned encoders such as CLIP and SigLIP capture high-level semantics. To assess whether \method remains beneficial in this setting, we pair DINOv2 with each language-aligned vision encoder---CLIP, SigLIP, and \method---and evaluate them on LIBERO under the same frozen configuration as in Section~\ref{sec:libero}. The image encoders are concatenated at the feature level before being passed to the diffusion policy, and the corresponding language encoder is paired with each setup. 

Table~\ref{tab:libero_multi_encoder_results} reports the results. DINOv2 + \method achieves the highest mean success rate, outperforming both DINOv2 + CLIP and DINOv2 + SigLIP. We attribute this advantage to two complementary properties of \method's representations. First, like CLIP and SigLIP, \method aligns visual features with language and therefore provides the semantic grounding required for instruction following. Second, through dynamics-aware pre-training, \method focuses on control-relevant regions critical for manipulation---a signal that purely image-text contrastive encoders do not provide. Together, these properties make \method a more effective language-aligned counterpart to DINOv2's fine-grained spatial features.

\begin{table}[h]
    \centering
    \caption{\textbf{LIBERO benchmark results with paired image encoders.} We combine DINOv2 with various language-aligned vision encoders. All encoders are kept frozen, and only the diffusion policy is trained. The evaluation metric is success rate (\%). \textbf{Bold} and \underline{underline} numbers indicate the best and second-best results in each column, respectively.}
    \label{tab:libero_multi_encoder_results}
    \setlength{\tabcolsep}{4pt}
    \footnotesize
    \begin{tabular}{ll ccccc}
        \toprule
        \multirow{2}{*}{\begin{tabular}[c]{@{}l@{}}\textbf{Image}\\ \textbf{Encoders}\end{tabular}} & \multirow{2}{*}{\begin{tabular}[c]{@{}l@{}}\textbf{Language}\\ \textbf{Encoder}\end{tabular}} & \multicolumn{5}{c}{\textbf{Frozen}} \\
        \cmidrule(lr){3-7}
        & & \textbf{Goal} & \textbf{Object} & \textbf{Spatial} & \textbf{Long} & \textbf{Mean} \\
        \midrule
        DINOv2 + CLIP & CLIP & 68.0 & 52.0 & \textbf{53.5} & 24.0 & 49.4 \\
        DINOv2 + SigLIP & SigLIP & \textbf{75.5} & \underline{60.5} & 48.0 & \underline{25.0} & \underline{52.3} \\
        DINOv2 + \method (Ours) & \method (Ours) & \underline{72.5} & \textbf{73.5} & \underline{48.5} & \textbf{27.0} & \textbf{55.4} \\
        \bottomrule
    \end{tabular}
\end{table}

\subsection{Data Scaling}
\label{app:data_scaling}

To examine how \method scales with pre-training data, we pre-train on 5\%, 20\%, and 100\% of our 260K-trajectory dataset under an identical training schedule, and evaluate each checkpoint on LIBERO-Goal and LIBERO-Object with Diffusion Policy. Both the image and language encoders remain frozen, so the comparison reflects representation quality rather than policy adaptation. Table~\ref{tab:data_scaling} reports the results. Performance increases consistently with data size on both suites, suggesting that \method benefits from scaling to larger human and robot video corpora. We leave a model-size scaling study to future work, as pre-training larger backbones exceeds our available compute.

\begin{table}[ht]
\centering
\caption{\textbf{Data scaling.} All encoders are kept frozen, and only the diffusion policy is trained. We report success rate (\%) on LIBERO-Goal and LIBERO-Object as a function of the fraction of pre-training data. \textbf{Bold} indicates the best result in each column.}
\label{tab:data_scaling}
\small
\begin{tabular}{lccc}
\toprule
\textbf{Data fraction} & \textbf{Goal} & \textbf{Object} & \textbf{Mean} \\
\midrule
5\%   & 46.0 & 3.0  & 24.5 \\
20\%  & 62.5 & 23.5 & 43.0 \\
100\% & \textbf{70.5} & \textbf{37.5} & \textbf{54.0} \\
\bottomrule
\end{tabular}
\end{table}

\subsection{Backbone Generality}
\label{app:backbone_generality}

The \method objective contains no component specific to DINOv2~\cite{oquab2024dinov2}. To verify that it applies to other visual backbones, we train a ResNet-50~\cite{he2016deep} from random initialization with the same objective and evaluate it on MetaWorld under the frozen-encoder protocol of Section~\ref{sec:analysis}. Table~\ref{tab:backbone_generality} compares this variant with our default DINOv2-B/14 backbone. The ResNet-50 variant reaches a mean success rate of 78.5\%, close to the DINOv2-B/14 variant (78.9\%) and above DINOv2 without our pre-training (74.9\%, Table~\ref{tab:metaworld_results}). The gains therefore stem from the objective rather than from properties specific to DINOv2. We adopt DINOv2 as the default initialization because recent VLA models predominantly use ViT backbones~\cite{kim2025openvla, black2024pi_0}, allowing direct integration, and because its fine-grained spatial features provide a strong starting point for manipulation.

\begin{table}[ht]
\centering
\caption{\textbf{Backbone generality.} The image encoder is kept frozen, and only the MLP policy is trained. We report MetaWorld success rates (\%) grouped by task difficulty for \method pre-trained with different visual backbones. \textbf{Bold} indicates the best result in each column.}
\label{tab:backbone_generality}
\small
\begin{tabular}{lcccc}
\toprule
\textbf{Backbone} & \textbf{Easy (7)} & \textbf{Medium (5)} & \textbf{Hard \& Very Hard (3)} & \textbf{Mean} \\
\midrule
ResNet-50 (from scratch) & 77.4 & \textbf{85.5} & 69.3 & 78.5 \\
DINOv2-B/14 (default)    & \textbf{81.1} & 81.6 & 69.3 & \textbf{78.9} \\
\bottomrule
\end{tabular}
\end{table}

\subsection{Grad-CAM Visualizations}
\label{app:additional_gradcam}

\noindent\textbf{Visualization protocol.}
We use the PyTorch-Grad-CAM library to generate Grad-CAM visualizations and identify the image regions that contribute most to downstream action prediction. For each frozen image encoder, we compute Grad-CAM with respect to a scalar target defined as the negative mean squared error between the action predicted by the trained three-layer MLP policy head and the ground-truth action. With this choice, the resulting heatmap highlights the visual regions that most strongly support accurate action prediction. As the target layer, we use the final convolutional layer (\texttt{model.layer4[-1]}) for CNN-based encoders and the pre-attention normalization layer in the last Transformer block (\texttt{model.blocks[-1].norm1}) for ViT-based encoders.

\noindent\textbf{Additional visualizations.}
We provide additional Grad-CAM visualizations to complement the qualitative analysis in Section~\ref{sec:analysis}. These examples further show that \method consistently focuses on task-relevant objects and interaction regions, while baseline encoders more often exhibit diffuse attention or place substantial emphasis on less control-relevant regions.

{\setlength{\tabcolsep}{0.25mm}
\begin{longtable}{
    >{\centering\arraybackslash}m{0.0175\linewidth}
    *{8}{>{\centering\arraybackslash}m{0.1175\linewidth}}
@{}}
    \caption{Grad-CAM visualizations}\label{tab:gradcam_appendix} \\
    \toprule
    & \small Task & \small R3M & \small VC-1 & \small LIV & \small CLIP & \small DINOv2 & \small SigLIP & \small \textbf{\method} \\ \midrule
    \endfirsthead

    \caption[]{Grad-CAM visualizations (Continued)} \\
    \toprule
    & \small Task & \small R3M & \small VC-1 & \small LIV & \small CLIP & \small DINOv2 & \small SigLIP & \small \textbf{\method} \\ \midrule
    \endhead

    \midrule
    \multicolumn{9}{r}{\small Continued on next page} \\
    \endfoot

    \bottomrule
    \endlastfoot

    \rotatebox[origin=c]{90}{\scriptsize Assembly} & 
    \includegraphics[width=1\linewidth, valign=m]{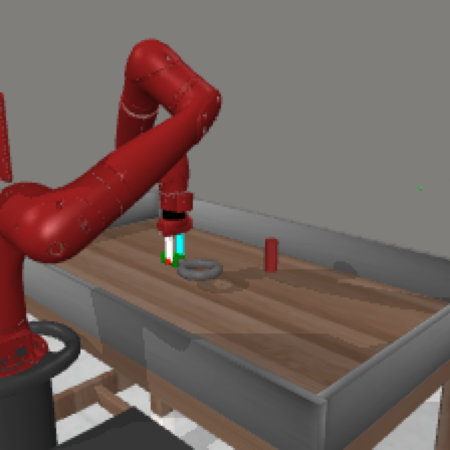} & 
    \includegraphics[width=1\linewidth, valign=m]{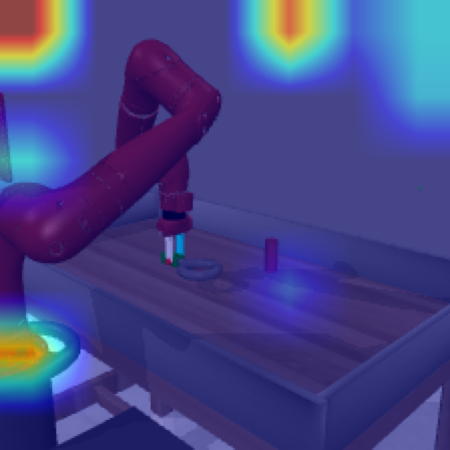} &
    \includegraphics[width=1\linewidth, valign=m]{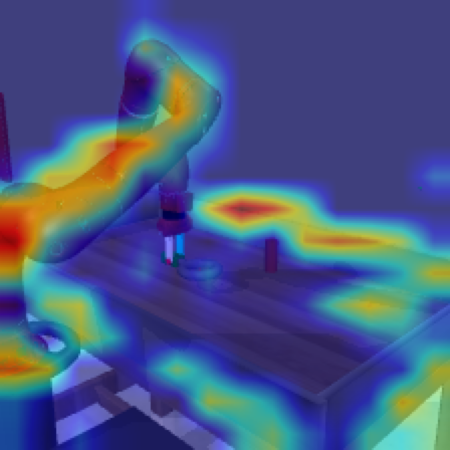} &
    \includegraphics[width=1\linewidth, valign=m]{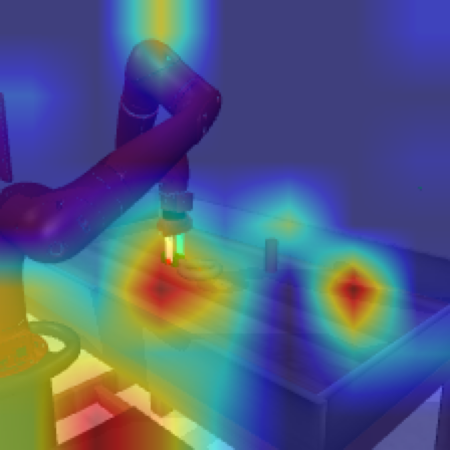} &
    \includegraphics[width=1\linewidth, valign=m]{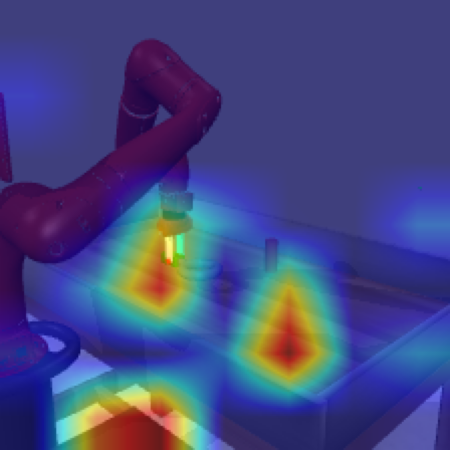} &
    \includegraphics[width=1\linewidth, valign=m]{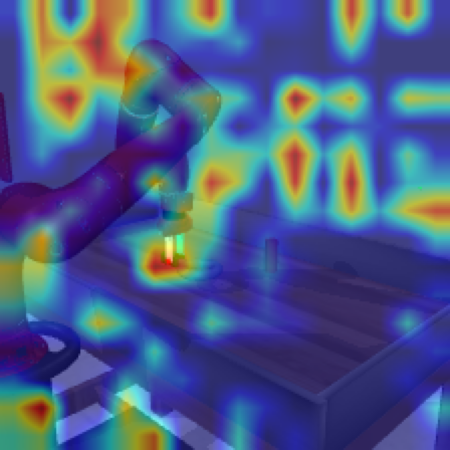} &
    \includegraphics[width=1\linewidth, valign=m]{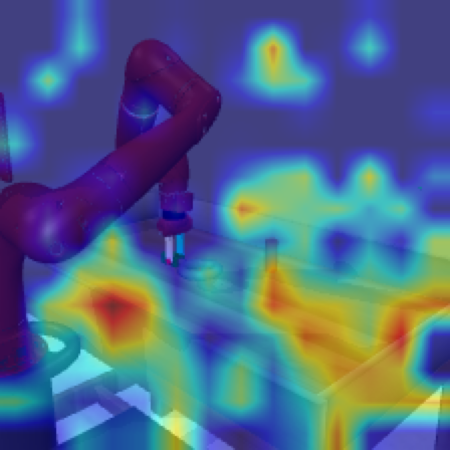} &
    \includegraphics[width=1\linewidth, valign=m]{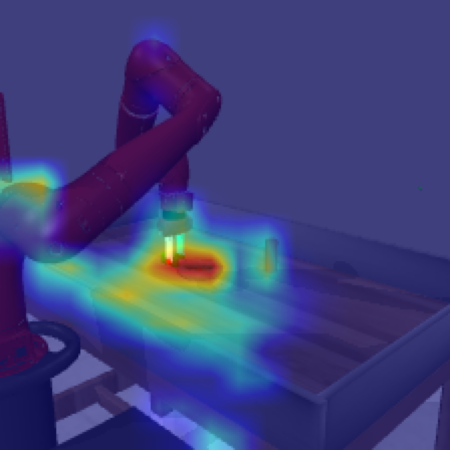} \\ \addlinespace[1ex]

    \rotatebox[origin=c]{90}{\scriptsize Bin-picking} & 
    \includegraphics[width=1\linewidth, valign=m]{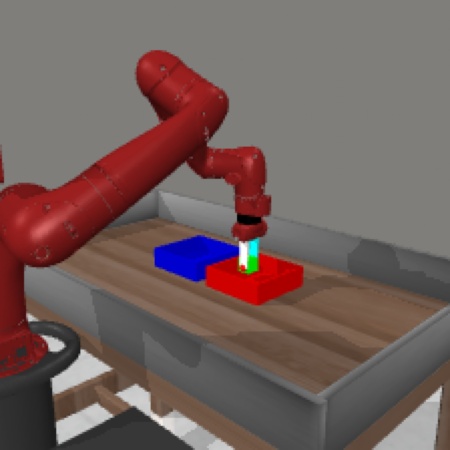} & 
    \includegraphics[width=1\linewidth, valign=m]{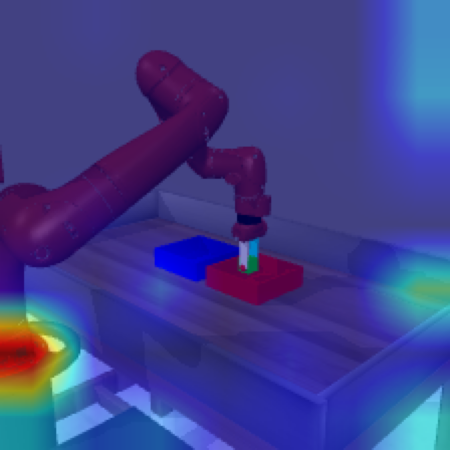} &
    \includegraphics[width=1\linewidth, valign=m]{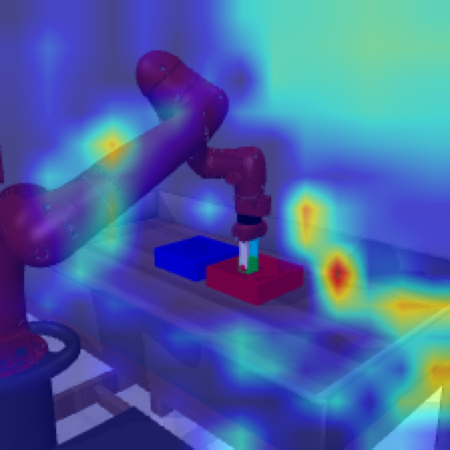} &
    \includegraphics[width=1\linewidth, valign=m]{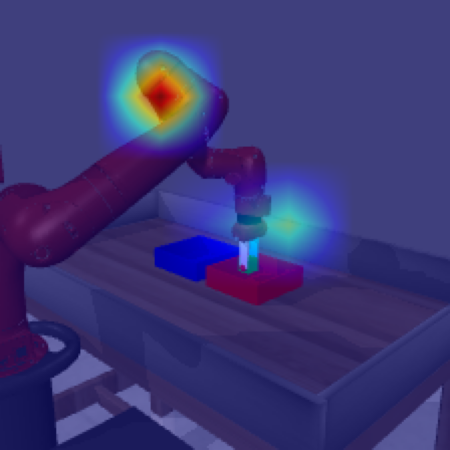} &
    \includegraphics[width=1\linewidth, valign=m]{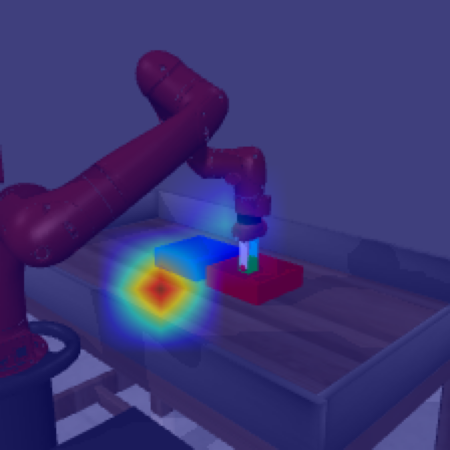} &
    \includegraphics[width=1\linewidth, valign=m]{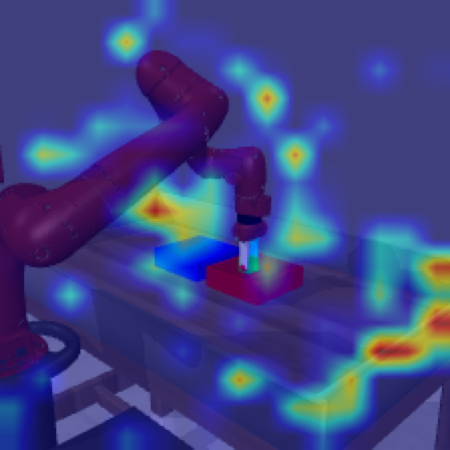} &
    \includegraphics[width=1\linewidth, valign=m]{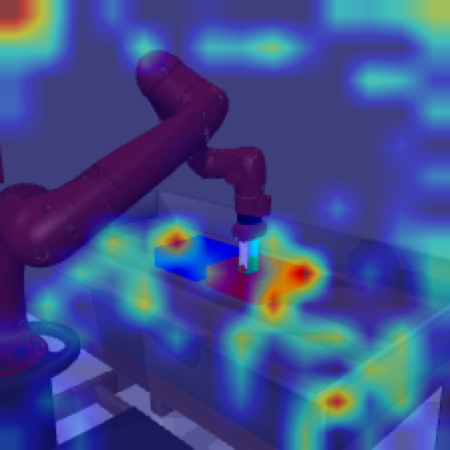} &
    \includegraphics[width=1\linewidth, valign=m]{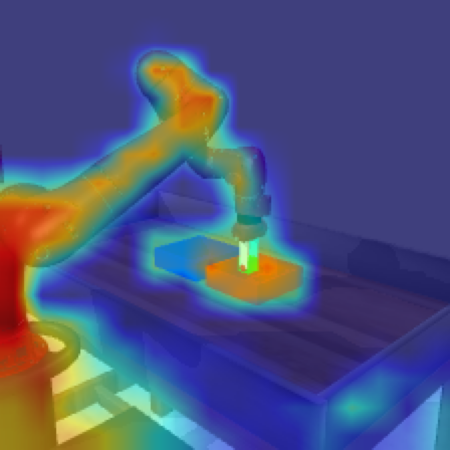} \\ \addlinespace[1ex]

    \rotatebox[origin=c]{90}{\scriptsize Box-close} & 
    \includegraphics[width=1\linewidth, valign=m]{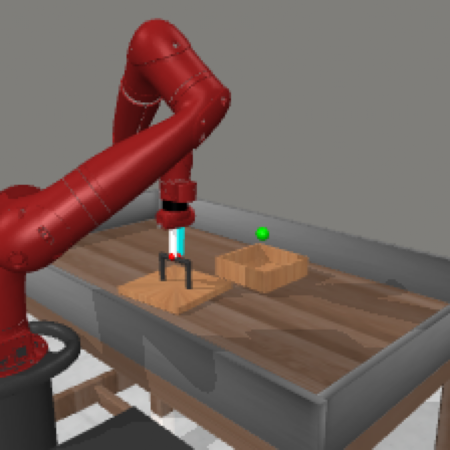} & 
    \includegraphics[width=1\linewidth, valign=m]{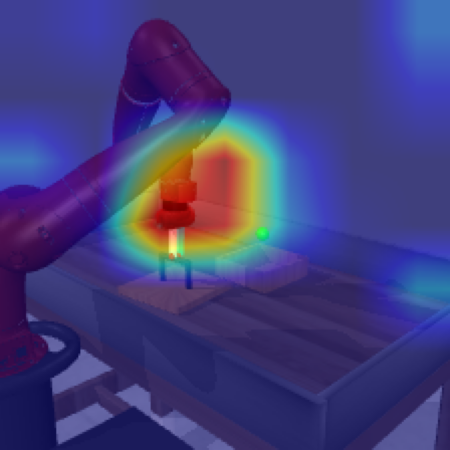} &
    \includegraphics[width=1\linewidth, valign=m]{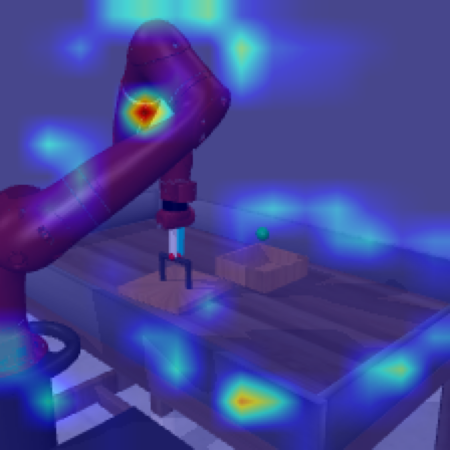} &
    \includegraphics[width=1\linewidth, valign=m]{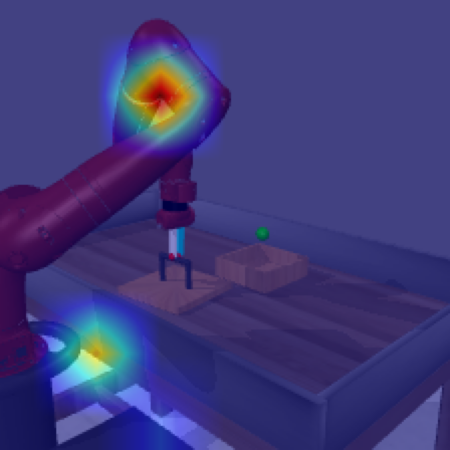} &
    \includegraphics[width=1\linewidth, valign=m]{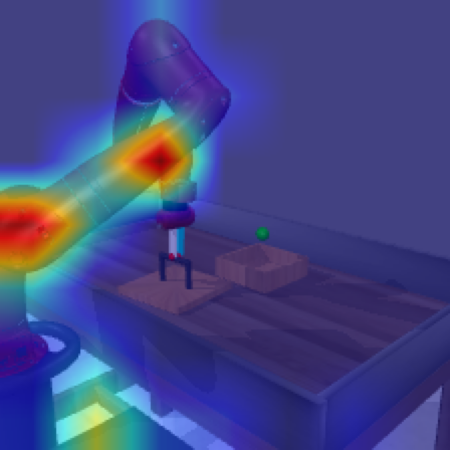} &
    \includegraphics[width=1\linewidth, valign=m]{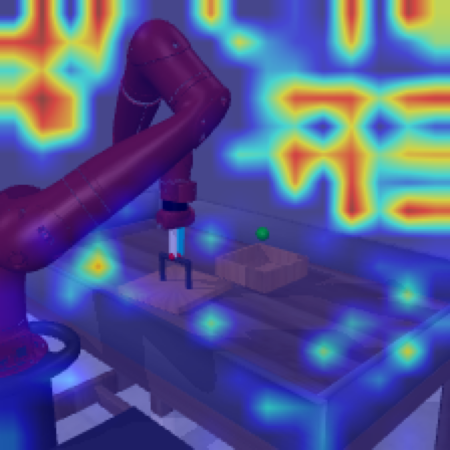} &
    \includegraphics[width=1\linewidth, valign=m]{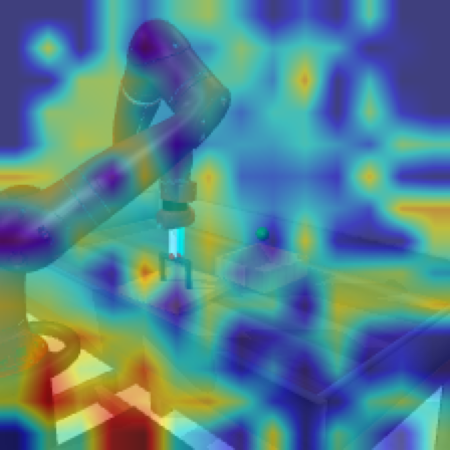} &
    \includegraphics[width=1\linewidth, valign=m]{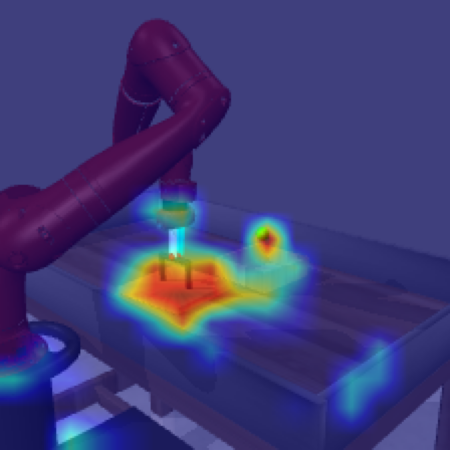} \\ \addlinespace[1ex]

    \rotatebox[origin=c]{90}{\scriptsize Button-press} & 
    \includegraphics[width=1\linewidth, valign=m]{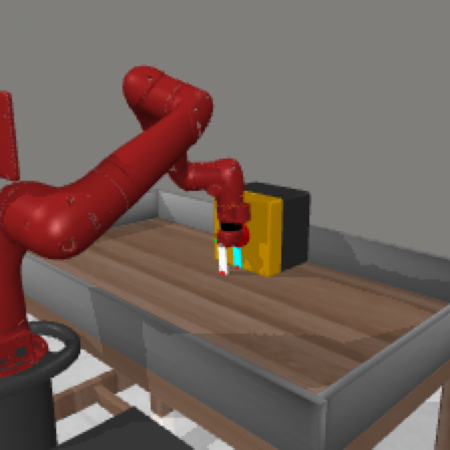} & 
    \includegraphics[width=1\linewidth, valign=m]{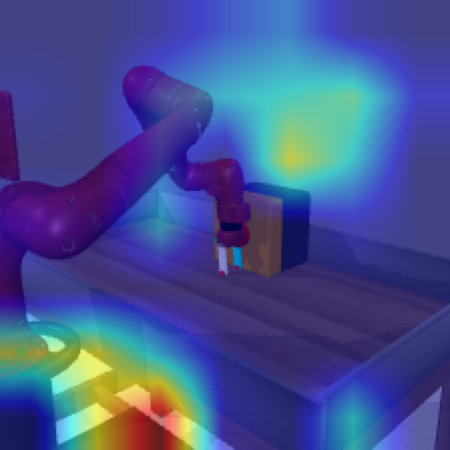} &
    \includegraphics[width=1\linewidth, valign=m]{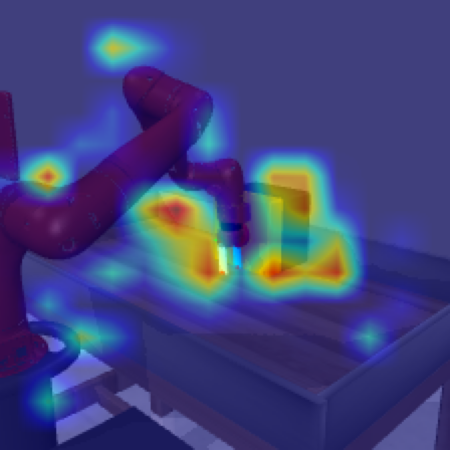} &
    \includegraphics[width=1\linewidth, valign=m]{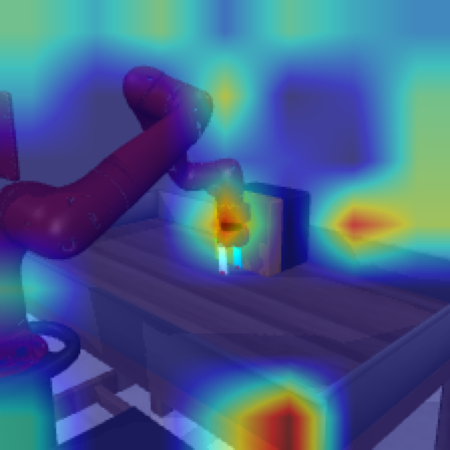} &
    \includegraphics[width=1\linewidth, valign=m]{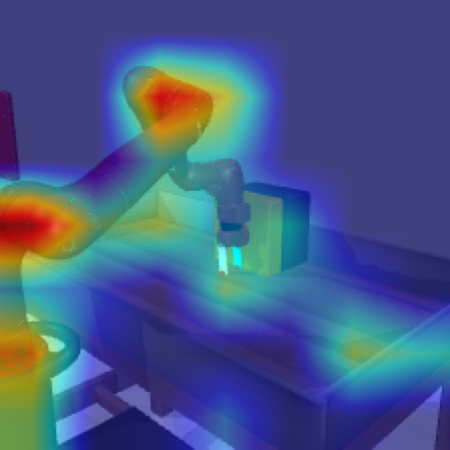} &
    \includegraphics[width=1\linewidth, valign=m]{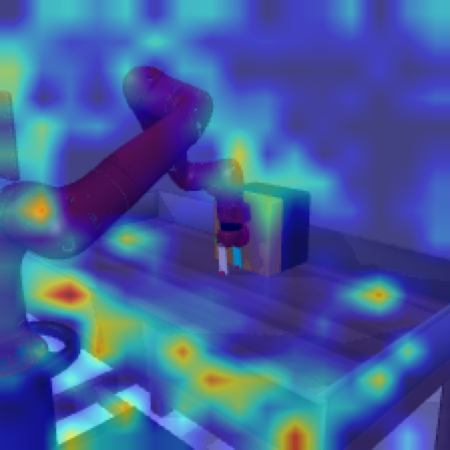} &
    \includegraphics[width=1\linewidth, valign=m]{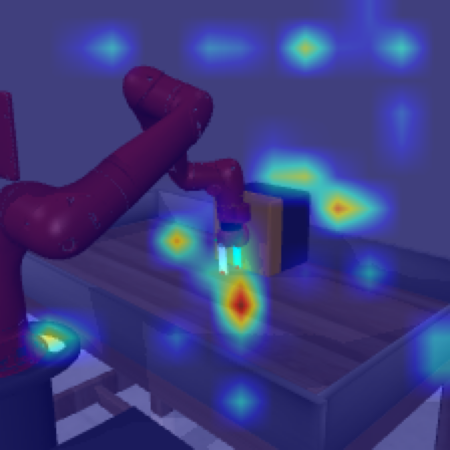} &
    \includegraphics[width=1\linewidth, valign=m]{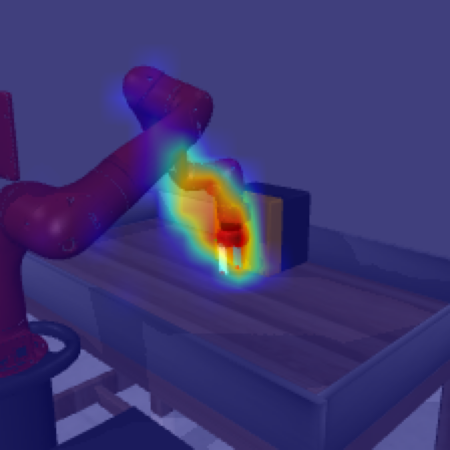} \\ \addlinespace[1ex]

    \rotatebox[origin=c]{90}{\scriptsize Dial-turn} & 
    \includegraphics[width=1\linewidth, valign=m]{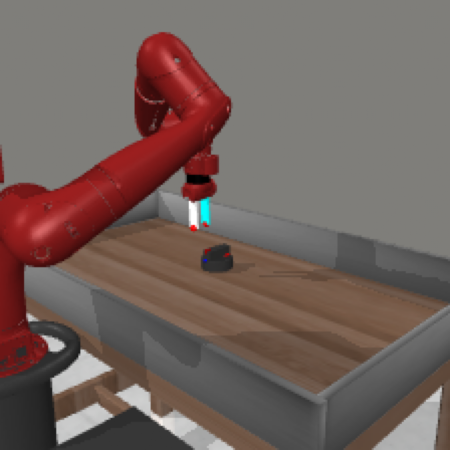} & 
    \includegraphics[width=1\linewidth, valign=m]{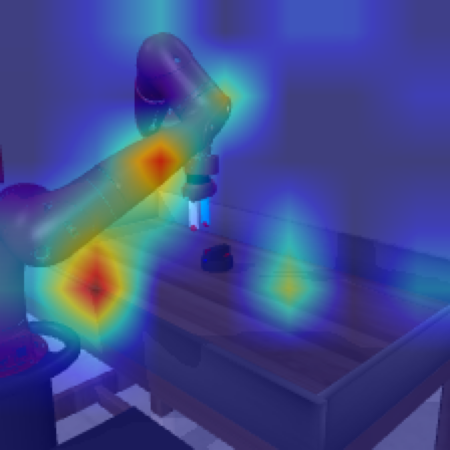} &
    \includegraphics[width=1\linewidth, valign=m]{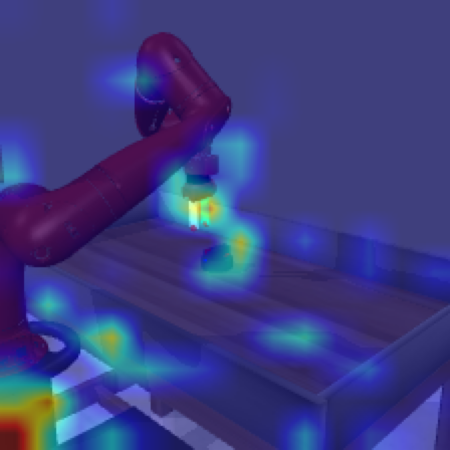} &
    \includegraphics[width=1\linewidth, valign=m]{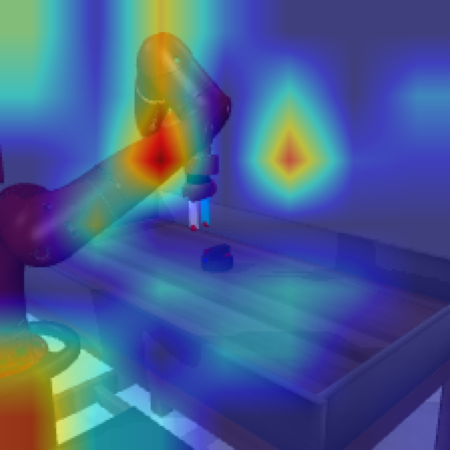} &
    \includegraphics[width=1\linewidth, valign=m]{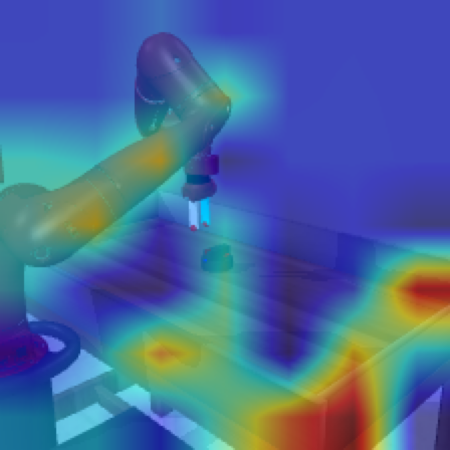} &
    \includegraphics[width=1\linewidth, valign=m]{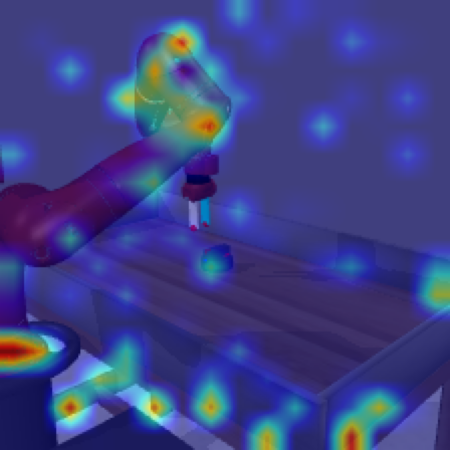} &
    \includegraphics[width=1\linewidth, valign=m]{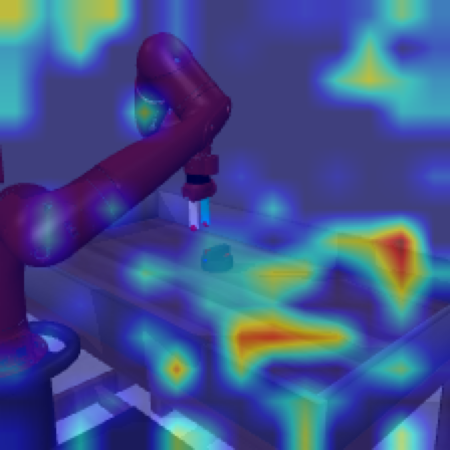} &
    \includegraphics[width=1\linewidth, valign=m]{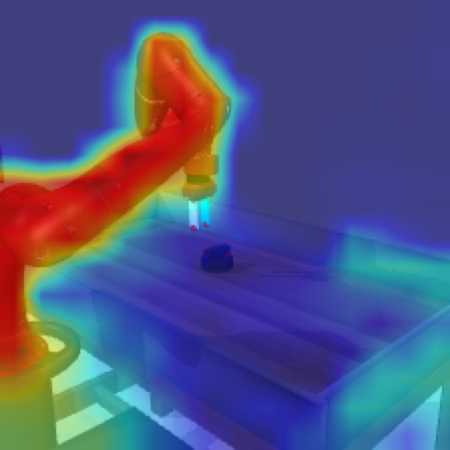} \\ \addlinespace[1ex]

    \rotatebox[origin=c]{90}{\scriptsize Drawer-open} & 
    \includegraphics[width=1\linewidth, valign=m]{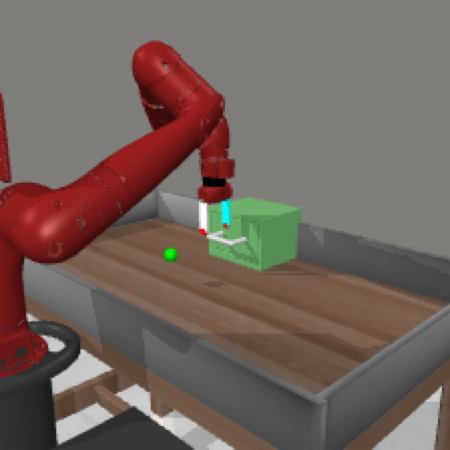} & 
    \includegraphics[width=1\linewidth, valign=m]{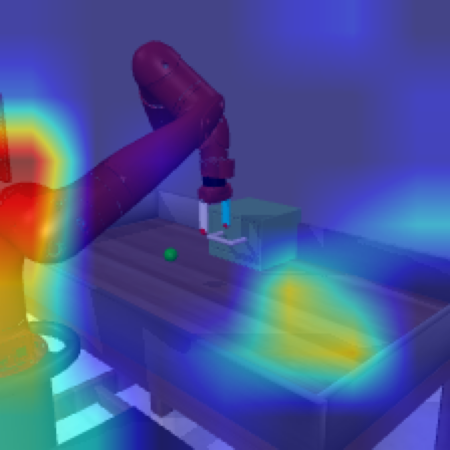} &
    \includegraphics[width=1\linewidth, valign=m]{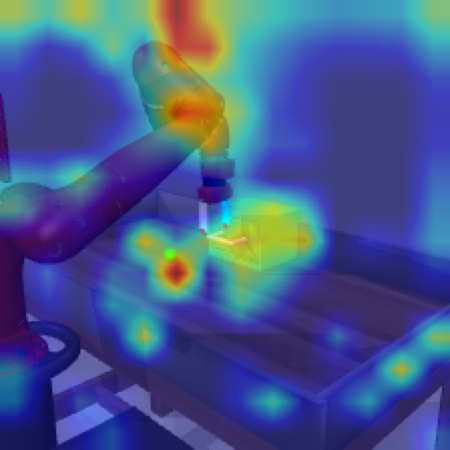} &
    \includegraphics[width=1\linewidth, valign=m]{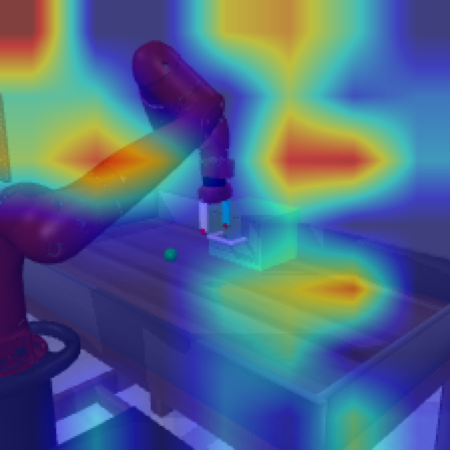} &
    \includegraphics[width=1\linewidth, valign=m]{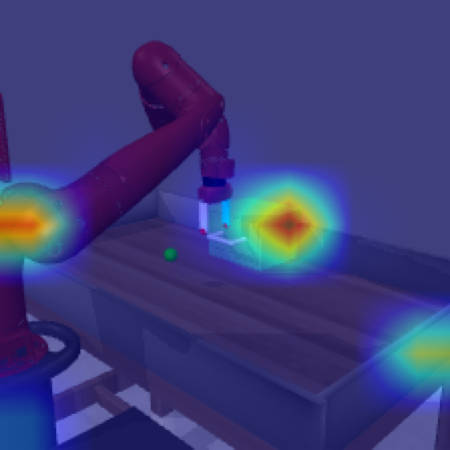} &
    \includegraphics[width=1\linewidth, valign=m]{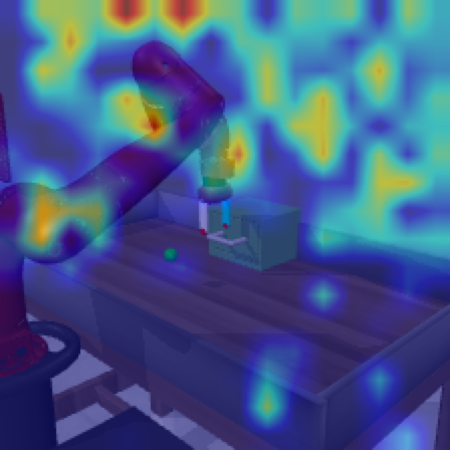} &
    \includegraphics[width=1\linewidth, valign=m]{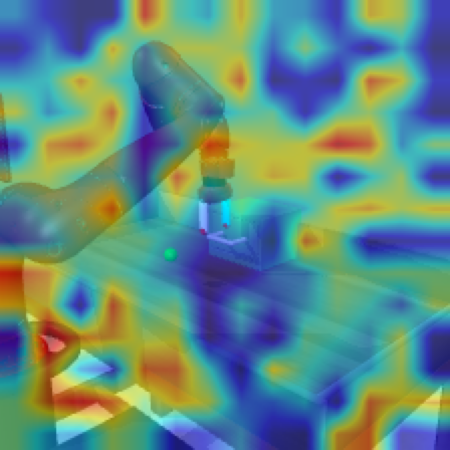} &
    \includegraphics[width=1\linewidth, valign=m]{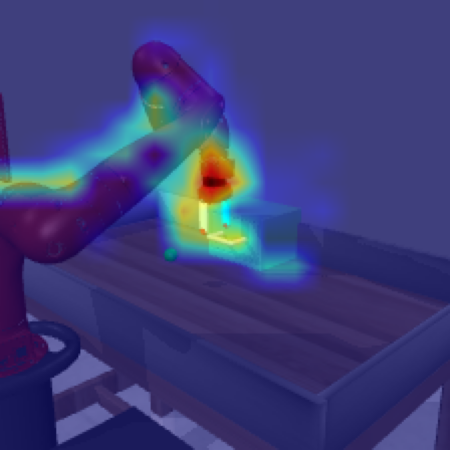} \\ \addlinespace[1ex]

    \rotatebox[origin=c]{90}{\scriptsize Hammer} & 
    \includegraphics[width=1\linewidth, valign=m]{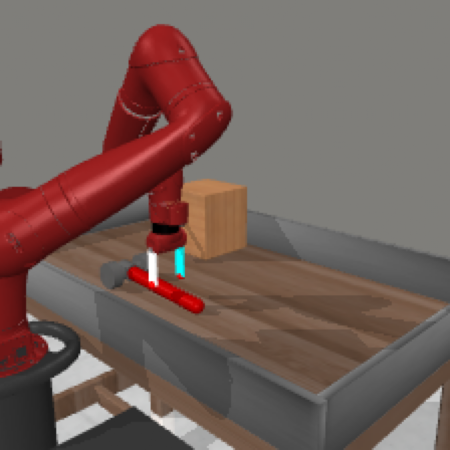} & 
    \includegraphics[width=1\linewidth, valign=m]{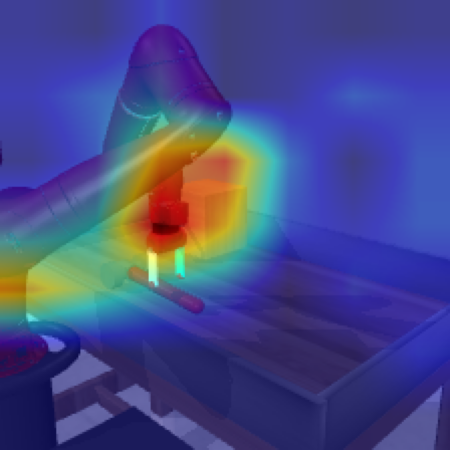} &
    \includegraphics[width=1\linewidth, valign=m]{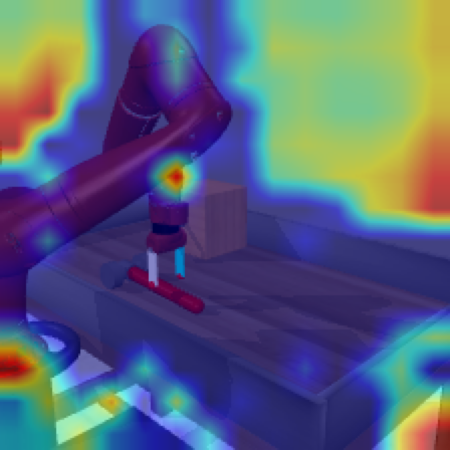} &
    \includegraphics[width=1\linewidth, valign=m]{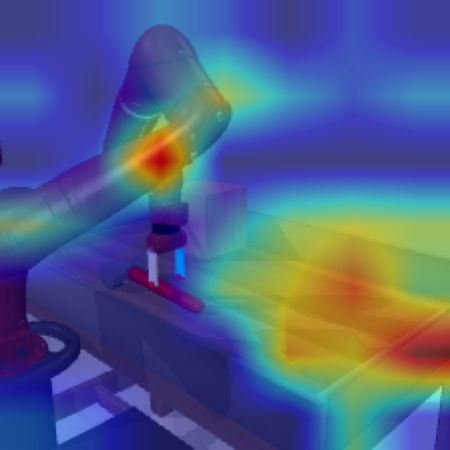} &
    \includegraphics[width=1\linewidth, valign=m]{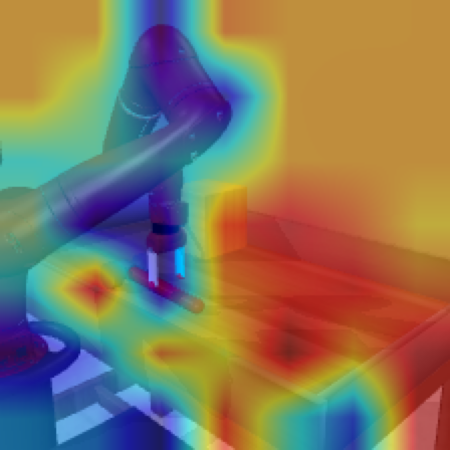} &
    \includegraphics[width=1\linewidth, valign=m]{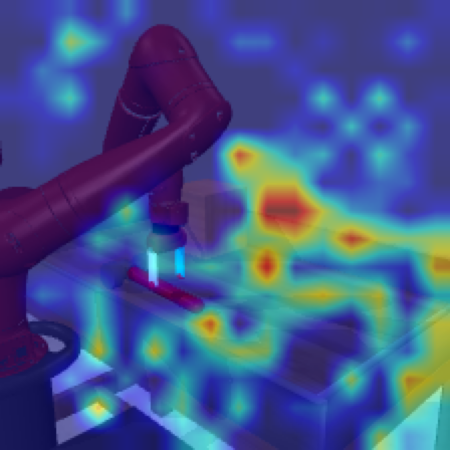} &
    \includegraphics[width=1\linewidth, valign=m]{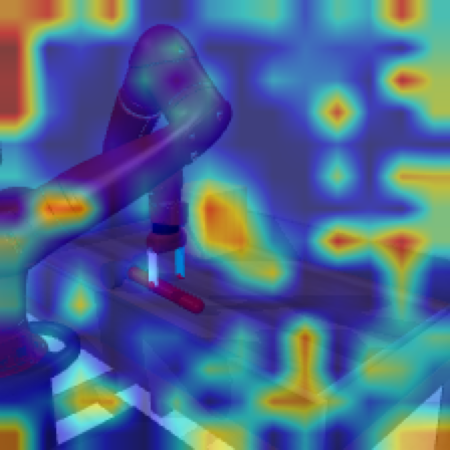} &
    \includegraphics[width=1\linewidth, valign=m]{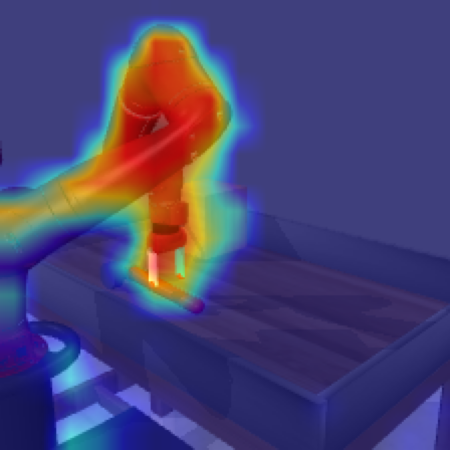} \\ \addlinespace[1ex]

    \rotatebox[origin=c]{90}{\scriptsize Hand-insert} & 
    \includegraphics[width=1\linewidth, valign=m]{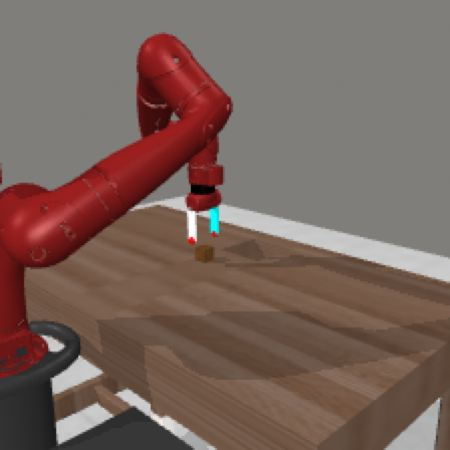} & 
    \includegraphics[width=1\linewidth, valign=m]{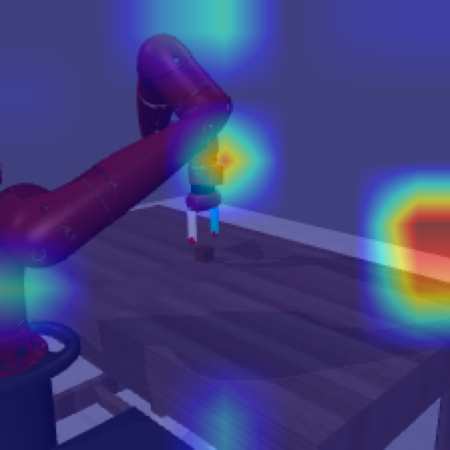} &
    \includegraphics[width=1\linewidth, valign=m]{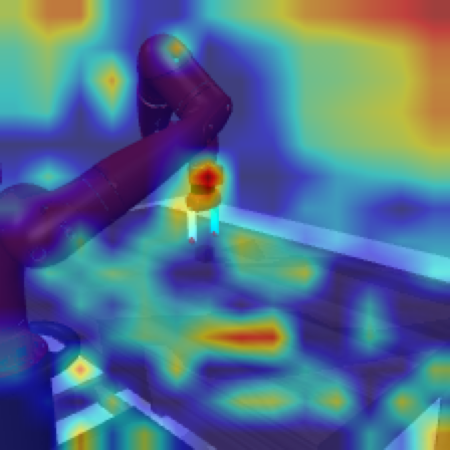} &
    \includegraphics[width=1\linewidth, valign=m]{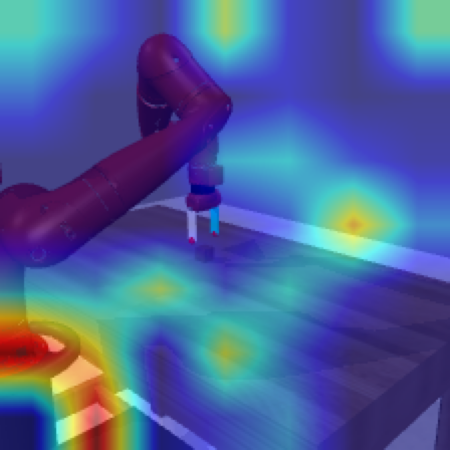} &
    \includegraphics[width=1\linewidth, valign=m]{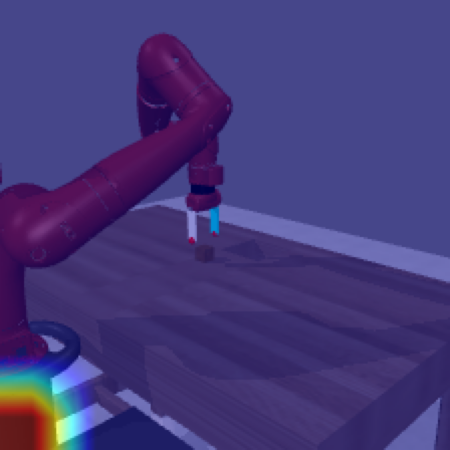} &
    \includegraphics[width=1\linewidth, valign=m]{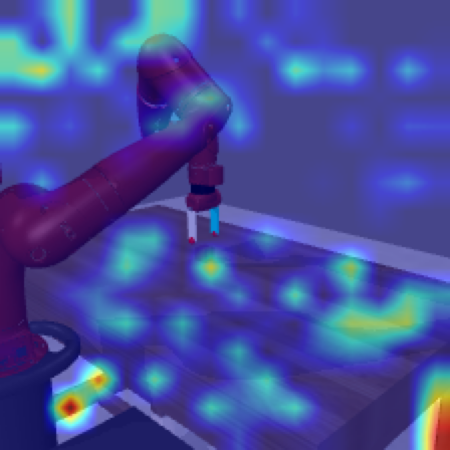} &
    \includegraphics[width=1\linewidth, valign=m]{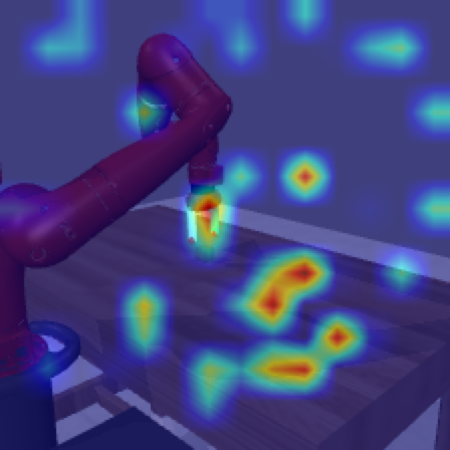} &
    \includegraphics[width=1\linewidth, valign=m]{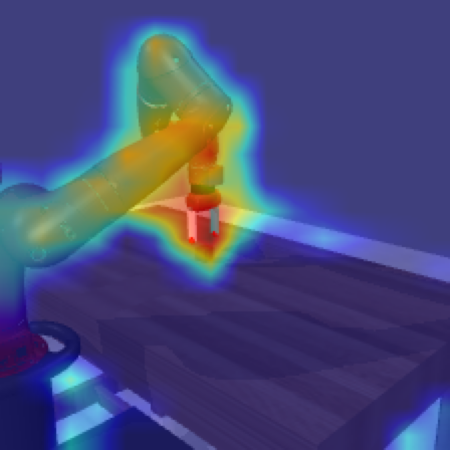} \\ \addlinespace[1ex]

    \rotatebox[origin=c]{90}{\scriptsize Handle-pull} & 
    \includegraphics[width=1\linewidth, valign=m]{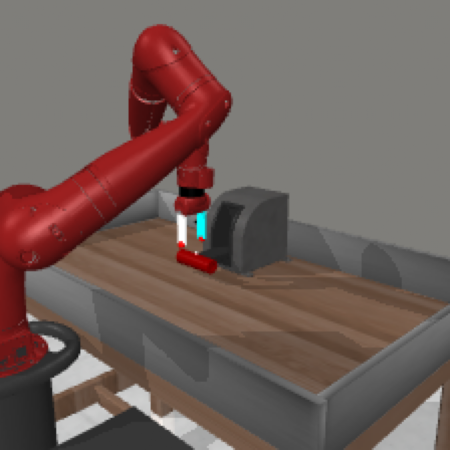} & 
    \includegraphics[width=1\linewidth, valign=m]{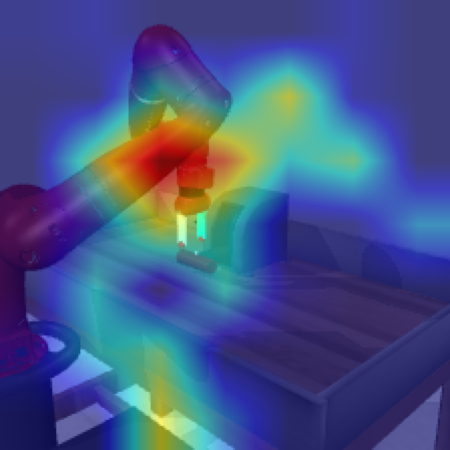} &
    \includegraphics[width=1\linewidth, valign=m]{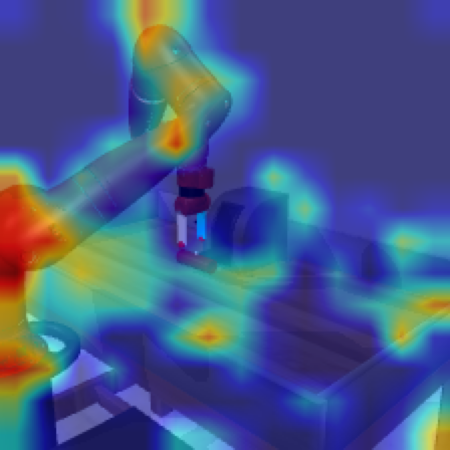} &
    \includegraphics[width=1\linewidth, valign=m]{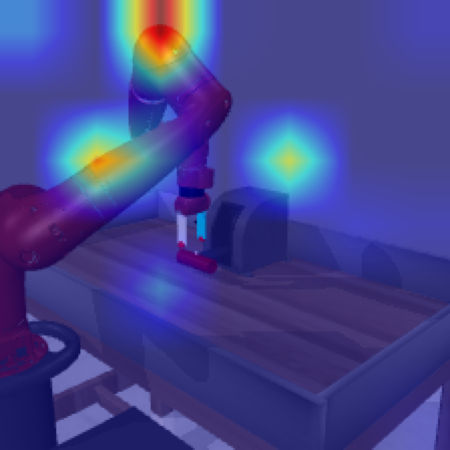} &
    \includegraphics[width=1\linewidth, valign=m]{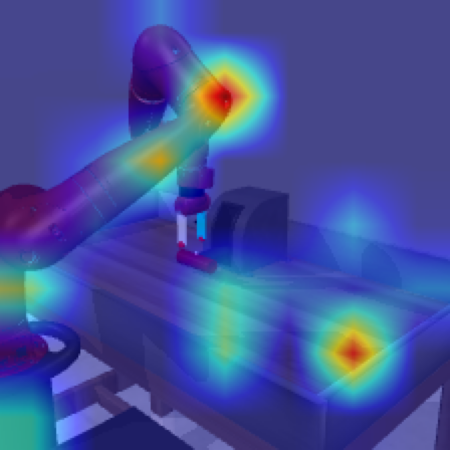} &
    \includegraphics[width=1\linewidth, valign=m]{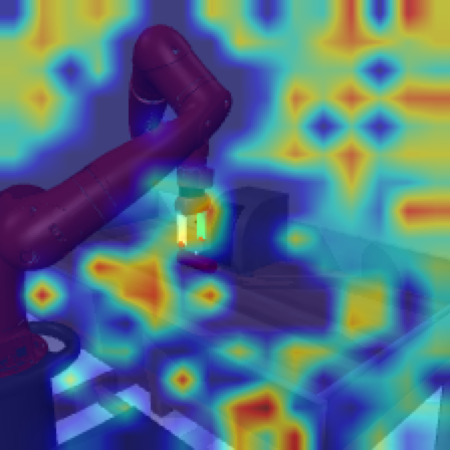} &
    \includegraphics[width=1\linewidth, valign=m]{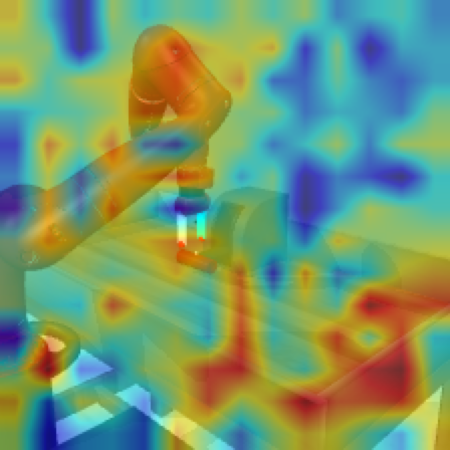} &
    \includegraphics[width=1\linewidth, valign=m]{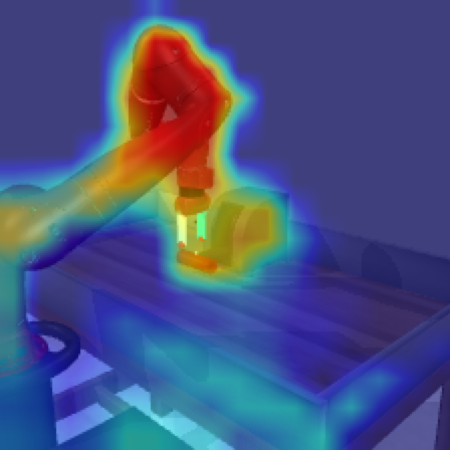} \\ \addlinespace[1ex]

    \rotatebox[origin=c]{90}{\scriptsize Lever-pull} & 
    \includegraphics[width=1\linewidth, valign=m]{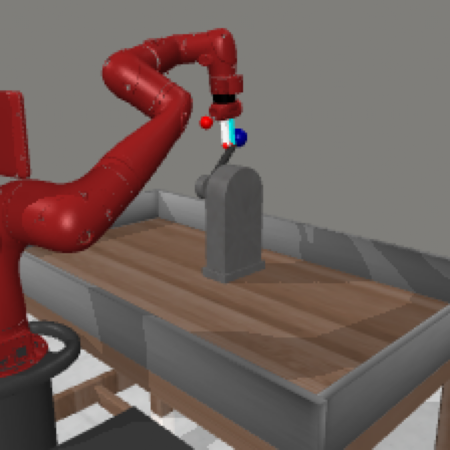} & 
    \includegraphics[width=1\linewidth, valign=m]{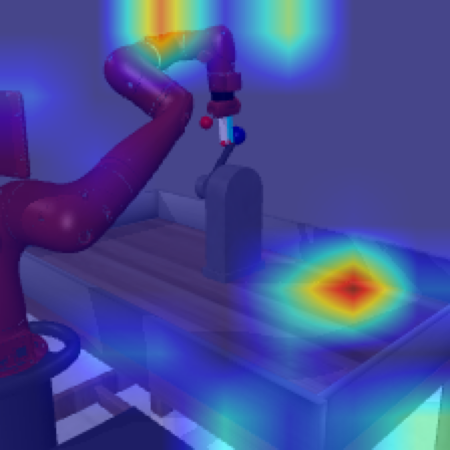} &
    \includegraphics[width=1\linewidth, valign=m]{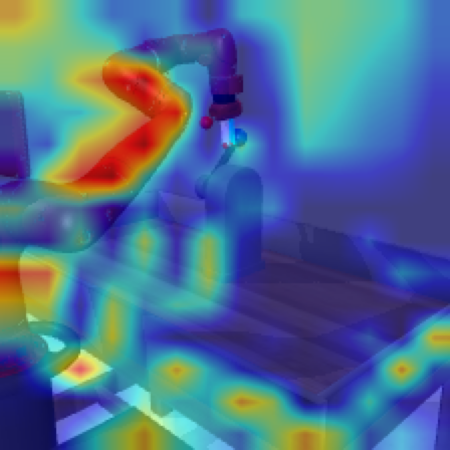} &
    \includegraphics[width=1\linewidth, valign=m]{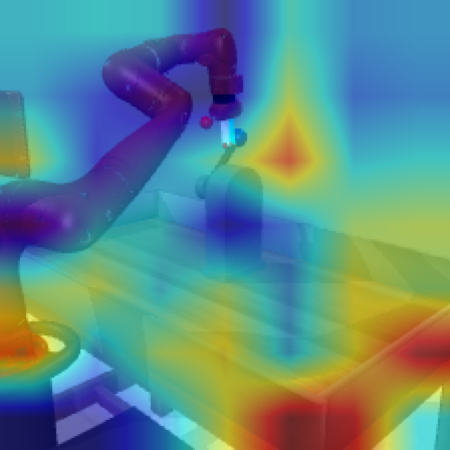} &
    \includegraphics[width=1\linewidth, valign=m]{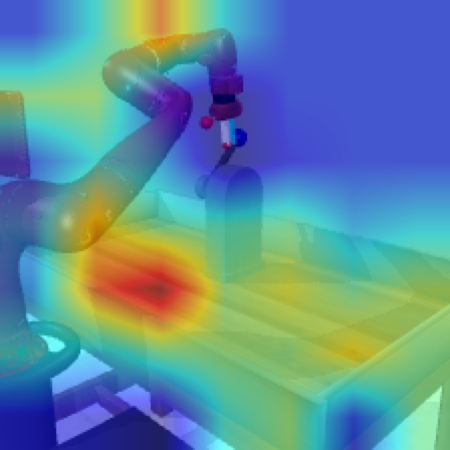} &
    \includegraphics[width=1\linewidth, valign=m]{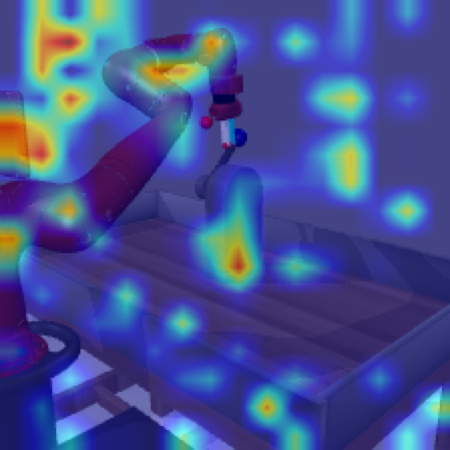} &
    \includegraphics[width=1\linewidth, valign=m]{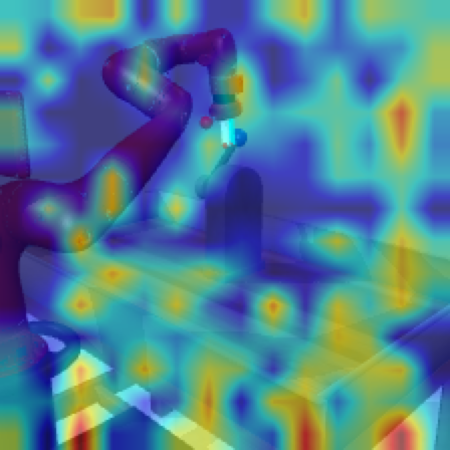} &
    \includegraphics[width=1\linewidth, valign=m]{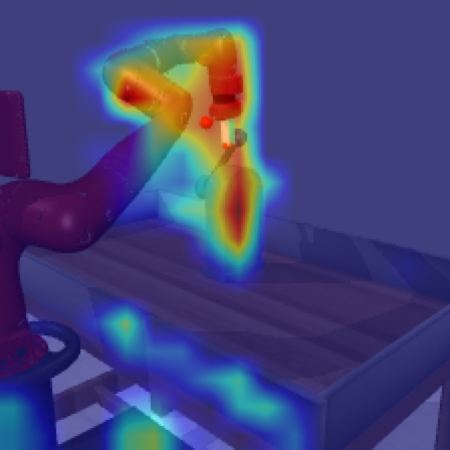} \\ \addlinespace[1ex]

    \rotatebox[origin=c]{90}{\scriptsize Peg-unplug-side} & 
    \includegraphics[width=1\linewidth, valign=m]{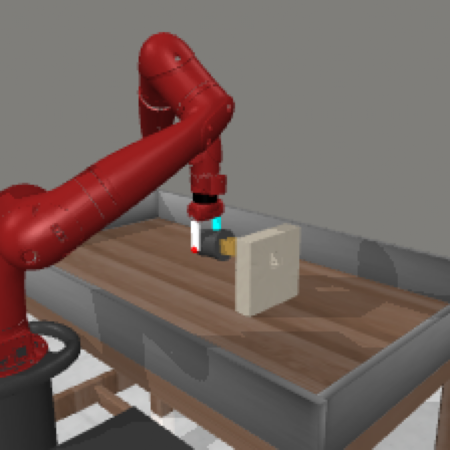} & 
    \includegraphics[width=1\linewidth, valign=m]{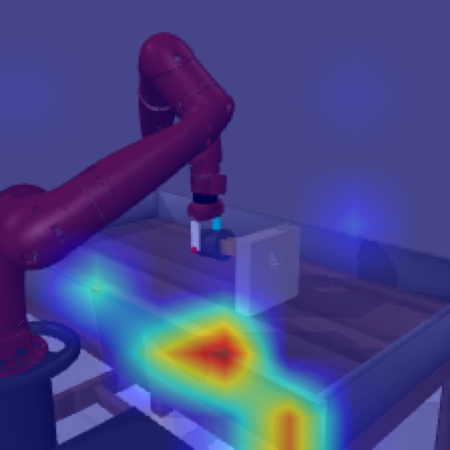} &
    \includegraphics[width=1\linewidth, valign=m]{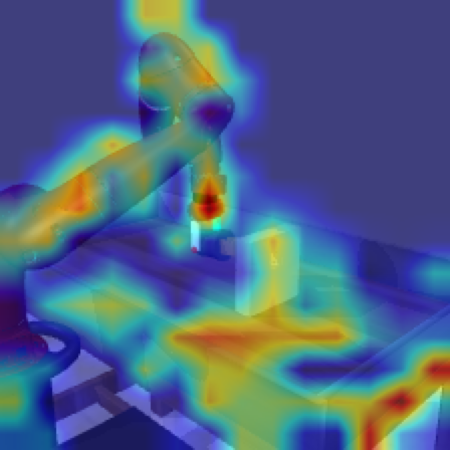} &
    \includegraphics[width=1\linewidth, valign=m]{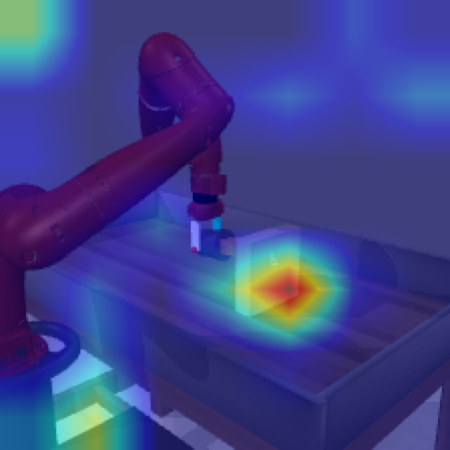} &
    \includegraphics[width=1\linewidth, valign=m]{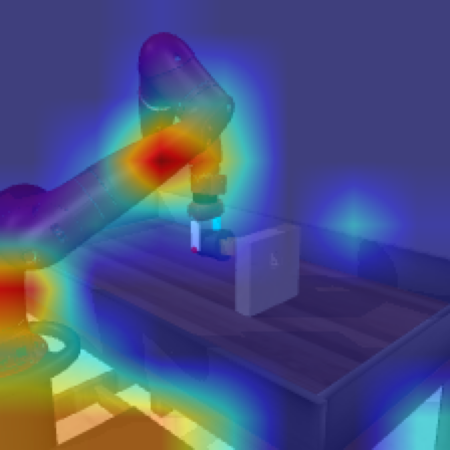} &
    \includegraphics[width=1\linewidth, valign=m]{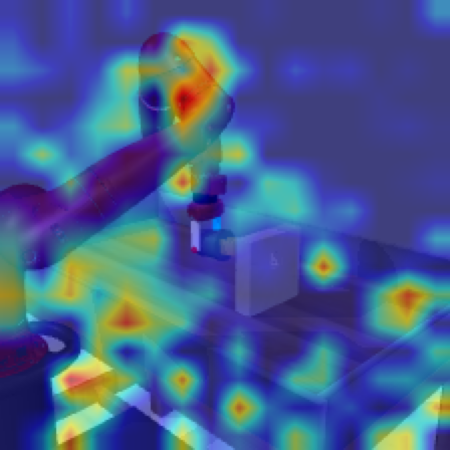} &
    \includegraphics[width=1\linewidth, valign=m]{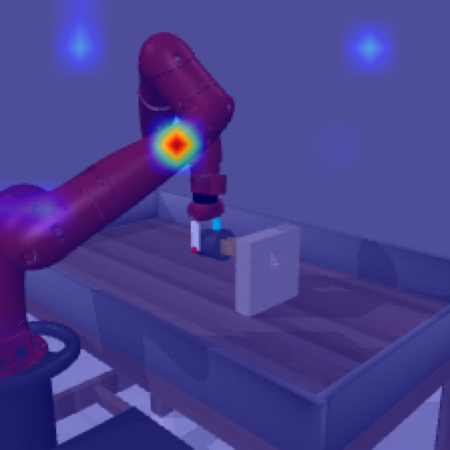} &
    \includegraphics[width=1\linewidth, valign=m]{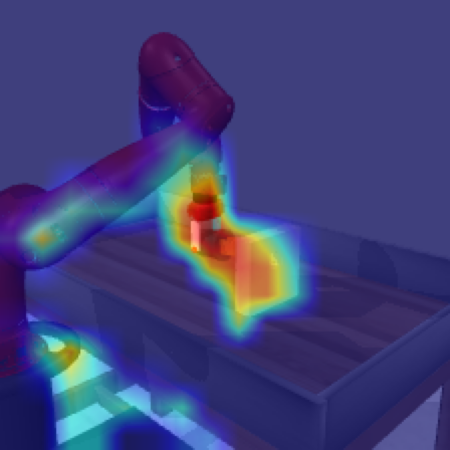} \\ \addlinespace[1ex]

    \rotatebox[origin=c]{90}{\scriptsize Push-wall} & 
    \includegraphics[width=1\linewidth, valign=m]{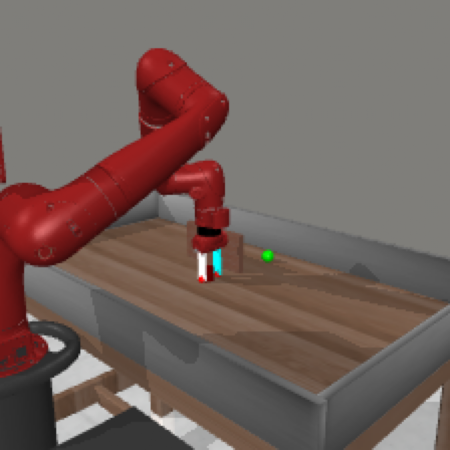} & 
    \includegraphics[width=1\linewidth, valign=m]{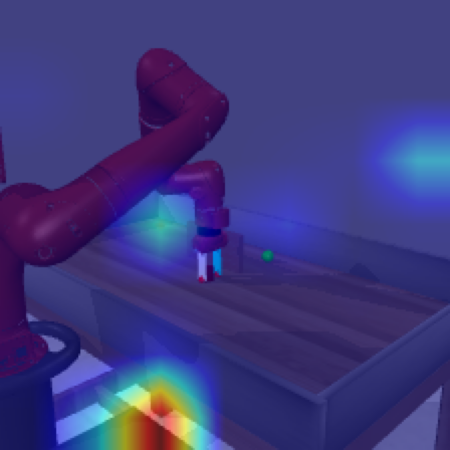} &
    \includegraphics[width=1\linewidth, valign=m]{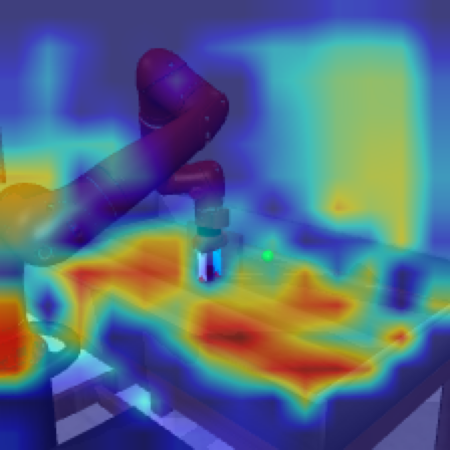} &
    \includegraphics[width=1\linewidth, valign=m]{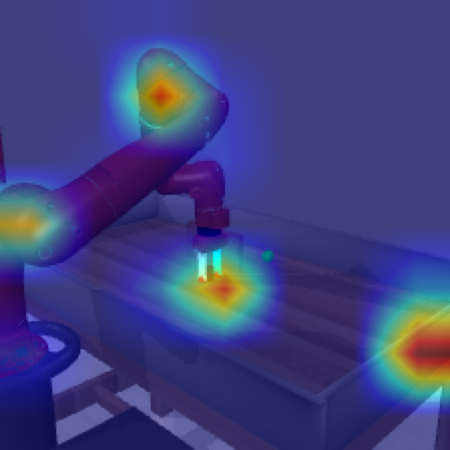} &
    \includegraphics[width=1\linewidth, valign=m]{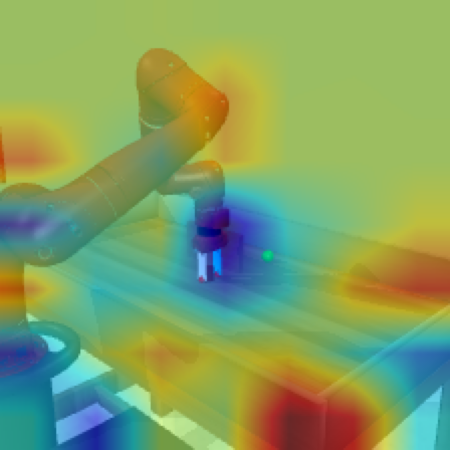} &
    \includegraphics[width=1\linewidth, valign=m]{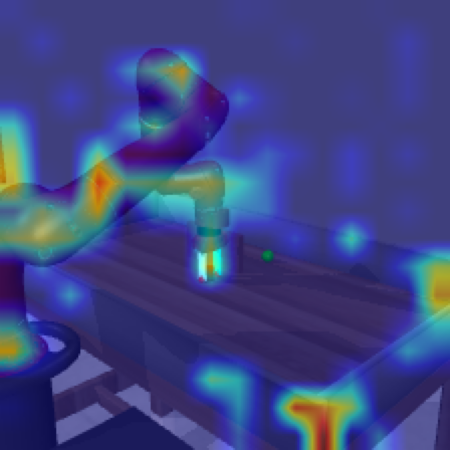} &
    \includegraphics[width=1\linewidth, valign=m]{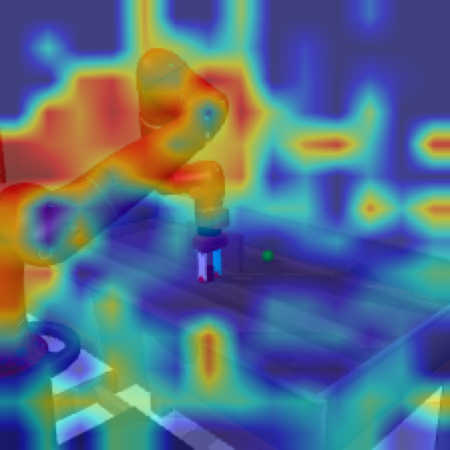} &
    \includegraphics[width=1\linewidth, valign=m]{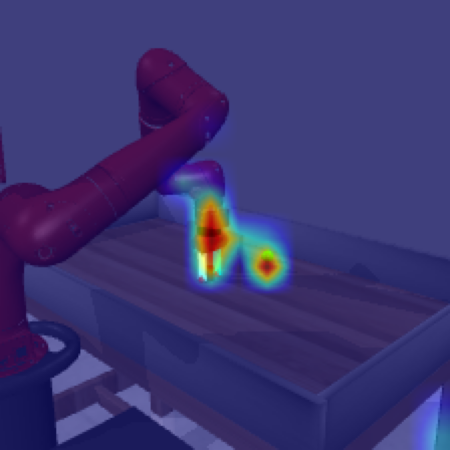} \\ \addlinespace[1ex]

    \rotatebox[origin=c]{90}{\scriptsize Reach} & 
    \includegraphics[width=1\linewidth, valign=m]{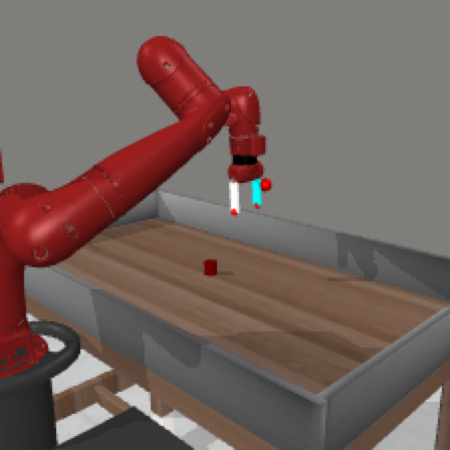} & 
    \includegraphics[width=1\linewidth, valign=m]{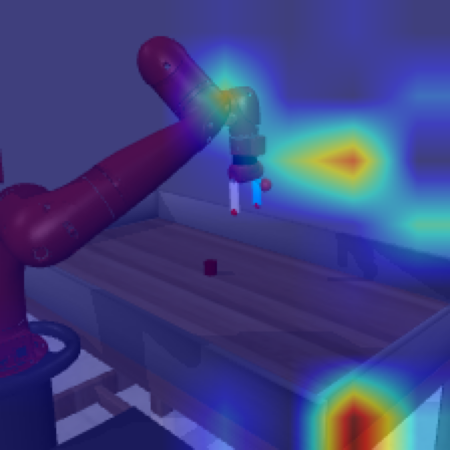} &
    \includegraphics[width=1\linewidth, valign=m]{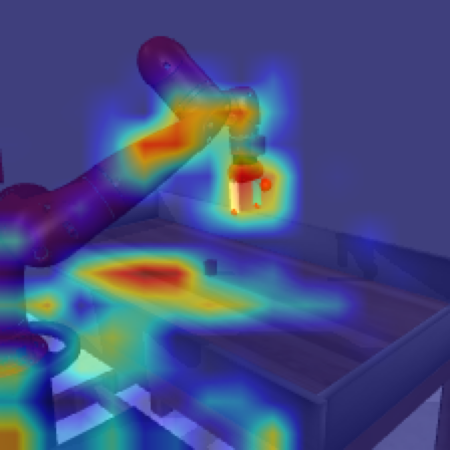} &
    \includegraphics[width=1\linewidth, valign=m]{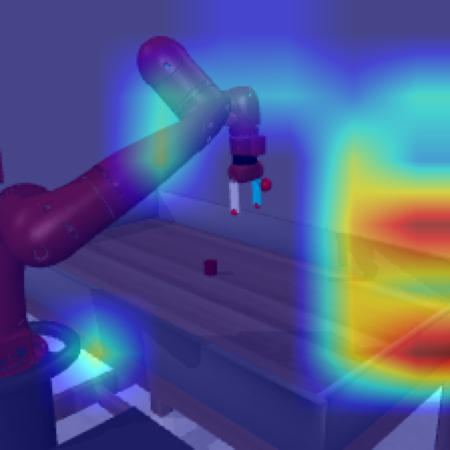} &
    \includegraphics[width=1\linewidth, valign=m]{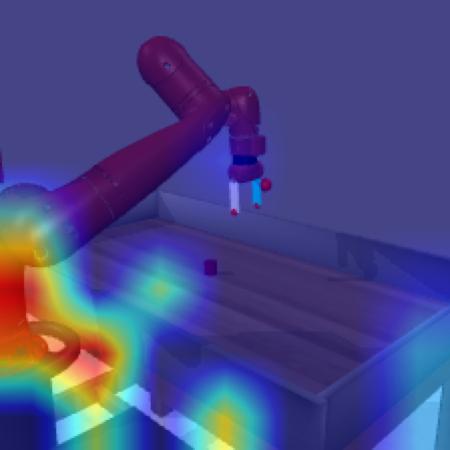} &
    \includegraphics[width=1\linewidth, valign=m]{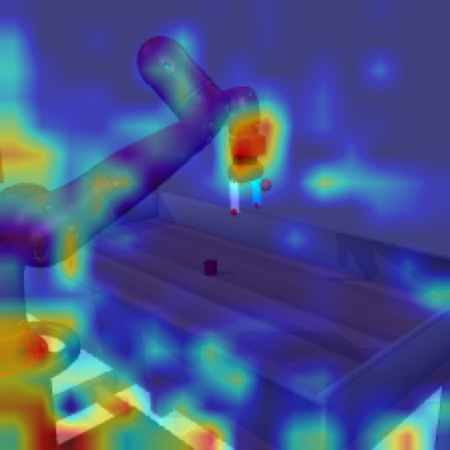} &
    \includegraphics[width=1\linewidth, valign=m]{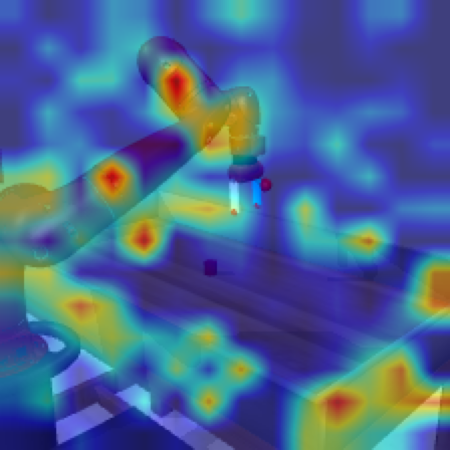} &
    \includegraphics[width=1\linewidth, valign=m]{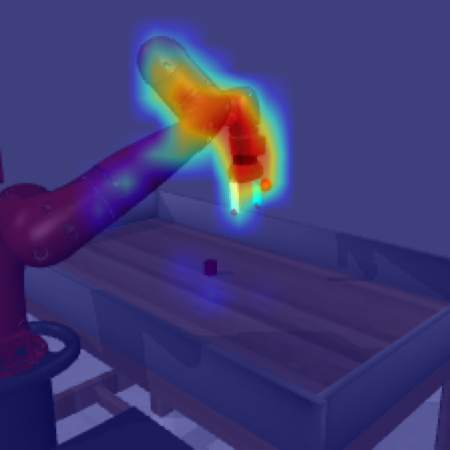} \\ \addlinespace[1ex]

    \rotatebox[origin=c]{90}{\scriptsize Shelf-place} & 
    \includegraphics[width=1\linewidth, valign=m]{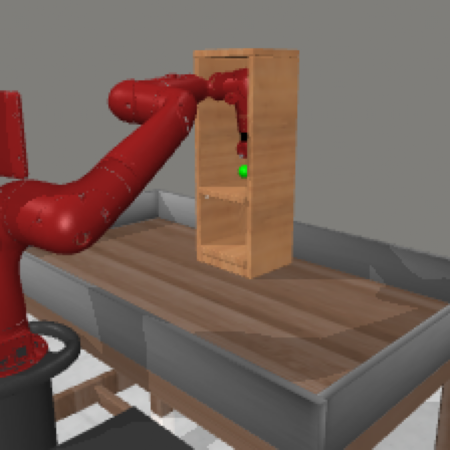} & 
    \includegraphics[width=1\linewidth, valign=m]{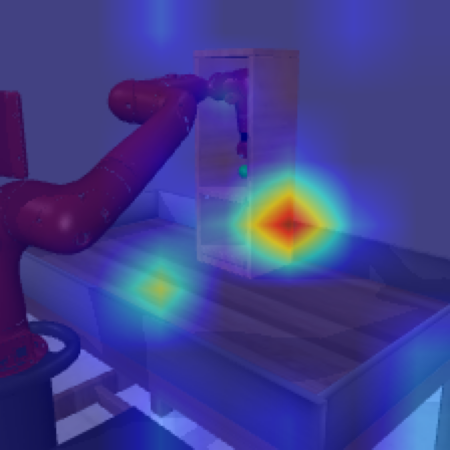} &
    \includegraphics[width=1\linewidth, valign=m]{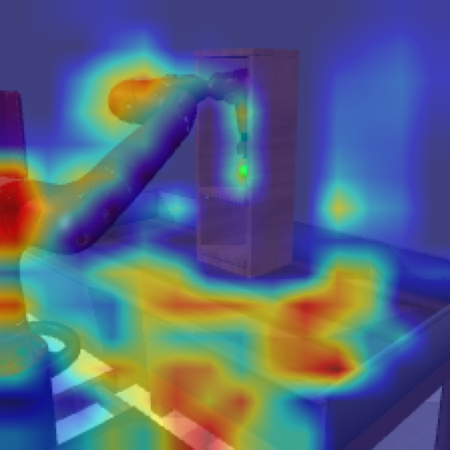} &
    \includegraphics[width=1\linewidth, valign=m]{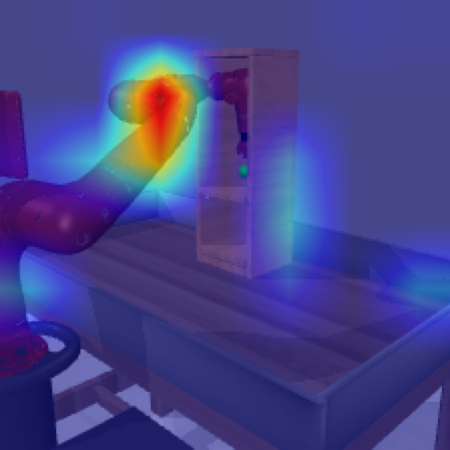} &
    \includegraphics[width=1\linewidth, valign=m]{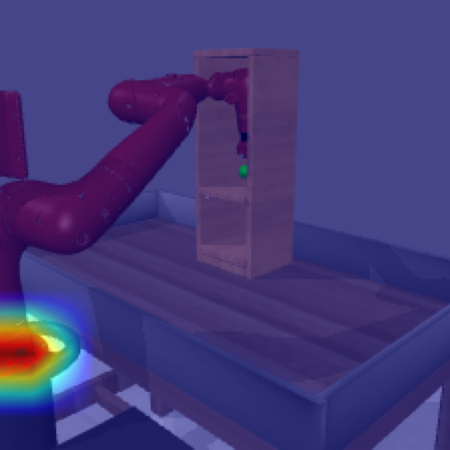} &
    \includegraphics[width=1\linewidth, valign=m]{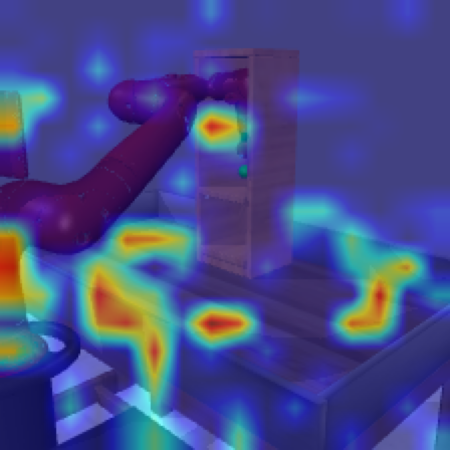} &
    \includegraphics[width=1\linewidth, valign=m]{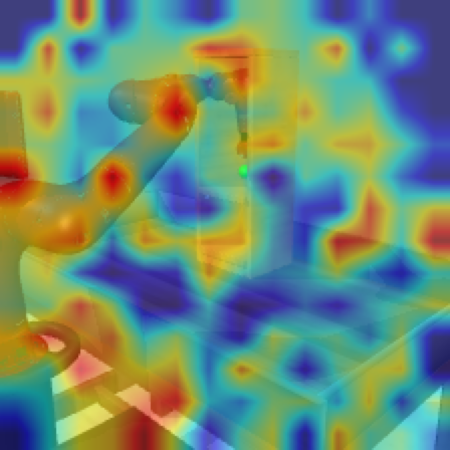} &
    \includegraphics[width=1\linewidth, valign=m]{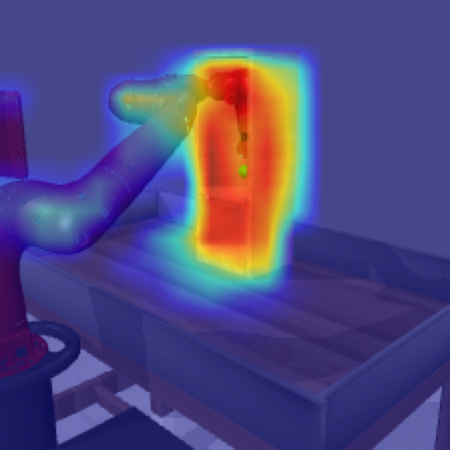} \\ \addlinespace[1ex]

    \rotatebox[origin=c]{90}{\scriptsize Sweep-into} & 
    \includegraphics[width=1\linewidth, valign=m]{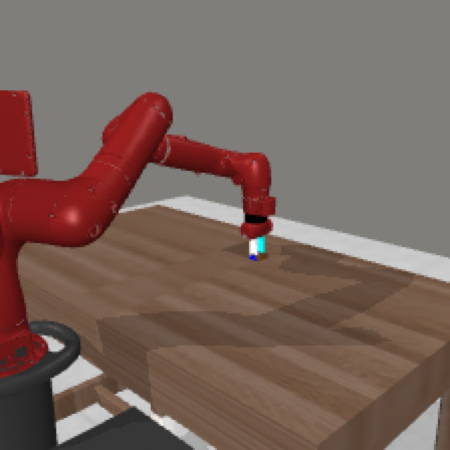} & 
    \includegraphics[width=1\linewidth, valign=m]{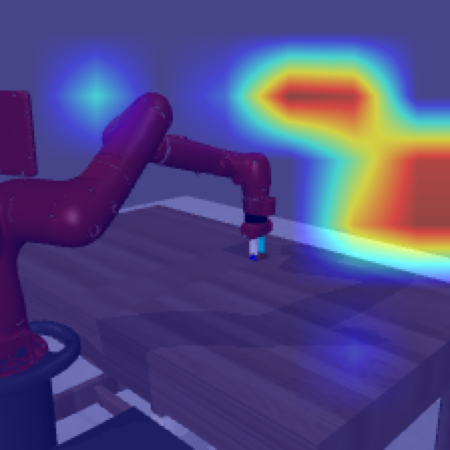} &
    \includegraphics[width=1\linewidth, valign=m]{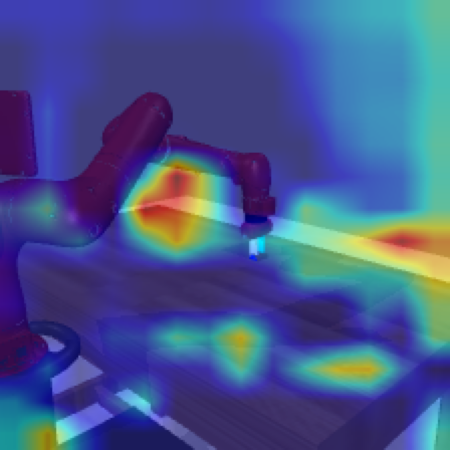} &
    \includegraphics[width=1\linewidth, valign=m]{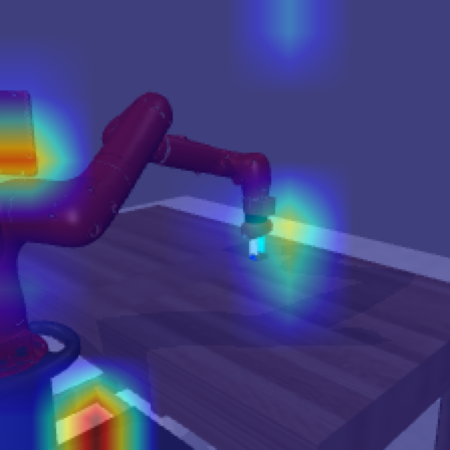} &
    \includegraphics[width=1\linewidth, valign=m]{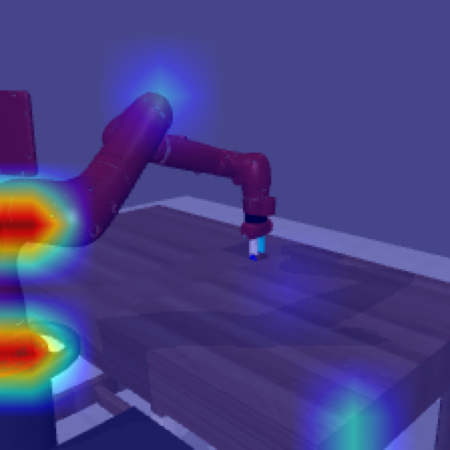} &
    \includegraphics[width=1\linewidth, valign=m]{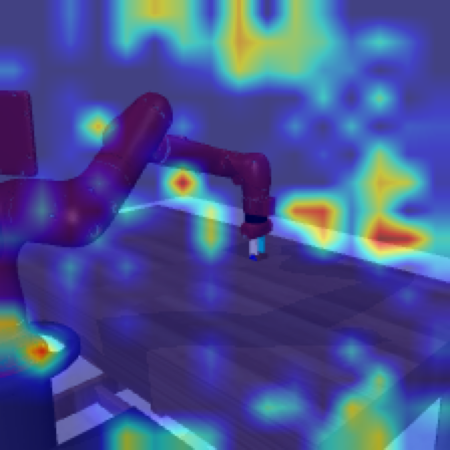} &
    \includegraphics[width=1\linewidth, valign=m]{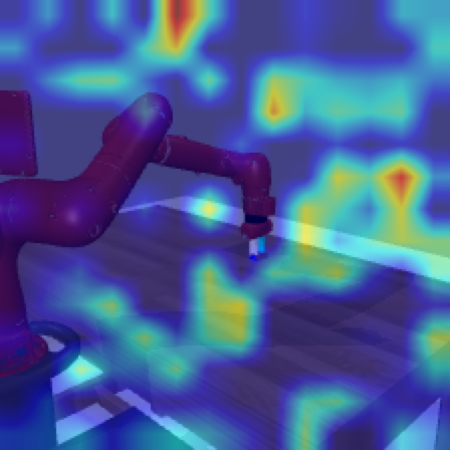} &
    \includegraphics[width=1\linewidth, valign=m]{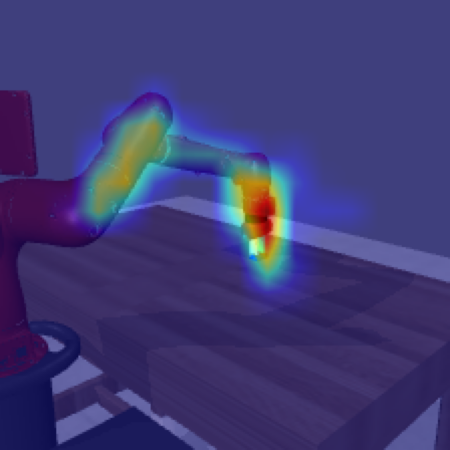} \\ \addlinespace[1ex]

    \rotatebox[origin=c]{90}{\scriptsize Close-box} & 
    \includegraphics[width=1\linewidth, valign=m]{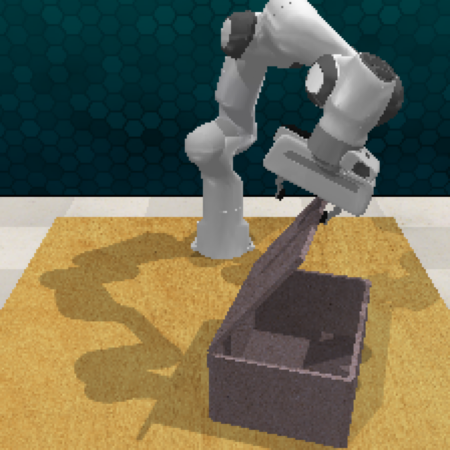} & 
    \includegraphics[width=1\linewidth, valign=m]{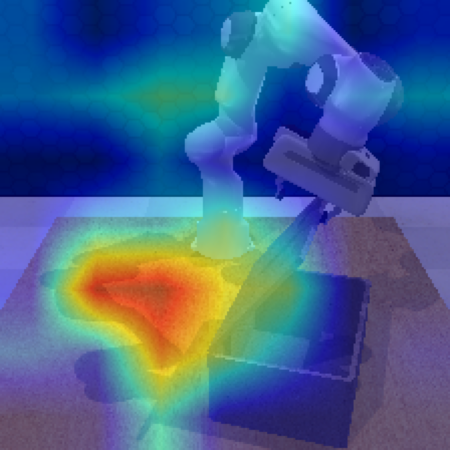} &
    \includegraphics[width=1\linewidth, valign=m]{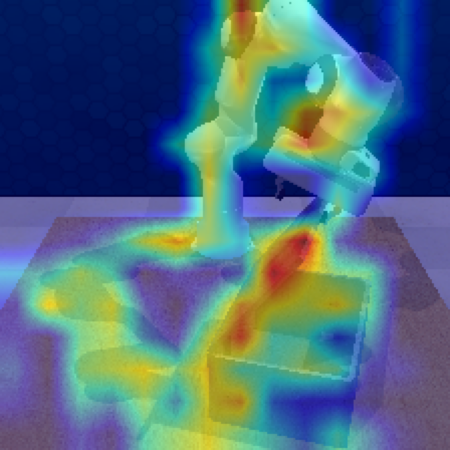} &
    \includegraphics[width=1\linewidth, valign=m]{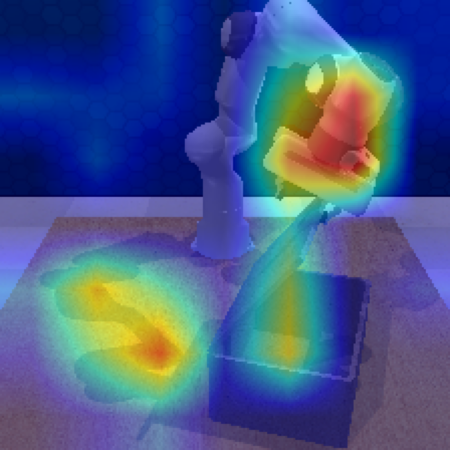} &
    \includegraphics[width=1\linewidth, valign=m]{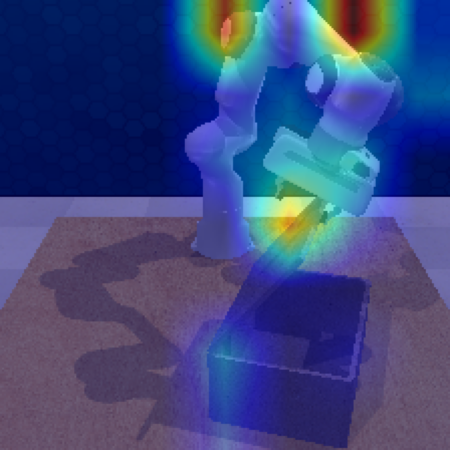} &
    \includegraphics[width=1\linewidth, valign=m]{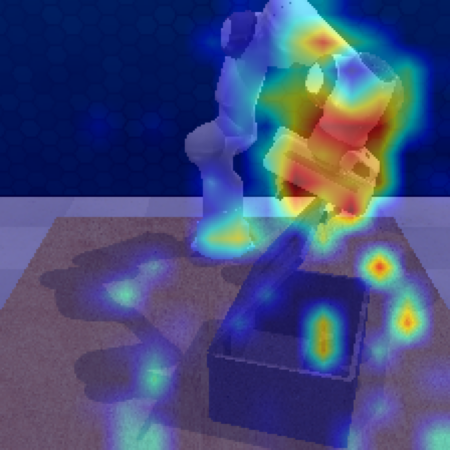} &
    \includegraphics[width=1\linewidth, valign=m]{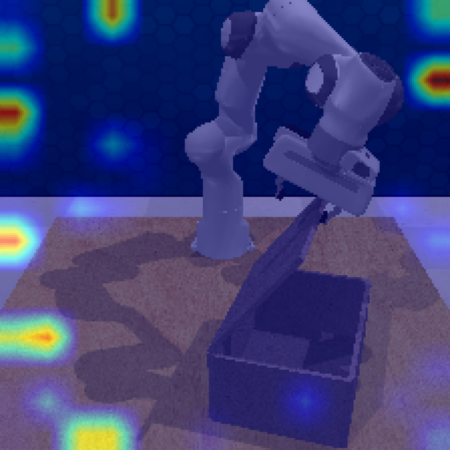} &
    \includegraphics[width=1\linewidth, valign=m]{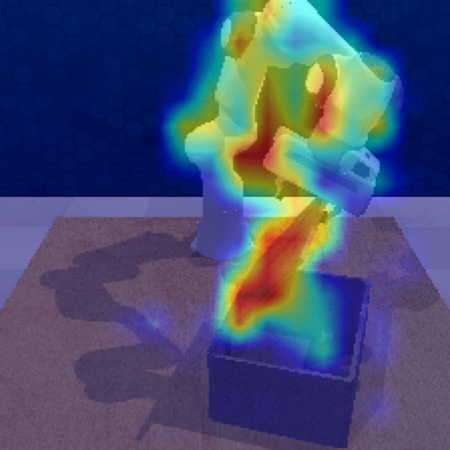} \\ \addlinespace[1ex]

    \rotatebox[origin=c]{90}{\scriptsize Close-laptop-lid} & 
    \includegraphics[width=1\linewidth, valign=m]{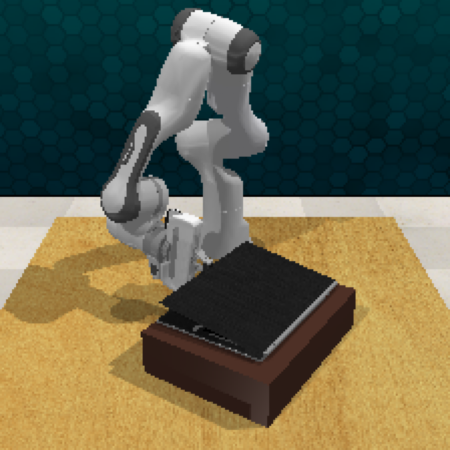} & 
    \includegraphics[width=1\linewidth, valign=m]{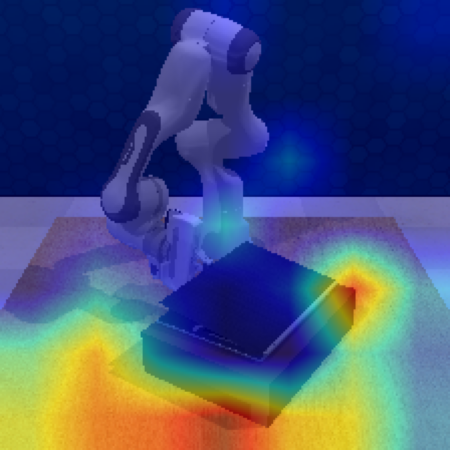} &
    \includegraphics[width=1\linewidth, valign=m]{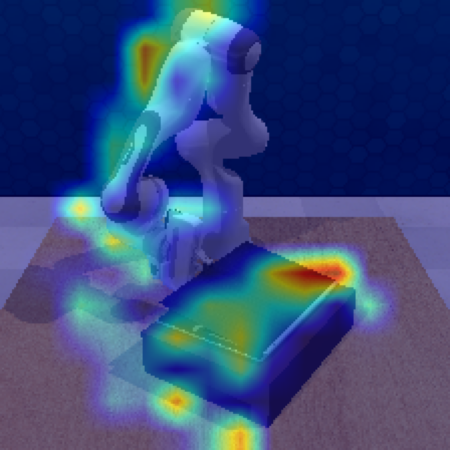} &
    \includegraphics[width=1\linewidth, valign=m]{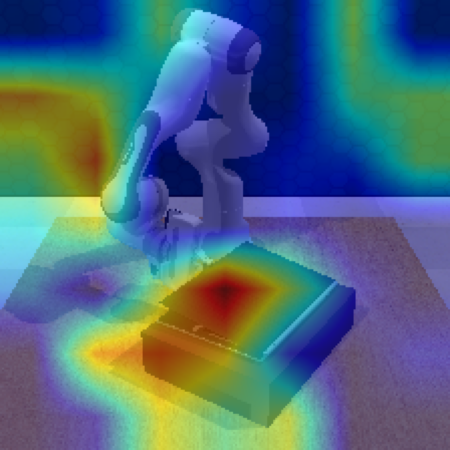} &
    \includegraphics[width=1\linewidth, valign=m]{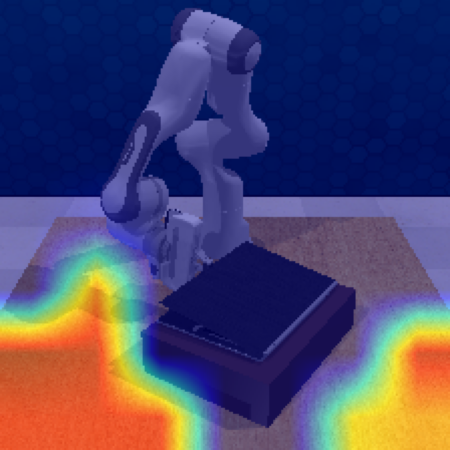} &
    \includegraphics[width=1\linewidth, valign=m]{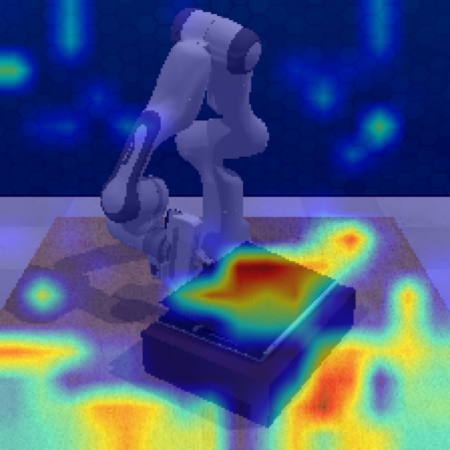} &
    \includegraphics[width=1\linewidth, valign=m]{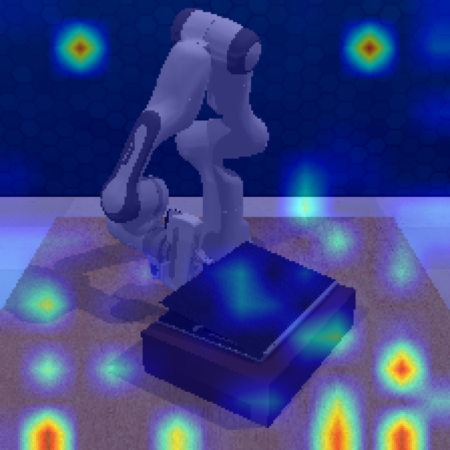} &
    \includegraphics[width=1\linewidth, valign=m]{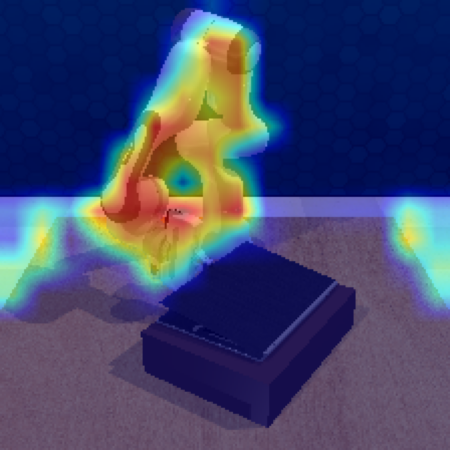} \\ \addlinespace[1ex]

    \rotatebox[origin=c]{90}{\scriptsize Unplug-charger} & 
    \includegraphics[width=1\linewidth, valign=m]{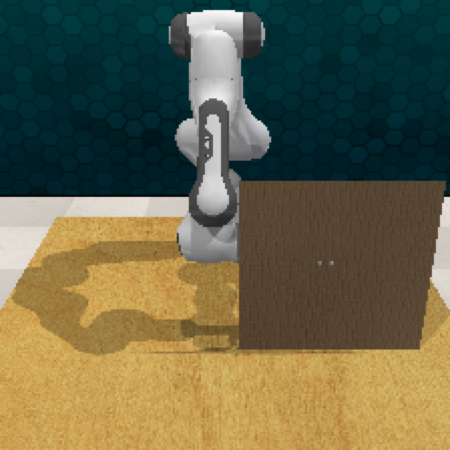} & 
    \includegraphics[width=1\linewidth, valign=m]{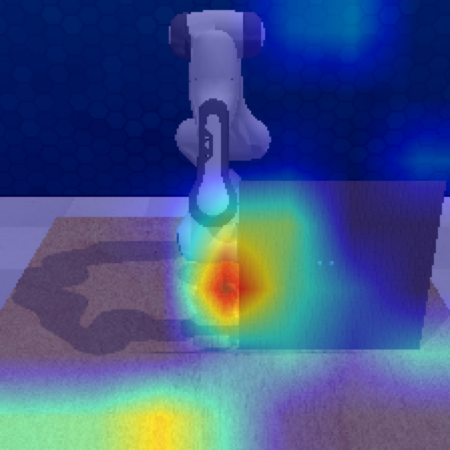} &
    \includegraphics[width=1\linewidth, valign=m]{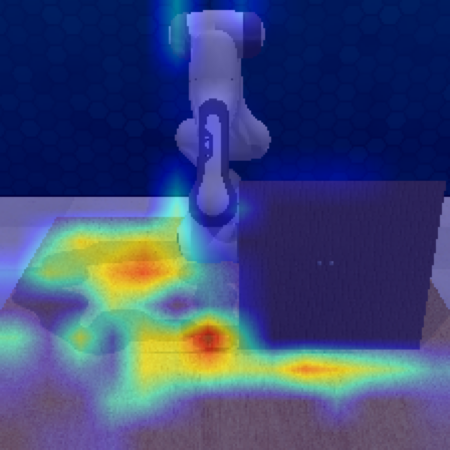} &
    \includegraphics[width=1\linewidth, valign=m]{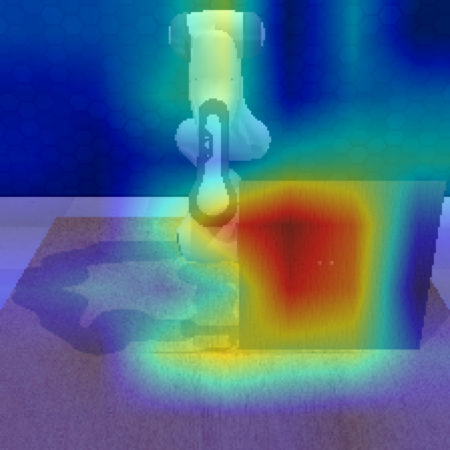} &
    \includegraphics[width=1\linewidth, valign=m]{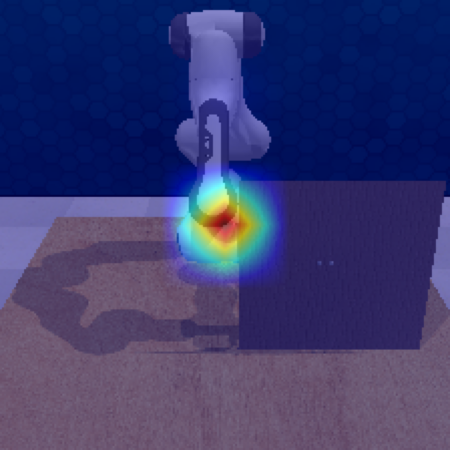} &
    \includegraphics[width=1\linewidth, valign=m]{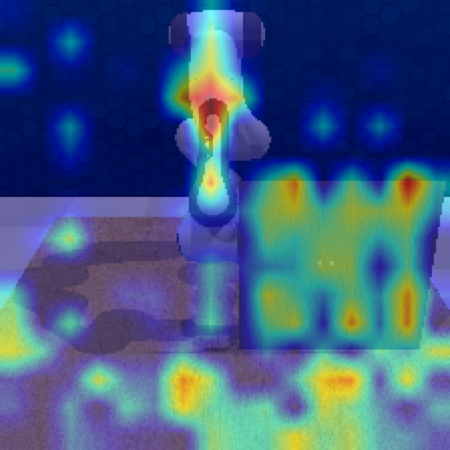} &
    \includegraphics[width=1\linewidth, valign=m]{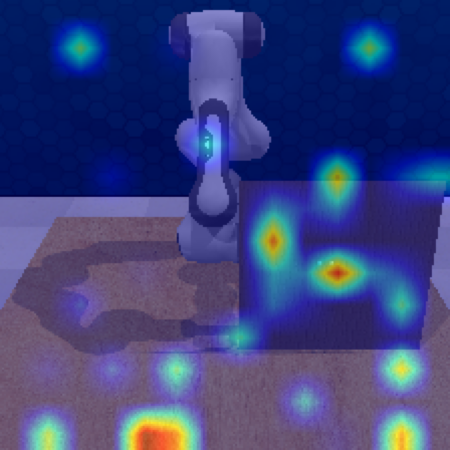} &
    \includegraphics[width=1\linewidth, valign=m]{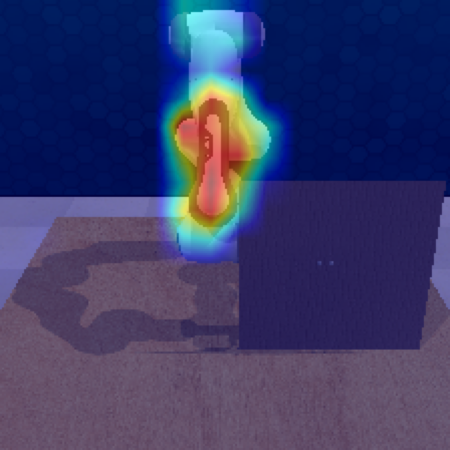} \\ \addlinespace[1ex]

    \rotatebox[origin=c]{90}{\scriptsize Water-plants} & 
    \includegraphics[width=1\linewidth, valign=m]{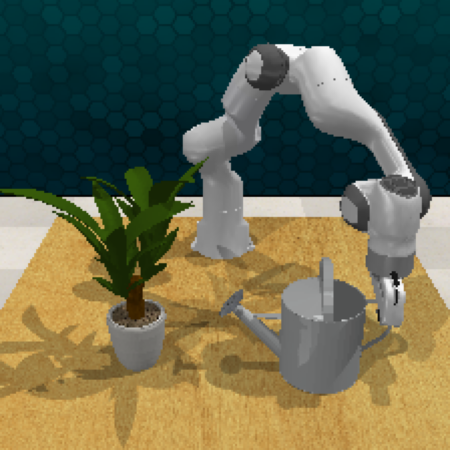} & 
    \includegraphics[width=1\linewidth, valign=m]{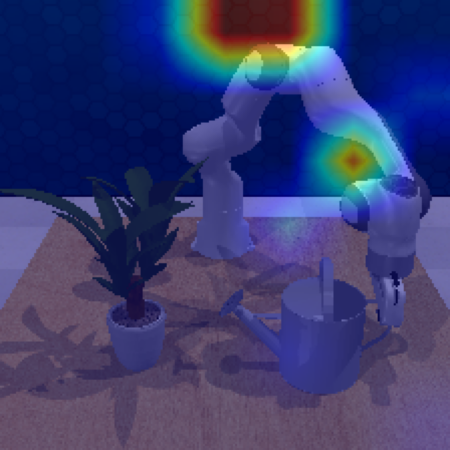} &
    \includegraphics[width=1\linewidth, valign=m]{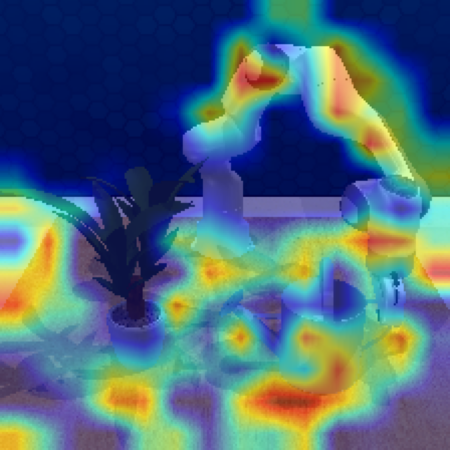} &
    \includegraphics[width=1\linewidth, valign=m]{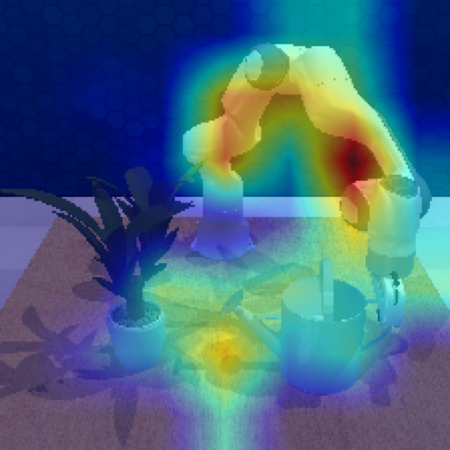} &
    \includegraphics[width=1\linewidth, valign=m]{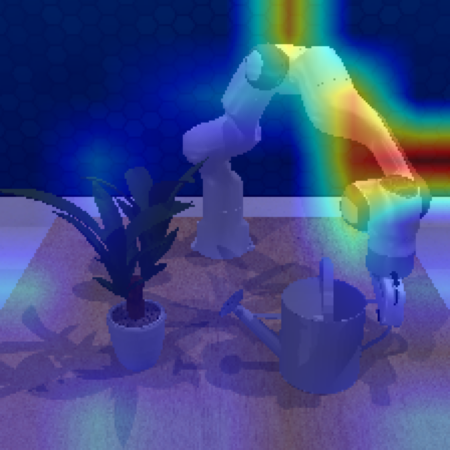} &
    \includegraphics[width=1\linewidth, valign=m]{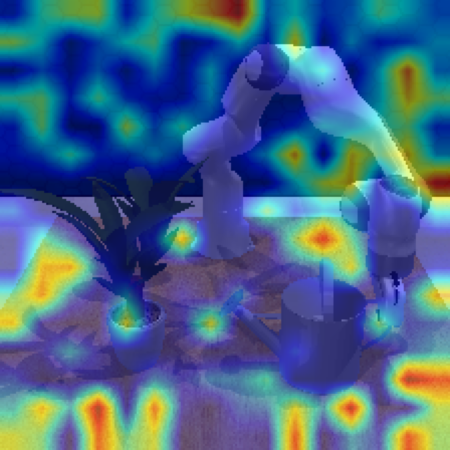} &
    \includegraphics[width=1\linewidth, valign=m]{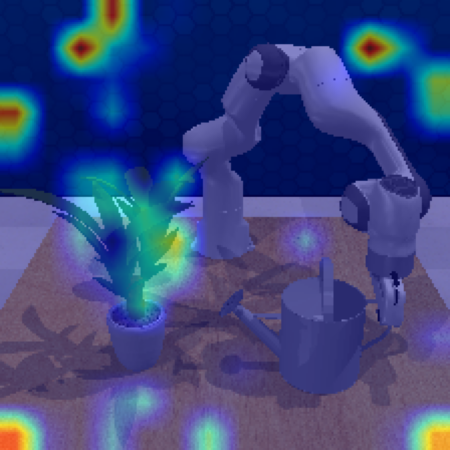} &
    \includegraphics[width=1\linewidth, valign=m]{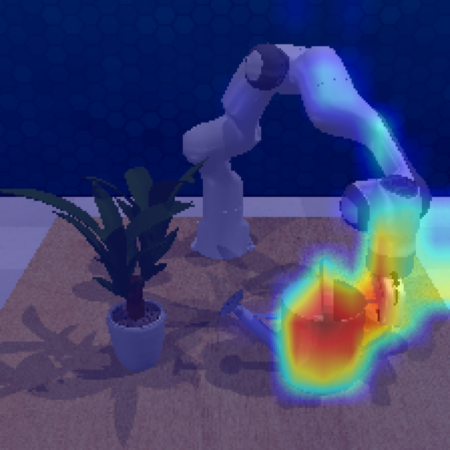} \\ \addlinespace[1ex]
\end{longtable}

}

\subsection{PCA Visualizations}
\label{app:additional_pca}

\noindent\textbf{Visualization protocol.}
We apply PCA to the spatial features of each encoder. For ViT-based encoders (VC-1, CLIP, DINOv2, SigLIP, DynaFLIP), we use the patch tokens; for CNN-based encoders (R3M, LIV), we use the $7 \times 7$ output of the final convolution block. We project the resulting features to 3 principal components and map them to RGB.

\noindent\textbf{Additional visualizations.} We provide additional PCA visualizations to complement the qualitative analysis in Section~\ref{sec:analysis}.

{\setlength{\tabcolsep}{0.25mm}
\begin{longtable}{
    >{\centering\arraybackslash}m{0.0175\linewidth}
    *{8}{>{\centering\arraybackslash}m{0.1175\linewidth}}
@{}}
    \caption{PCA visualizations of learned representations}\label{tab:pca_appendix} \\
    \toprule
    & \small Task & \small R3M & \small VC-1 & \small LIV & \small CLIP & \small DINOv2 & \small SigLIP & \small \textbf{\method} \\ \midrule
    \endfirsthead

    \caption[]{PCA visualizations of learned representations (Continued)} \\
    \toprule
    & \small Task & \small R3M & \small VC-1 & \small LIV & \small CLIP & \small DINOv2 & \small SigLIP & \small \textbf{\method} \\ \midrule
    \endhead

    \midrule
    \multicolumn{9}{r}{\small Continued on next page} \\
    \endfoot

    \bottomrule
    \endlastfoot

% 1. Assembly
    \rotatebox[origin=c]{90}{\scriptsize Assembly} &
    \includegraphics[width=1\linewidth, valign=m]{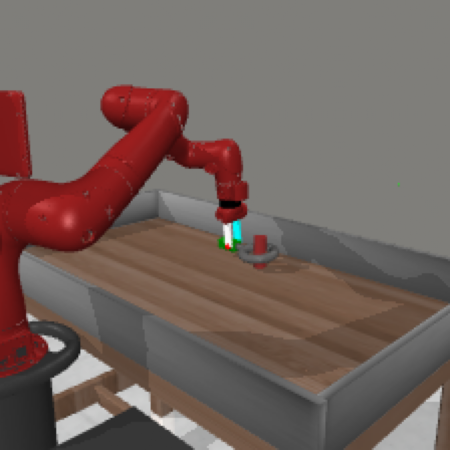} &
    \includegraphics[width=1\linewidth, valign=m]{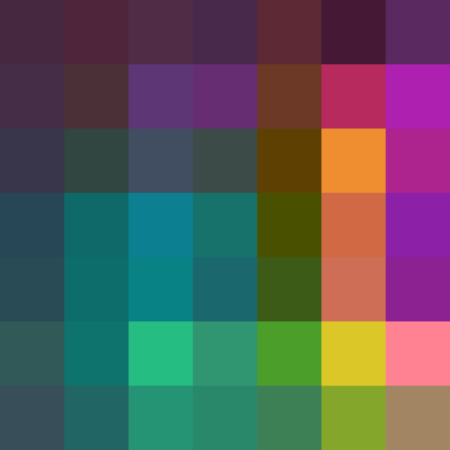} &
    \includegraphics[width=1\linewidth, valign=m]{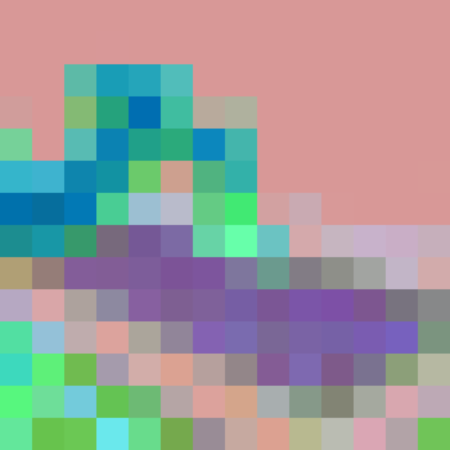} &
    \includegraphics[width=1\linewidth, valign=m]{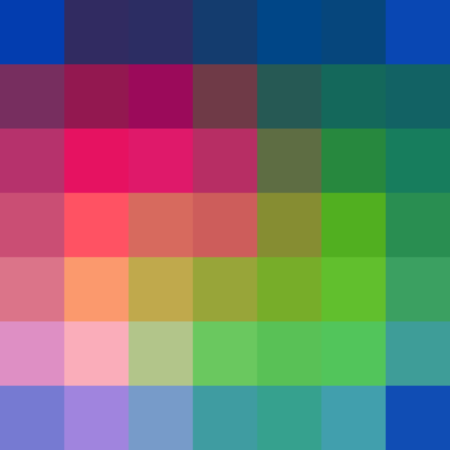} &
    \includegraphics[width=1\linewidth, valign=m]{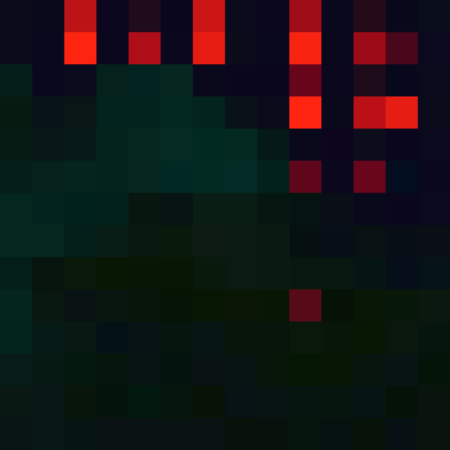} &
    \includegraphics[width=1\linewidth, valign=m]{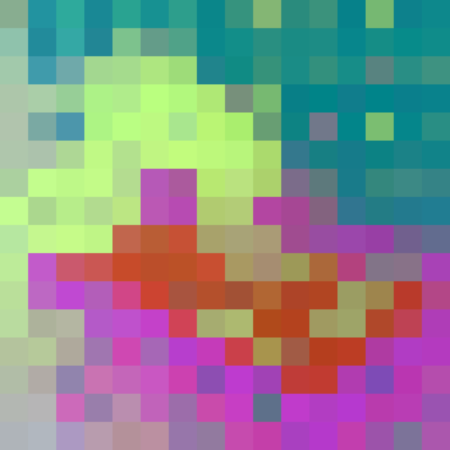} &
    \includegraphics[width=1\linewidth, valign=m]{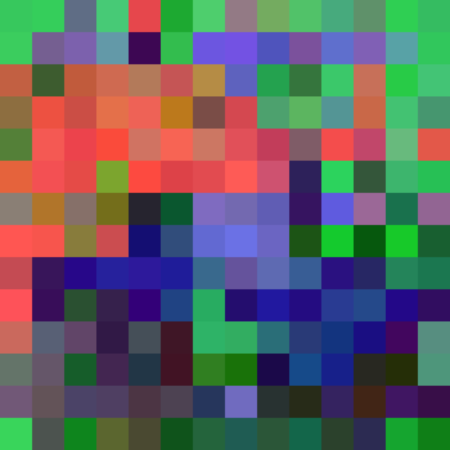} &
    \includegraphics[width=1\linewidth, valign=m]{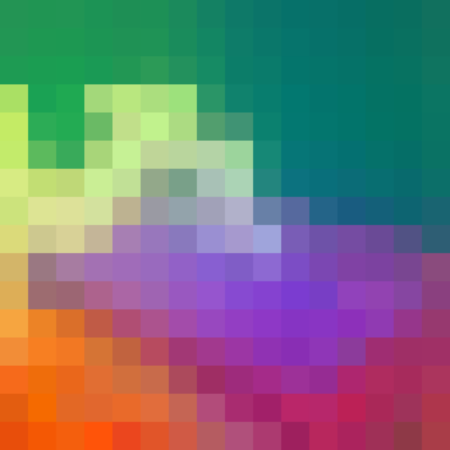} \\ \addlinespace[1ex]

% 2. Bin-picking
    \rotatebox[origin=c]{90}{\scriptsize Bin-picking} &
    \includegraphics[width=1\linewidth, valign=m]{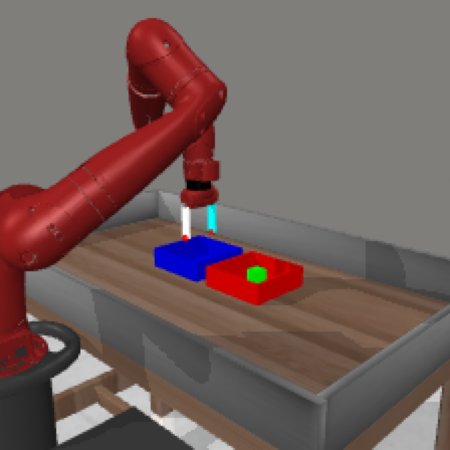} &
    \includegraphics[width=1\linewidth, valign=m]{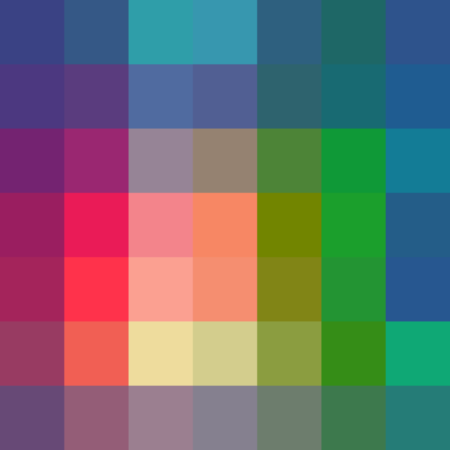} &
    \includegraphics[width=1\linewidth, valign=m]{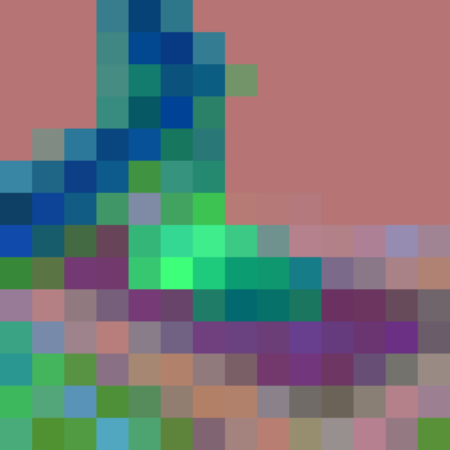} &
    \includegraphics[width=1\linewidth, valign=m]{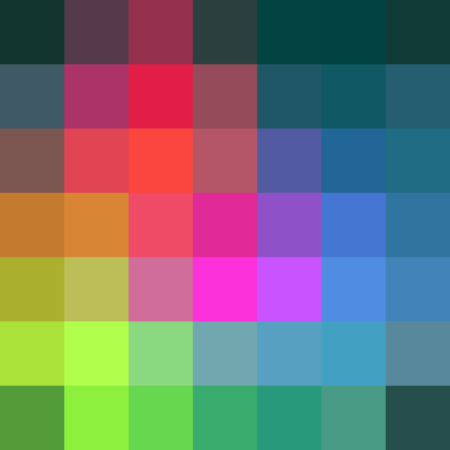} &
    \includegraphics[width=1\linewidth, valign=m]{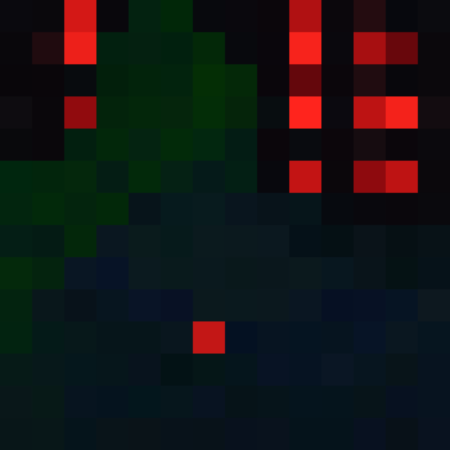} &
    \includegraphics[width=1\linewidth, valign=m]{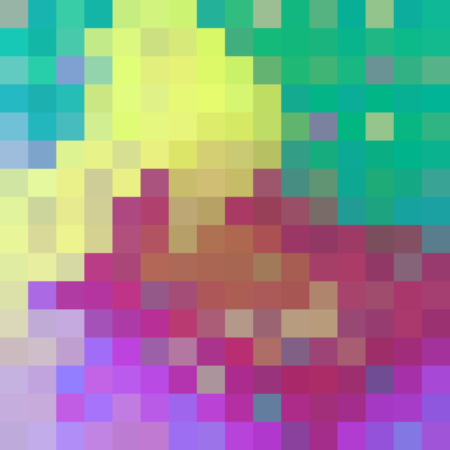} &
    \includegraphics[width=1\linewidth, valign=m]{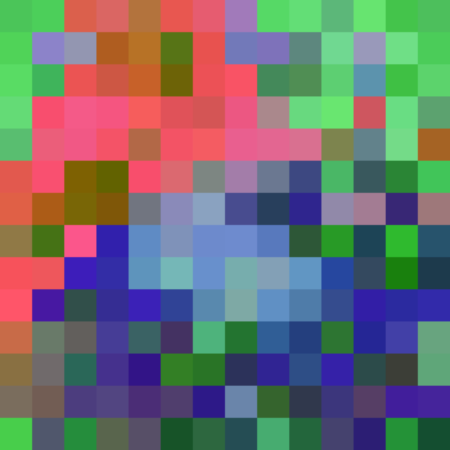} &
    \includegraphics[width=1\linewidth, valign=m]{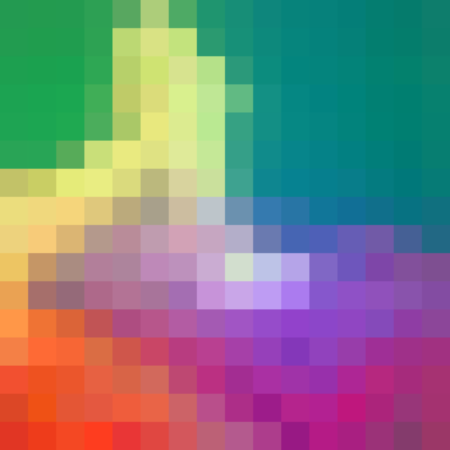} \\ \addlinespace[1ex]

% 3. Box-close
    \rotatebox[origin=c]{90}{\scriptsize Box-close} &
    \includegraphics[width=1\linewidth, valign=m]{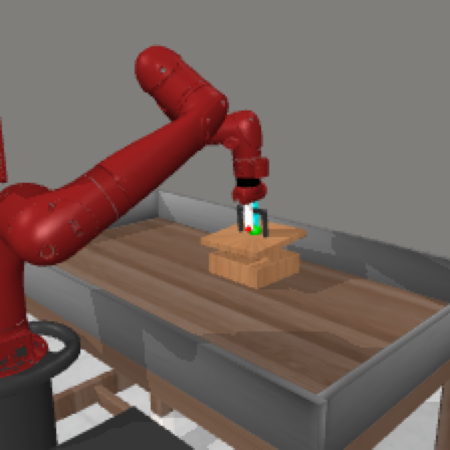} &
    \includegraphics[width=1\linewidth, valign=m]{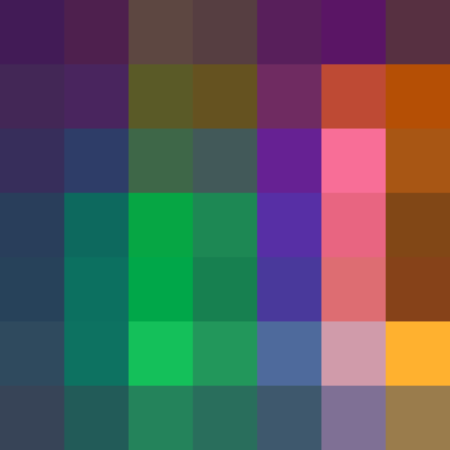} &
    \includegraphics[width=1\linewidth, valign=m]{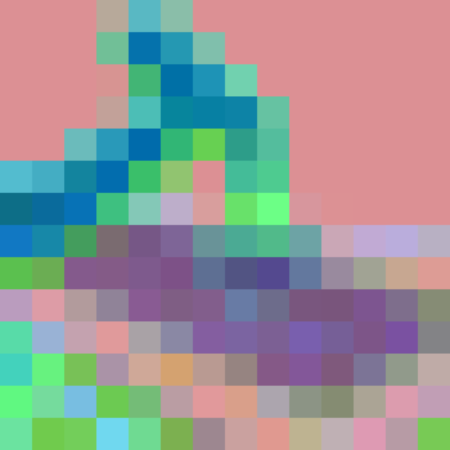} &
    \includegraphics[width=1\linewidth, valign=m]{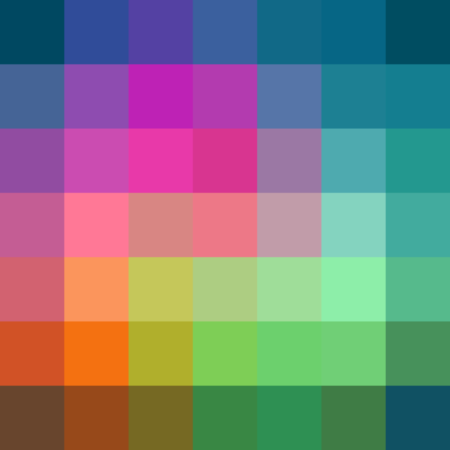} &
    \includegraphics[width=1\linewidth, valign=m]{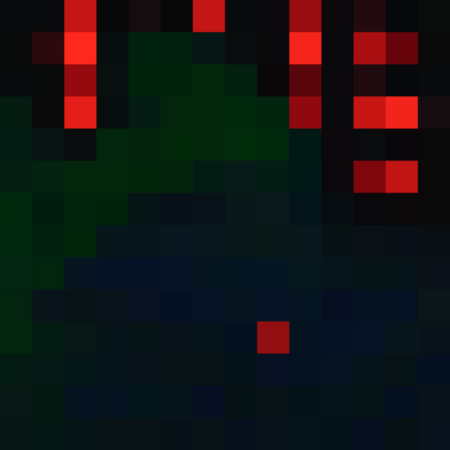} &
    \includegraphics[width=1\linewidth, valign=m]{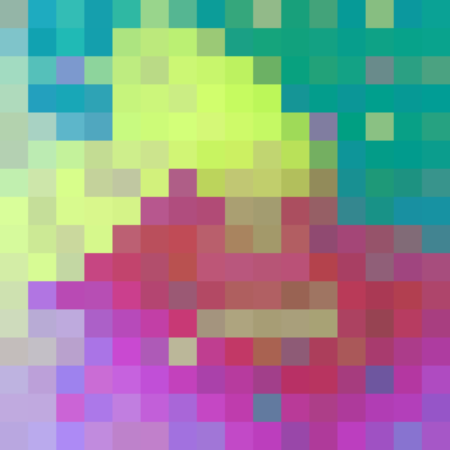} &
    \includegraphics[width=1\linewidth, valign=m]{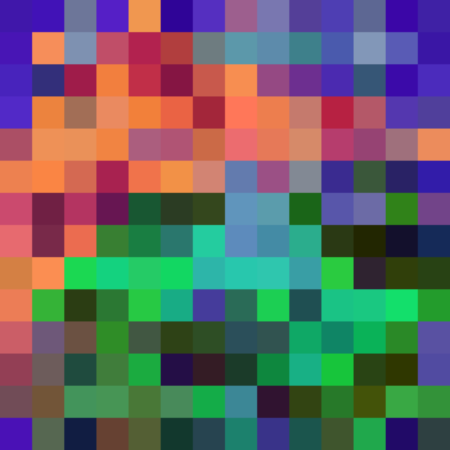} &
    \includegraphics[width=1\linewidth, valign=m]{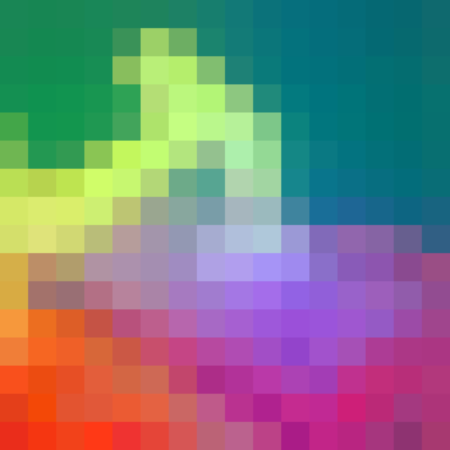} \\ \addlinespace[1ex]

% 4. Button-press
    \rotatebox[origin=c]{90}{\scriptsize Button-press} &
    \includegraphics[width=1\linewidth, valign=m]{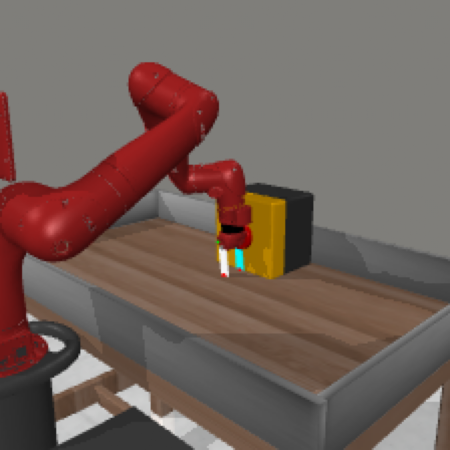} &
    \includegraphics[width=1\linewidth, valign=m]{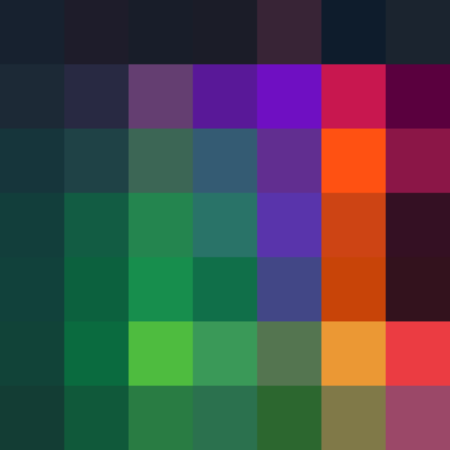} &
    \includegraphics[width=1\linewidth, valign=m]{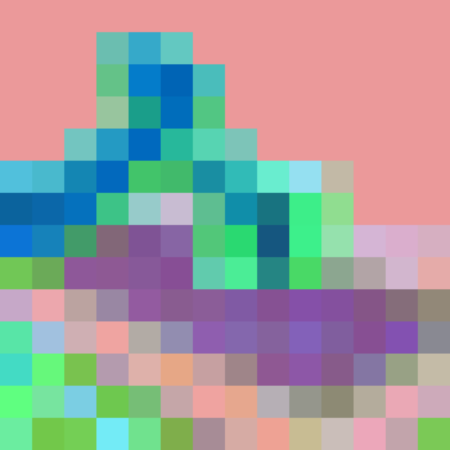} &
    \includegraphics[width=1\linewidth, valign=m]{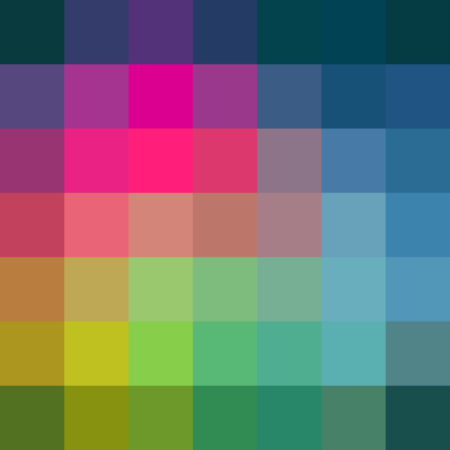} &
    \includegraphics[width=1\linewidth, valign=m]{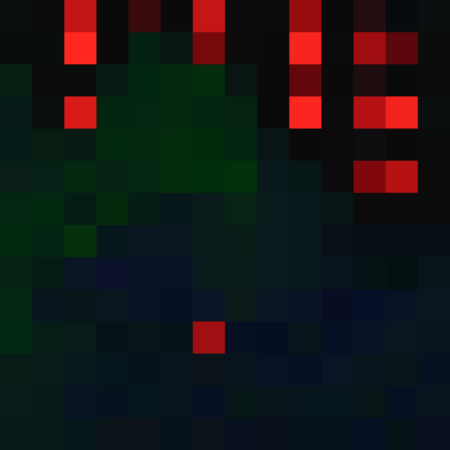} &
    \includegraphics[width=1\linewidth, valign=m]{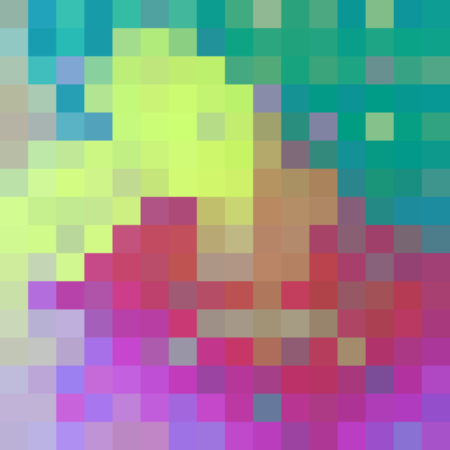} &
    \includegraphics[width=1\linewidth, valign=m]{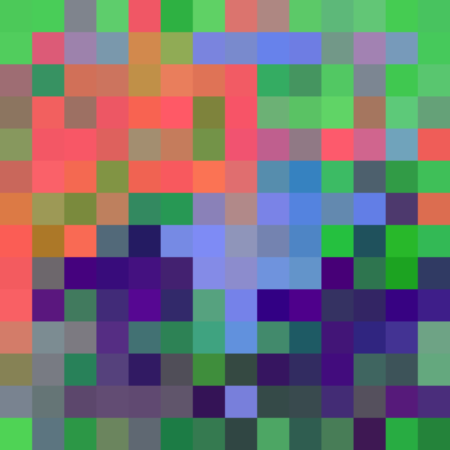} &
    \includegraphics[width=1\linewidth, valign=m]{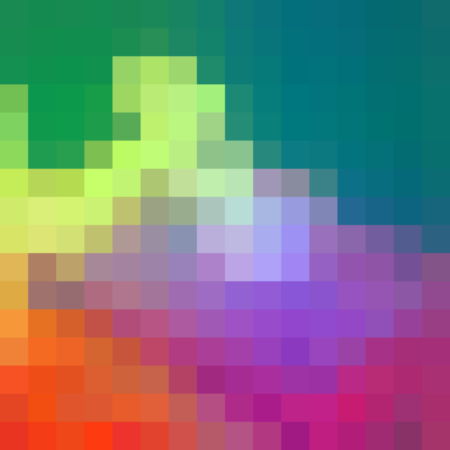} \\ \addlinespace[1ex]

% 5. Dial-turn
    \rotatebox[origin=c]{90}{\scriptsize Dial-turn} &
    \includegraphics[width=1\linewidth, valign=m]{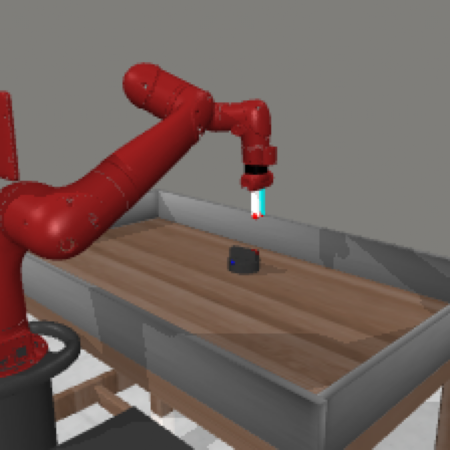} &
    \includegraphics[width=1\linewidth, valign=m]{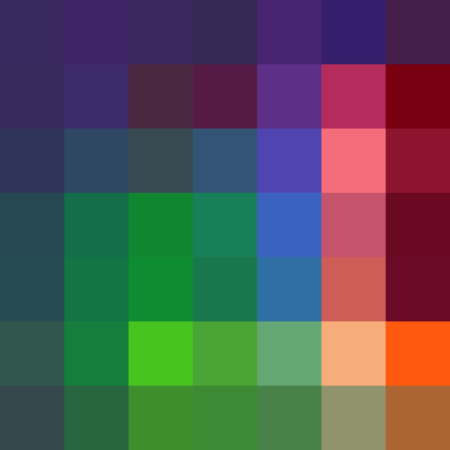} &
    \includegraphics[width=1\linewidth, valign=m]{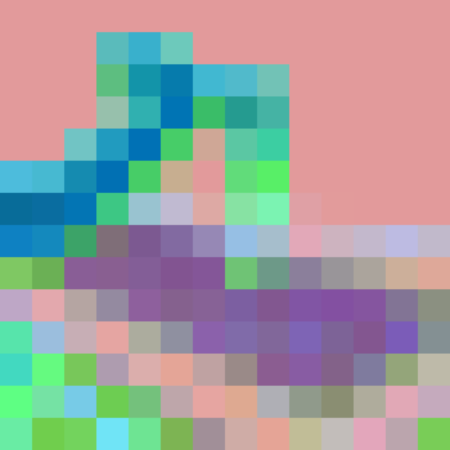} &
    \includegraphics[width=1\linewidth, valign=m]{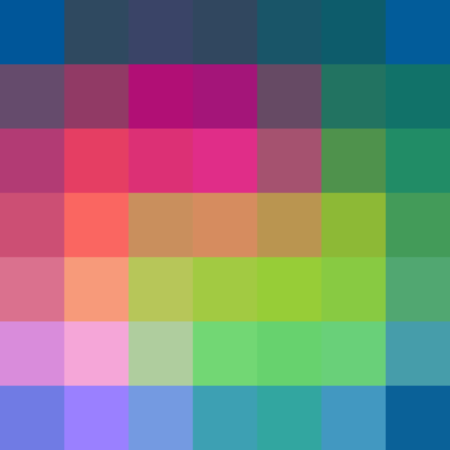} &
    \includegraphics[width=1\linewidth, valign=m]{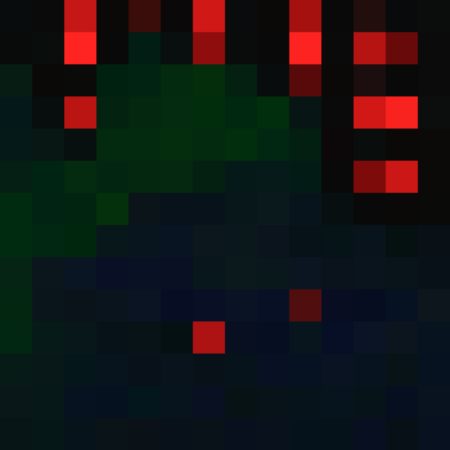} &
    \includegraphics[width=1\linewidth, valign=m]{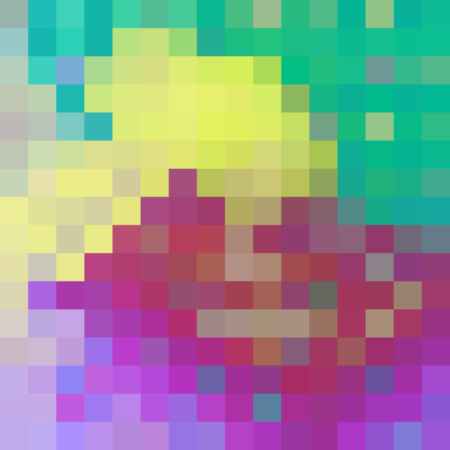} &
    \includegraphics[width=1\linewidth, valign=m]{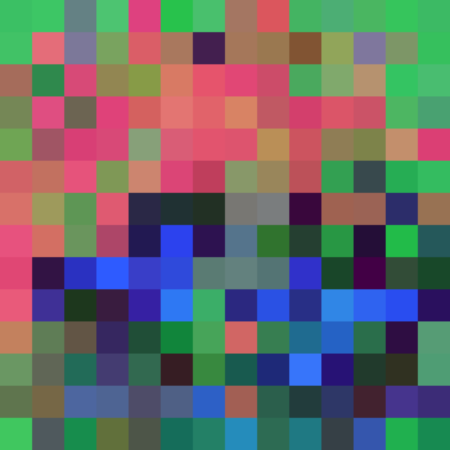} &
    \includegraphics[width=1\linewidth, valign=m]{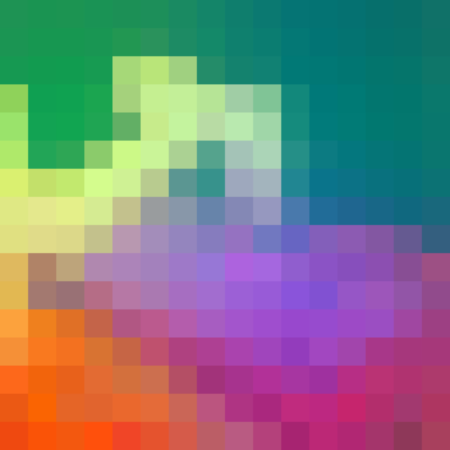} \\ \addlinespace[1ex]

% 6. Drawer-open
    \rotatebox[origin=c]{90}{\scriptsize Drawer-open} &
    \includegraphics[width=1\linewidth, valign=m]{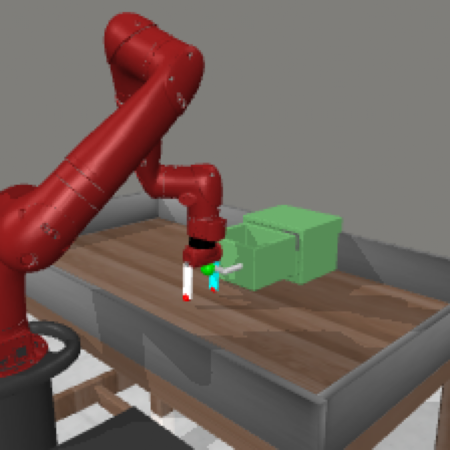} &
    \includegraphics[width=1\linewidth, valign=m]{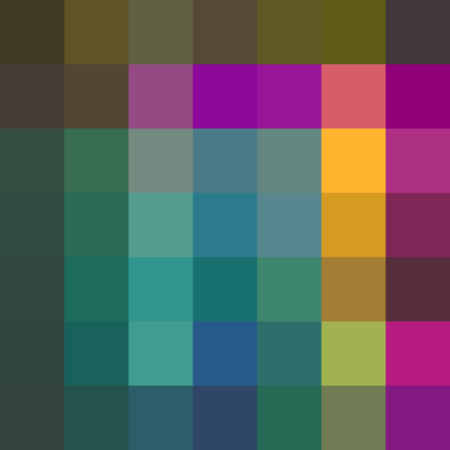} &
    \includegraphics[width=1\linewidth, valign=m]{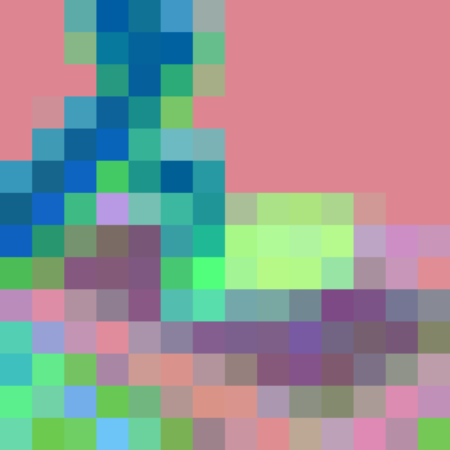} &
    \includegraphics[width=1\linewidth, valign=m]{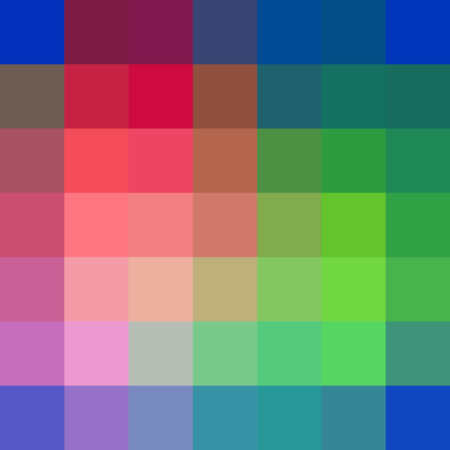} &
    \includegraphics[width=1\linewidth, valign=m]{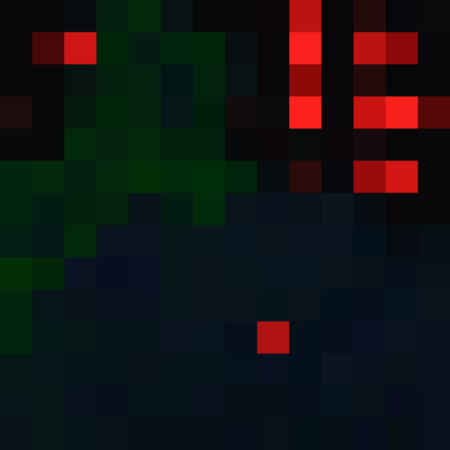} &
    \includegraphics[width=1\linewidth, valign=m]{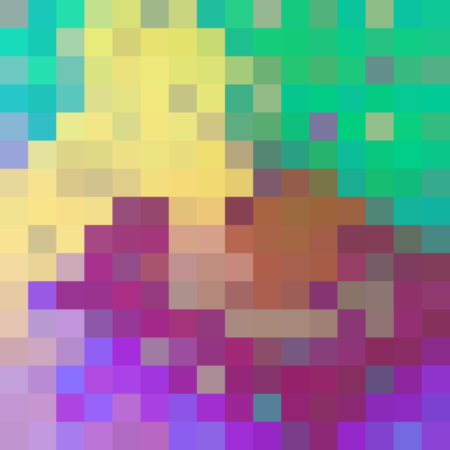} &
    \includegraphics[width=1\linewidth, valign=m]{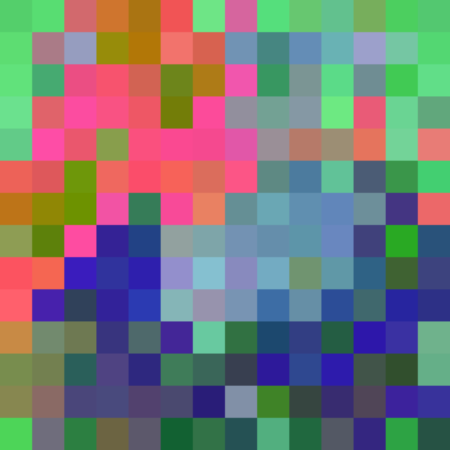} &
    \includegraphics[width=1\linewidth, valign=m]{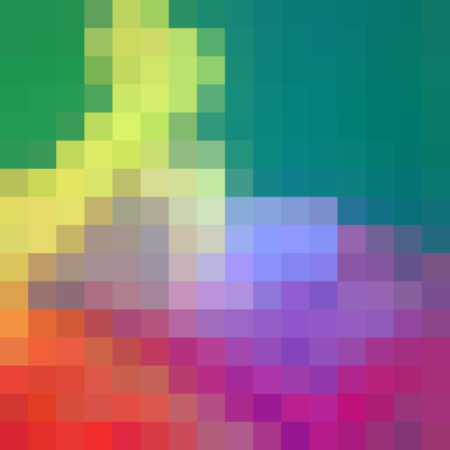} \\ \addlinespace[1ex]

% 7. Hammer
    \rotatebox[origin=c]{90}{\scriptsize Hammer} &
    \includegraphics[width=1\linewidth, valign=m]{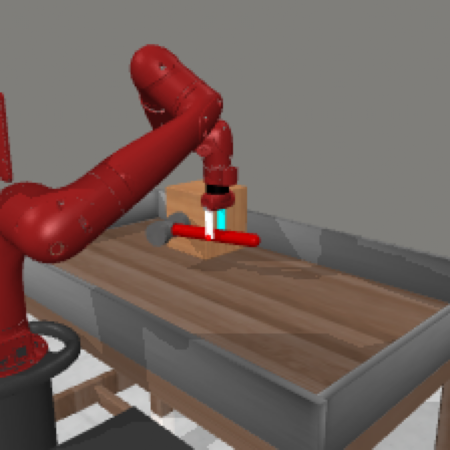} &
    \includegraphics[width=1\linewidth, valign=m]{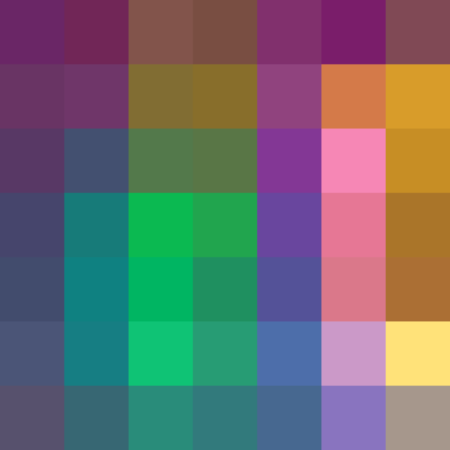} &
    \includegraphics[width=1\linewidth, valign=m]{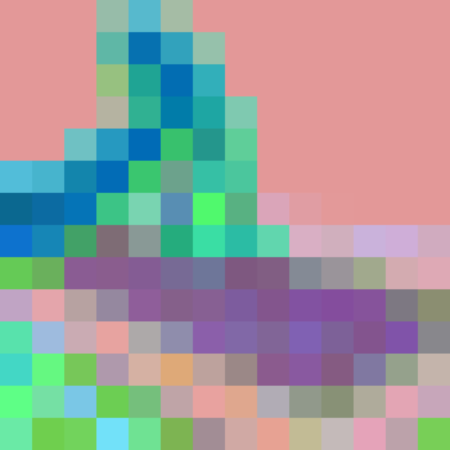} &
    \includegraphics[width=1\linewidth, valign=m]{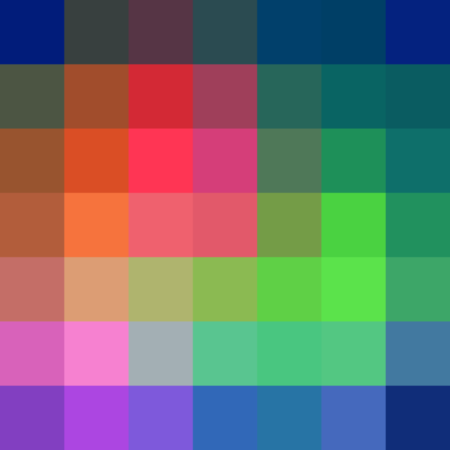} &
    \includegraphics[width=1\linewidth, valign=m]{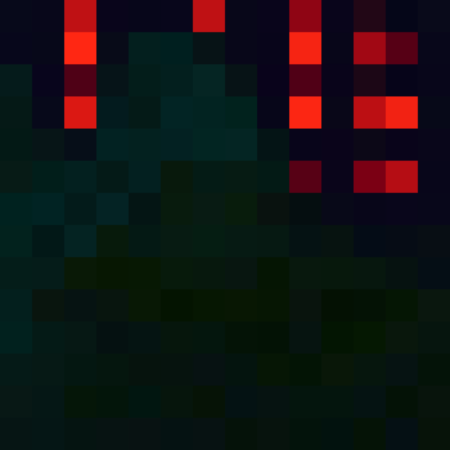} &
    \includegraphics[width=1\linewidth, valign=m]{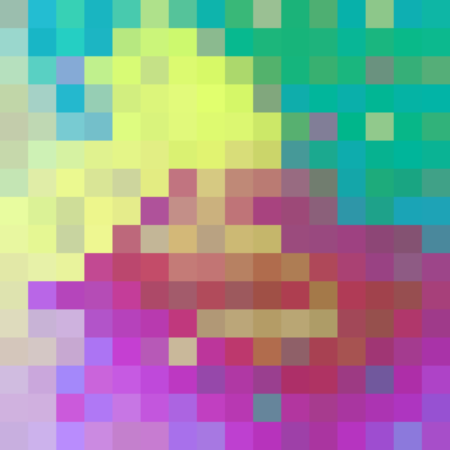} &
    \includegraphics[width=1\linewidth, valign=m]{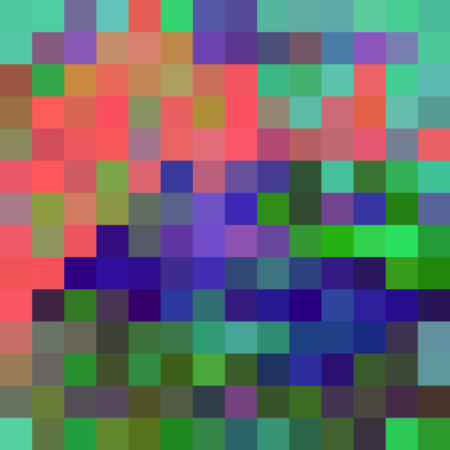} &
    \includegraphics[width=1\linewidth, valign=m]{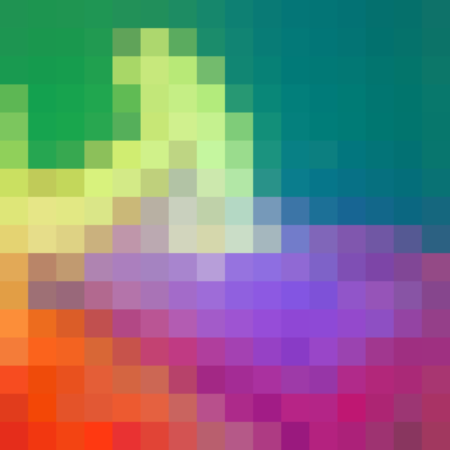} \\ \addlinespace[1ex]

% 8. Hand-insert
    \rotatebox[origin=c]{90}{\scriptsize Hand-insert} &
    \includegraphics[width=1\linewidth, valign=m]{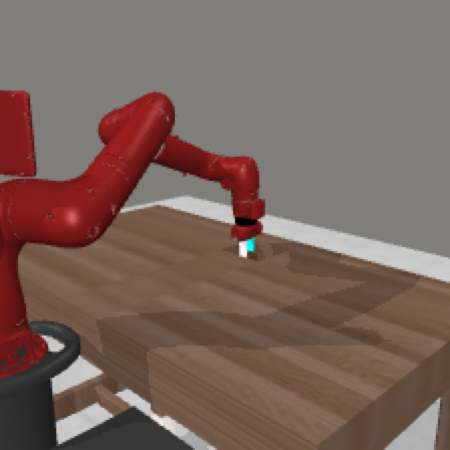} &
    \includegraphics[width=1\linewidth, valign=m]{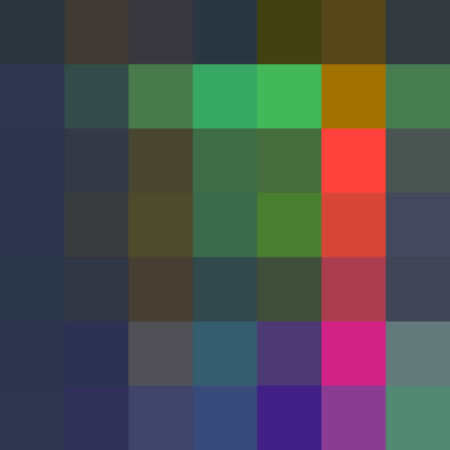} &
    \includegraphics[width=1\linewidth, valign=m]{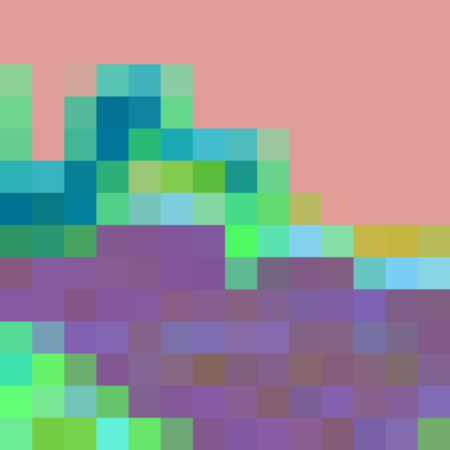} &
    \includegraphics[width=1\linewidth, valign=m]{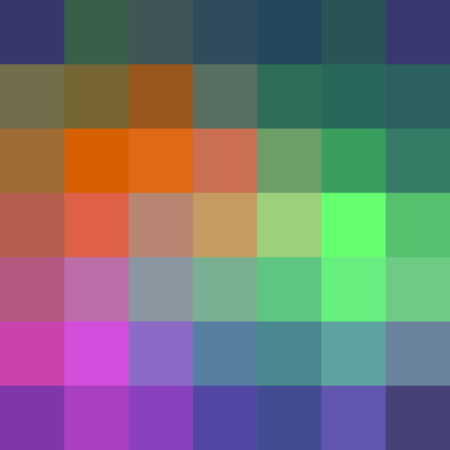} &
    \includegraphics[width=1\linewidth, valign=m]{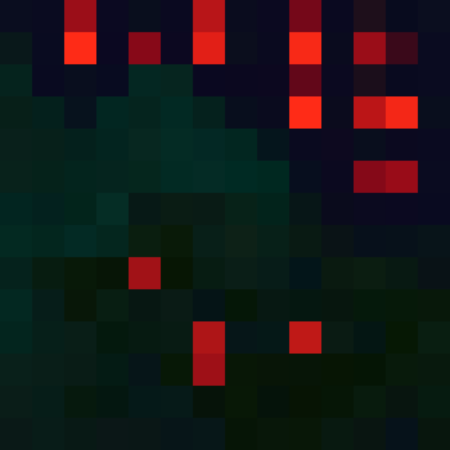} &
    \includegraphics[width=1\linewidth, valign=m]{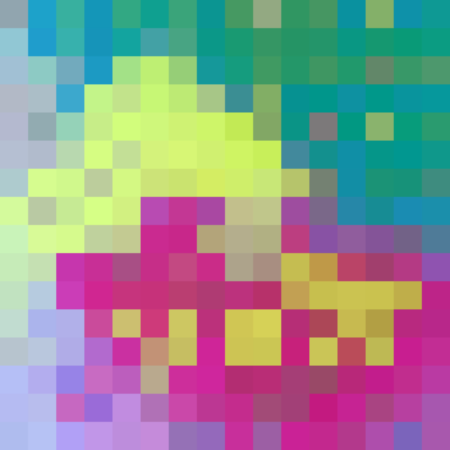} &
    \includegraphics[width=1\linewidth, valign=m]{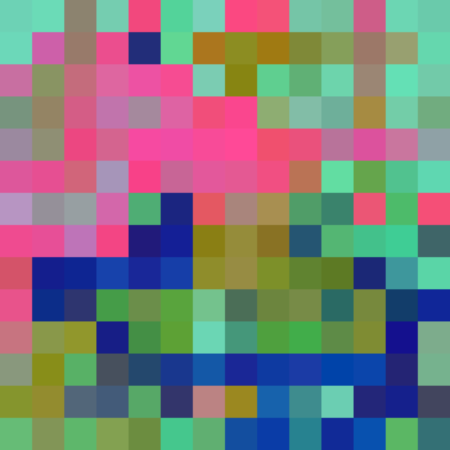} &
    \includegraphics[width=1\linewidth, valign=m]{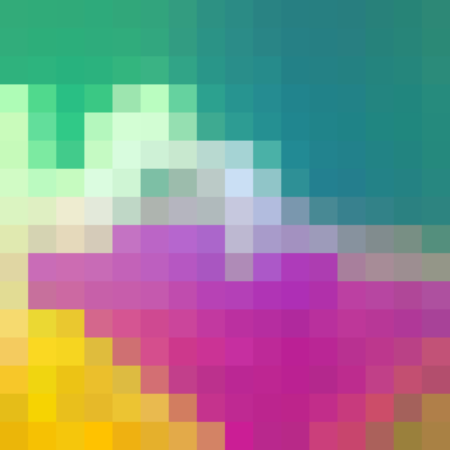} \\ \addlinespace[1ex]

% 9. Handle-pull
    \rotatebox[origin=c]{90}{\scriptsize Handle-pull} &
    \includegraphics[width=1\linewidth, valign=m]{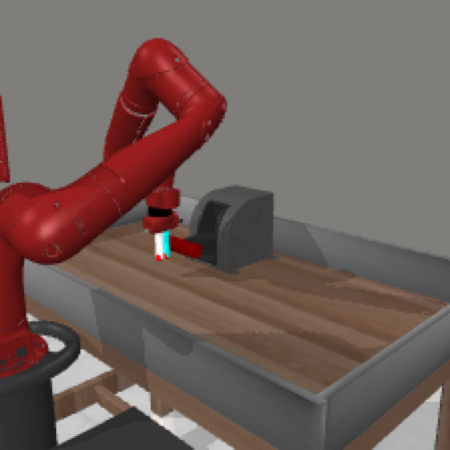} &
    \includegraphics[width=1\linewidth, valign=m]{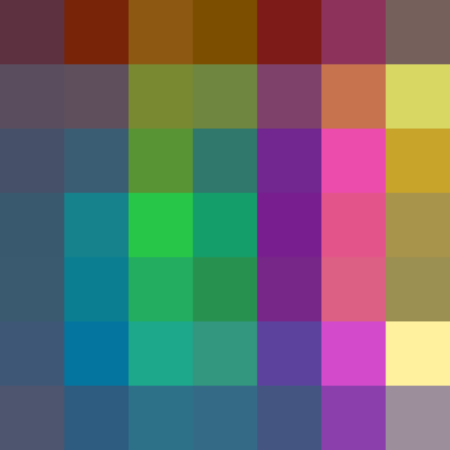} &
    \includegraphics[width=1\linewidth, valign=m]{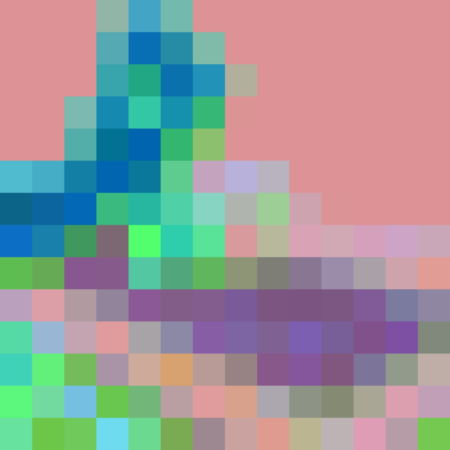} &
    \includegraphics[width=1\linewidth, valign=m]{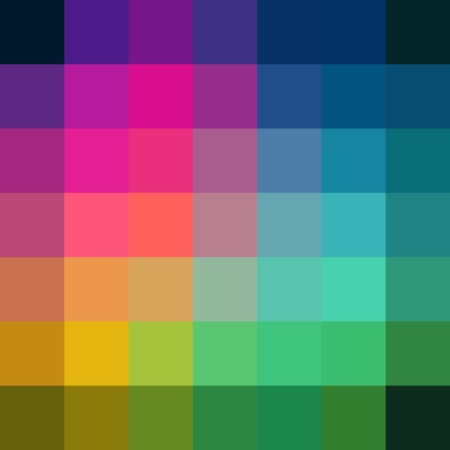} &
    \includegraphics[width=1\linewidth, valign=m]{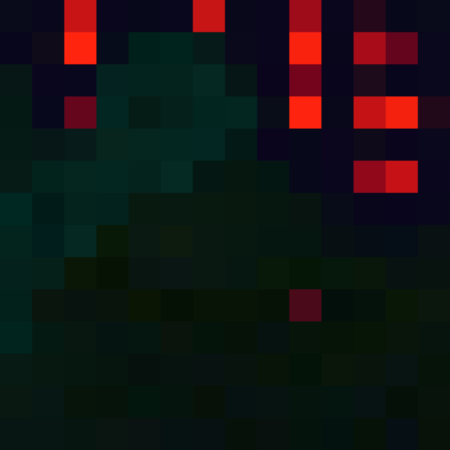} &
    \includegraphics[width=1\linewidth, valign=m]{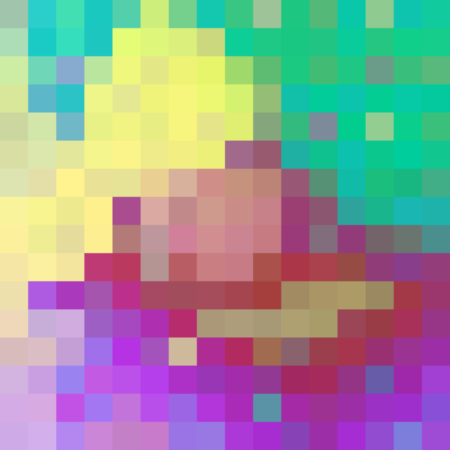} &
    \includegraphics[width=1\linewidth, valign=m]{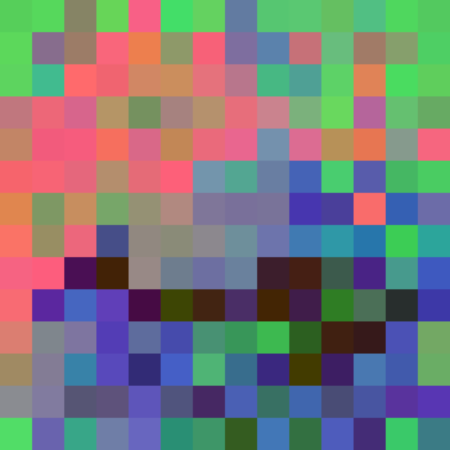} &
    \includegraphics[width=1\linewidth, valign=m]{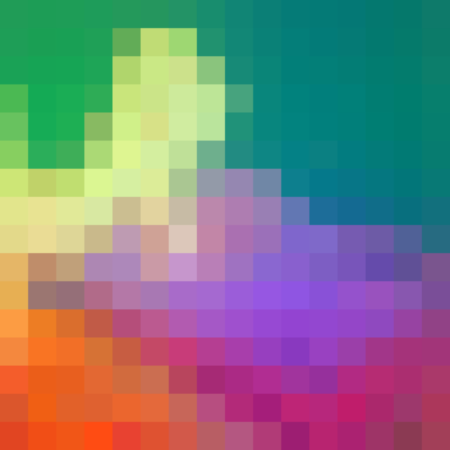} \\ \addlinespace[1ex]

% 10. Lever-pull
    \rotatebox[origin=c]{90}{\scriptsize Lever-pull} &
    \includegraphics[width=1\linewidth, valign=m]{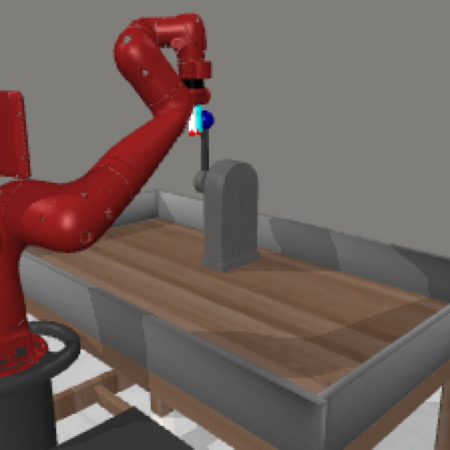} &
    \includegraphics[width=1\linewidth, valign=m]{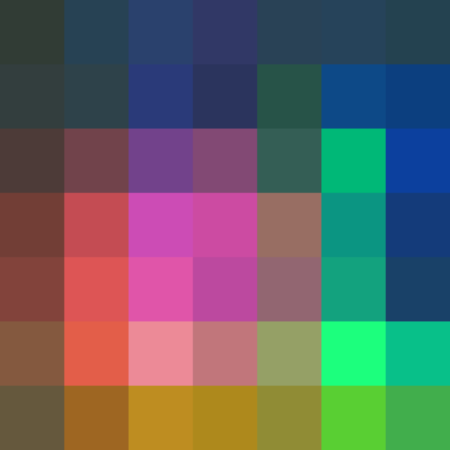} &
    \includegraphics[width=1\linewidth, valign=m]{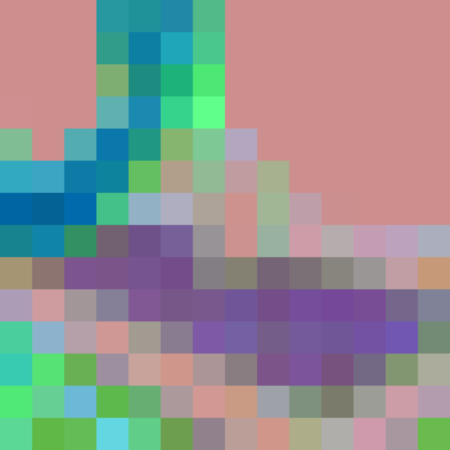} &
    \includegraphics[width=1\linewidth, valign=m]{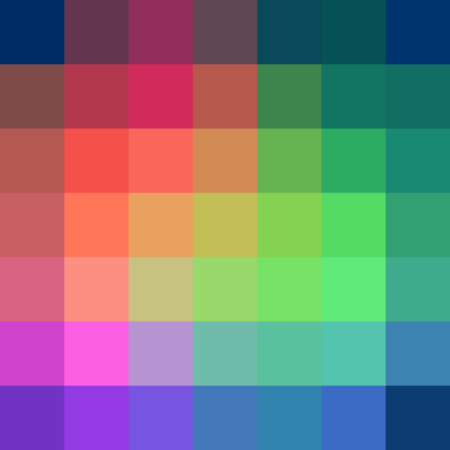} &
    \includegraphics[width=1\linewidth, valign=m]{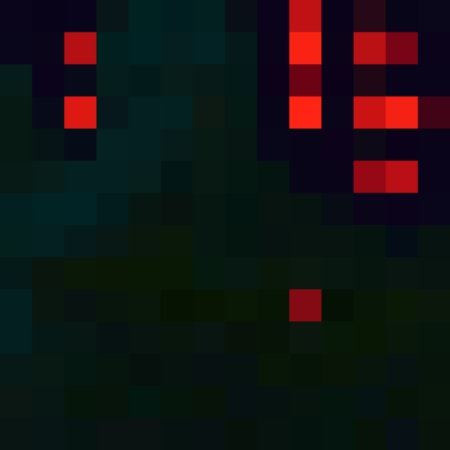} &
    \includegraphics[width=1\linewidth, valign=m]{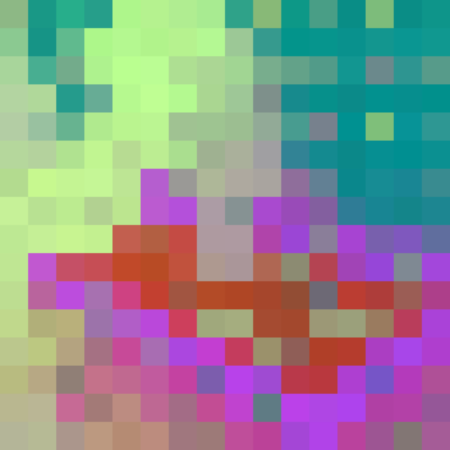} &
    \includegraphics[width=1\linewidth, valign=m]{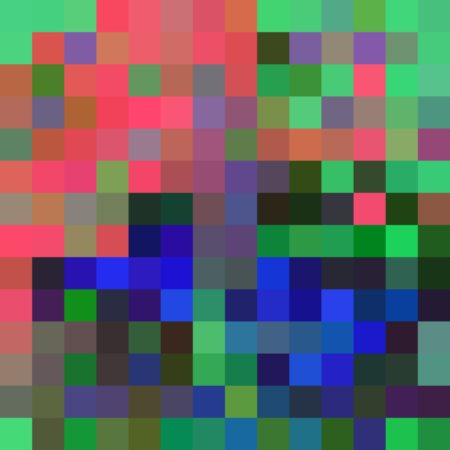} &
    \includegraphics[width=1\linewidth, valign=m]{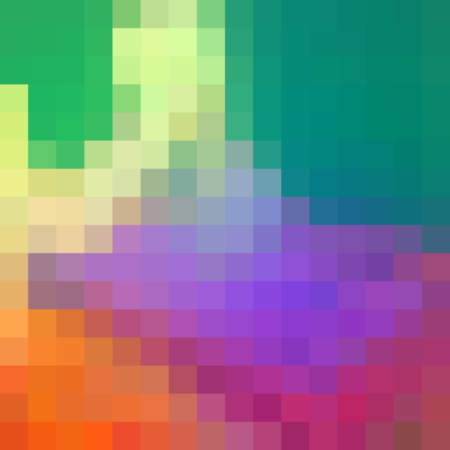} \\ \addlinespace[1ex]

% 11. Peg-unplug-side
    \rotatebox[origin=c]{90}{\scriptsize Peg-unplug-side} &
    \includegraphics[width=1\linewidth, valign=m]{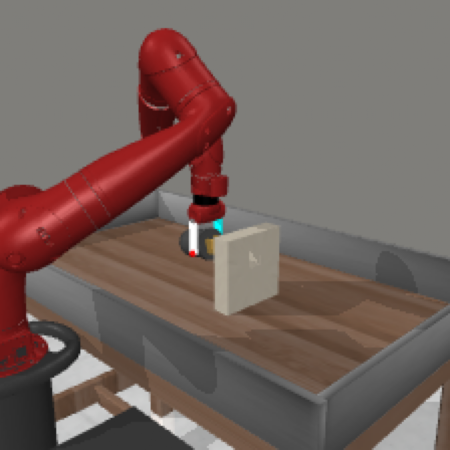} &
    \includegraphics[width=1\linewidth, valign=m]{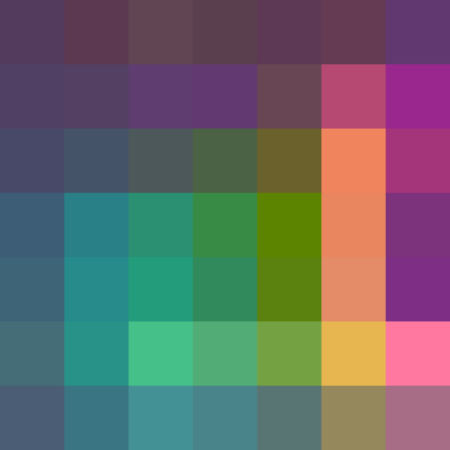} &
    \includegraphics[width=1\linewidth, valign=m]{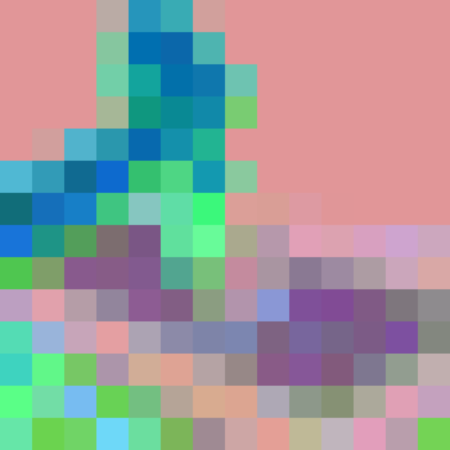} &
    \includegraphics[width=1\linewidth, valign=m]{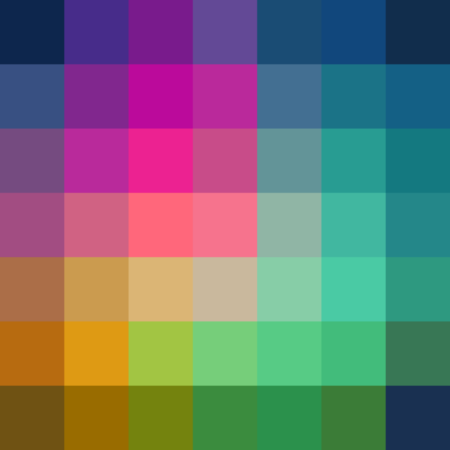} &
    \includegraphics[width=1\linewidth, valign=m]{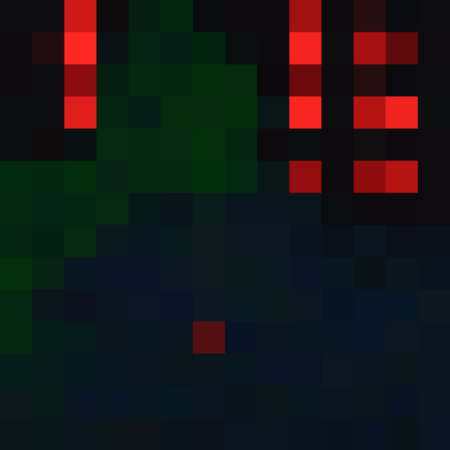} &
    \includegraphics[width=1\linewidth, valign=m]{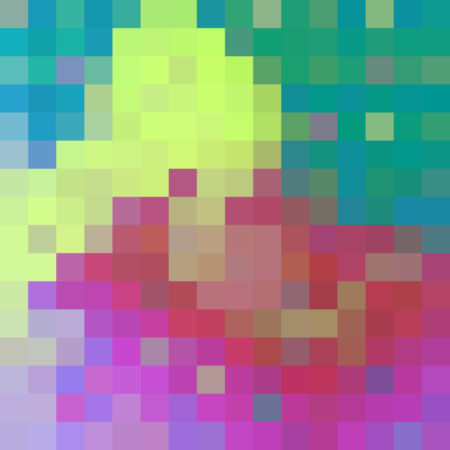} &
    \includegraphics[width=1\linewidth, valign=m]{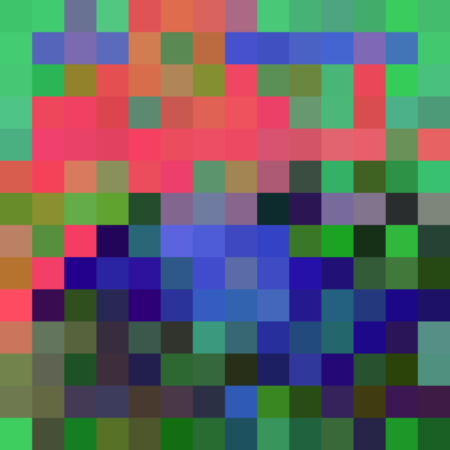} &
    \includegraphics[width=1\linewidth, valign=m]{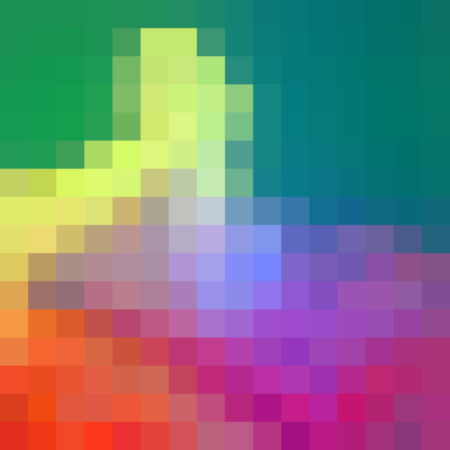} \\ \addlinespace[1ex]

% 12. Push-wall
    \rotatebox[origin=c]{90}{\scriptsize Push-wall} &
    \includegraphics[width=1\linewidth, valign=m]{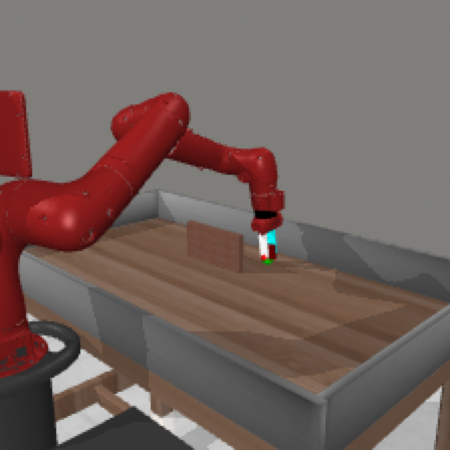} &
    \includegraphics[width=1\linewidth, valign=m]{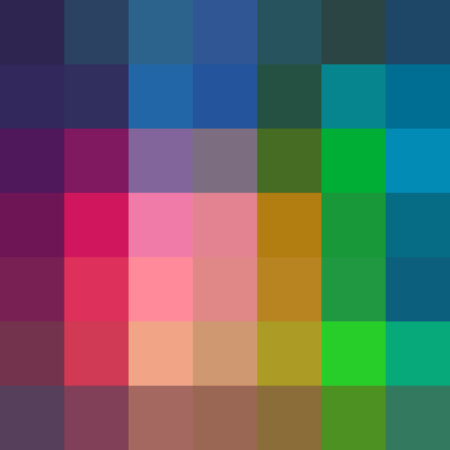} &
    \includegraphics[width=1\linewidth, valign=m]{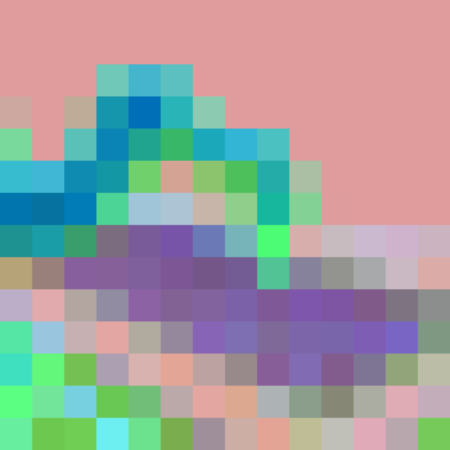} &
    \includegraphics[width=1\linewidth, valign=m]{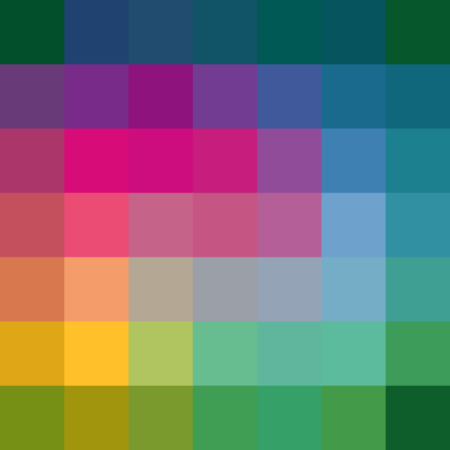} &
    \includegraphics[width=1\linewidth, valign=m]{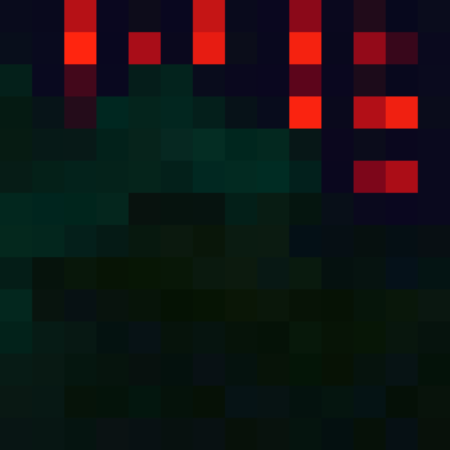} &
    \includegraphics[width=1\linewidth, valign=m]{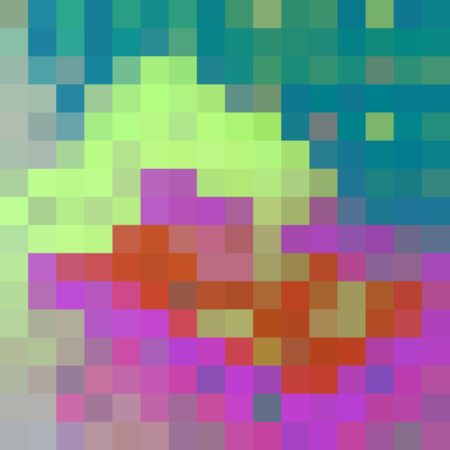} &
    \includegraphics[width=1\linewidth, valign=m]{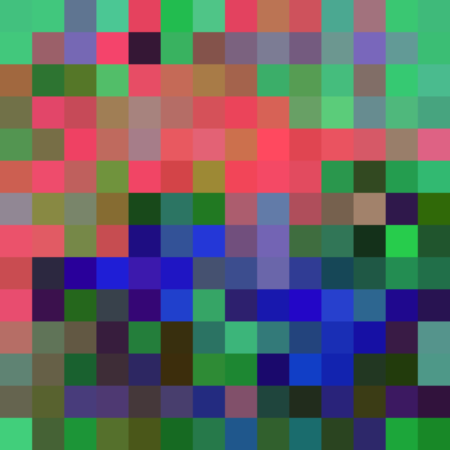} &
    \includegraphics[width=1\linewidth, valign=m]{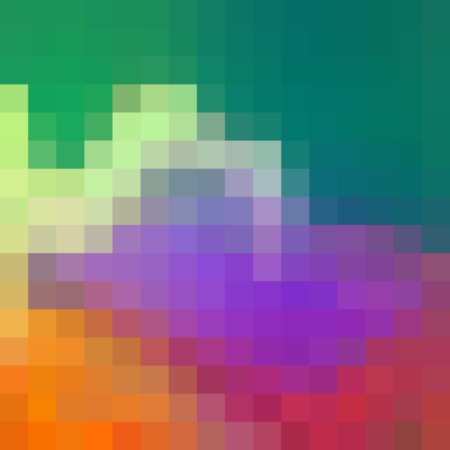} \\ \addlinespace[1ex]

% 13. Reach
    \rotatebox[origin=c]{90}{\scriptsize Reach} &
    \includegraphics[width=1\linewidth, valign=m]{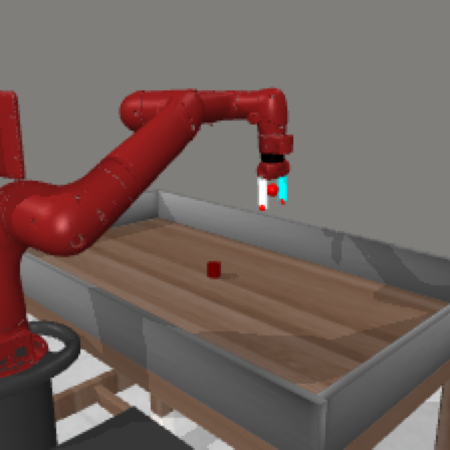} &
    \includegraphics[width=1\linewidth, valign=m]{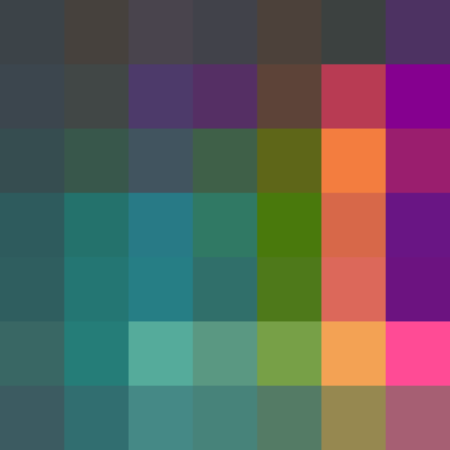} &
    \includegraphics[width=1\linewidth, valign=m]{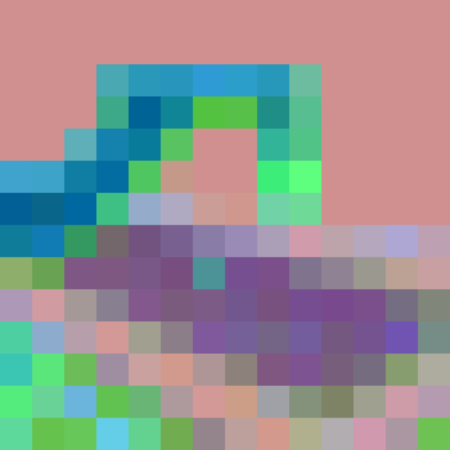} &
    \includegraphics[width=1\linewidth, valign=m]{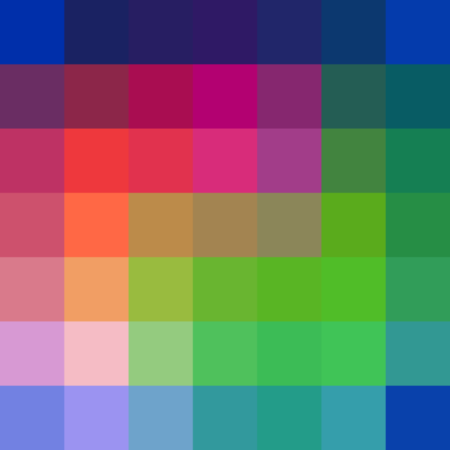} &
    \includegraphics[width=1\linewidth, valign=m]{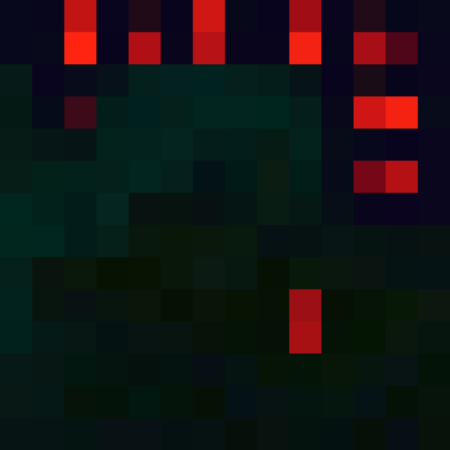} &
    \includegraphics[width=1\linewidth, valign=m]{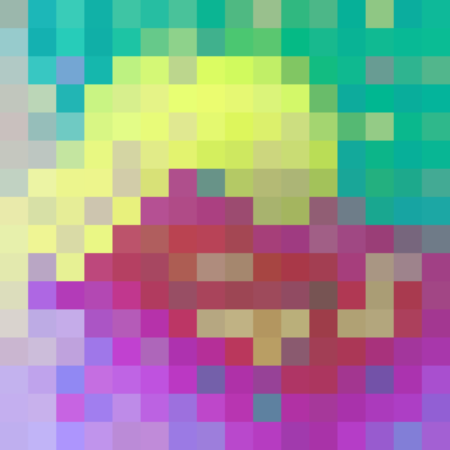} &
    \includegraphics[width=1\linewidth, valign=m]{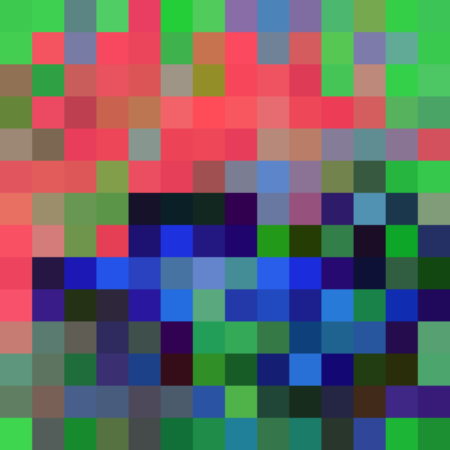} &
    \includegraphics[width=1\linewidth, valign=m]{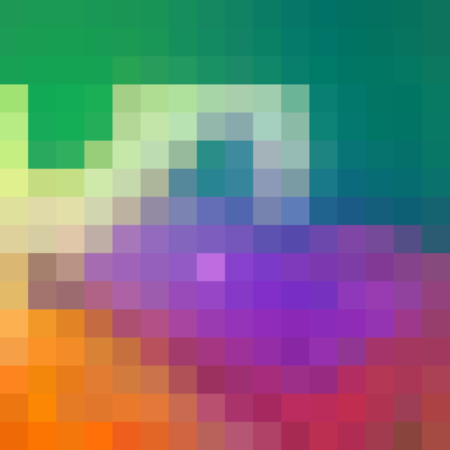} \\ \addlinespace[1ex]

% 14. Shelf-place
    \rotatebox[origin=c]{90}{\scriptsize Shelf-place} &
    \includegraphics[width=1\linewidth, valign=m]{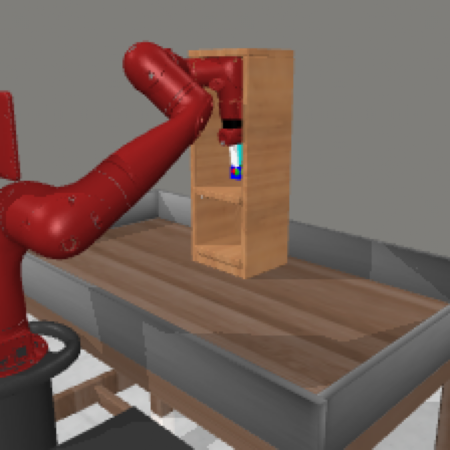} &
    \includegraphics[width=1\linewidth, valign=m]{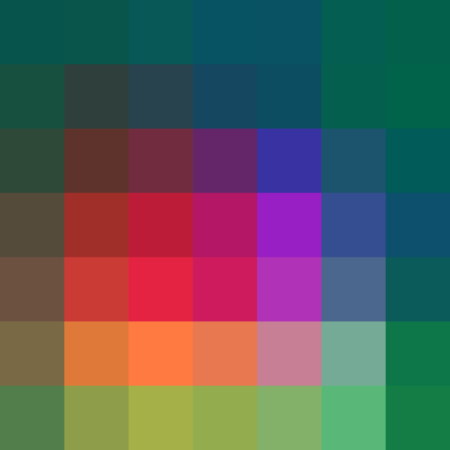} &
    \includegraphics[width=1\linewidth, valign=m]{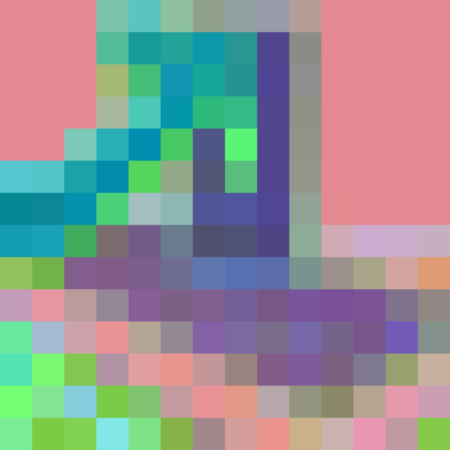} &
    \includegraphics[width=1\linewidth, valign=m]{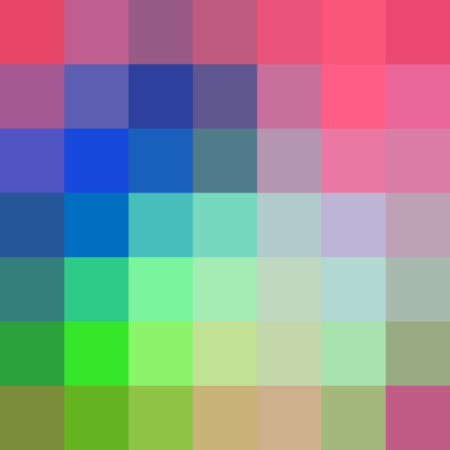} &
    \includegraphics[width=1\linewidth, valign=m]{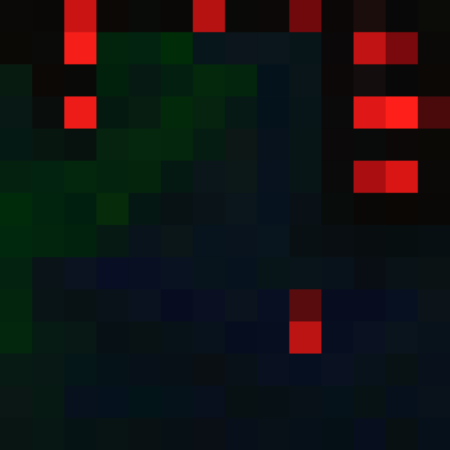} &
    \includegraphics[width=1\linewidth, valign=m]{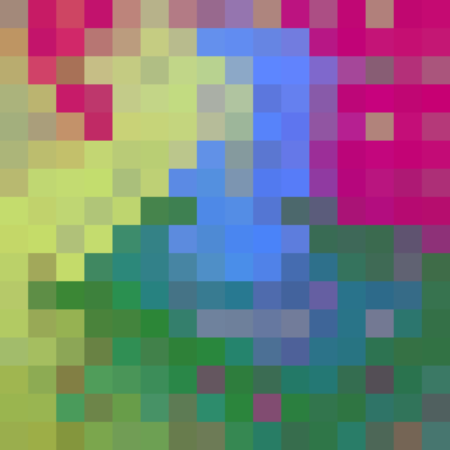} &
    \includegraphics[width=1\linewidth, valign=m]{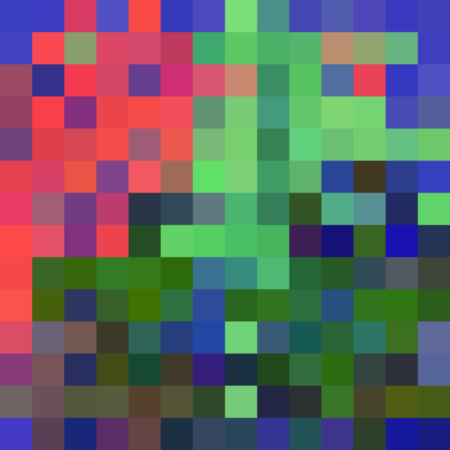} &
    \includegraphics[width=1\linewidth, valign=m]{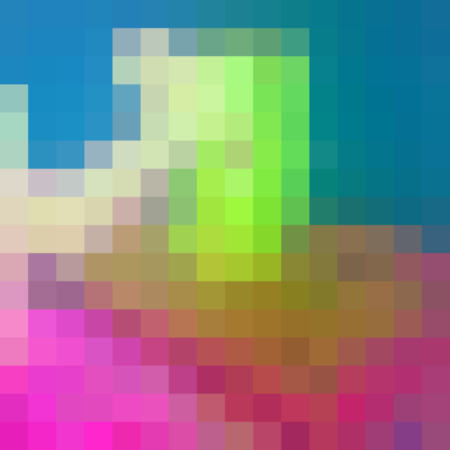} \\ \addlinespace[1ex]

% 15. Sweep-into
    \rotatebox[origin=c]{90}{\scriptsize Sweep-into} &
    \includegraphics[width=1\linewidth, valign=m]{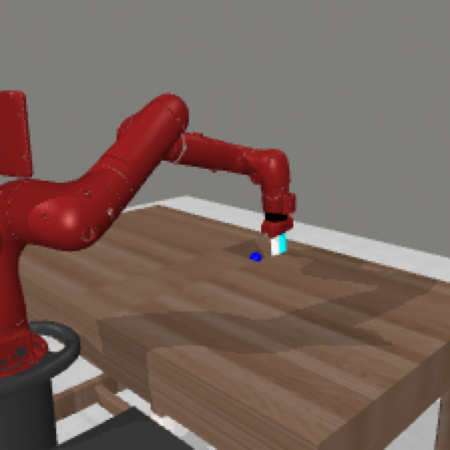} &
    \includegraphics[width=1\linewidth, valign=m]{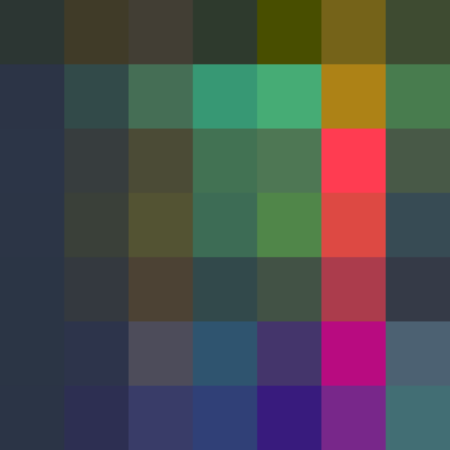} &
    \includegraphics[width=1\linewidth, valign=m]{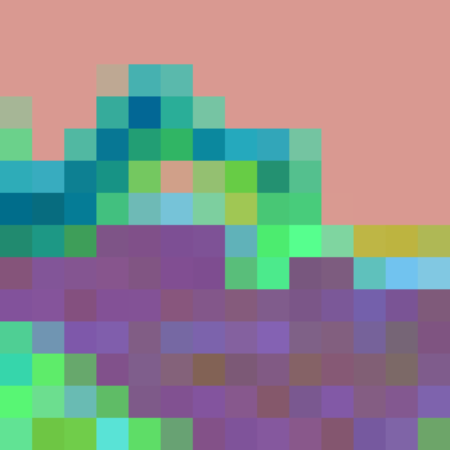} &
    \includegraphics[width=1\linewidth, valign=m]{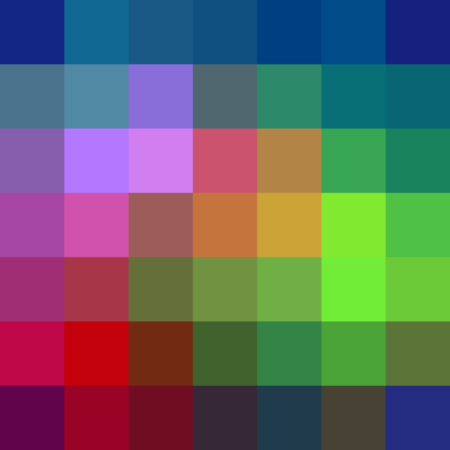} &
    \includegraphics[width=1\linewidth, valign=m]{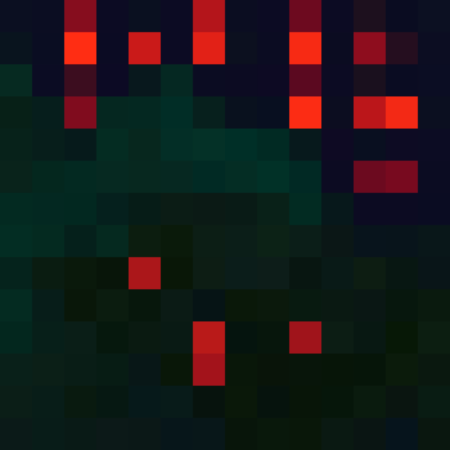} &
    \includegraphics[width=1\linewidth, valign=m]{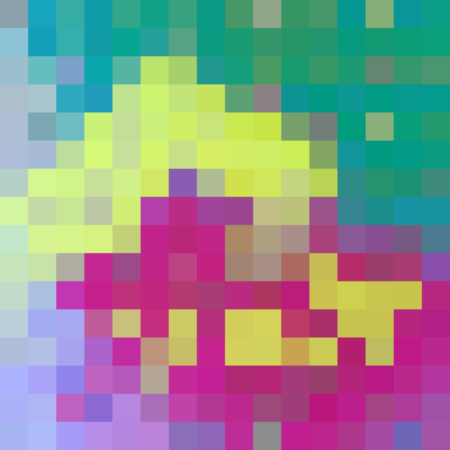} &
    \includegraphics[width=1\linewidth, valign=m]{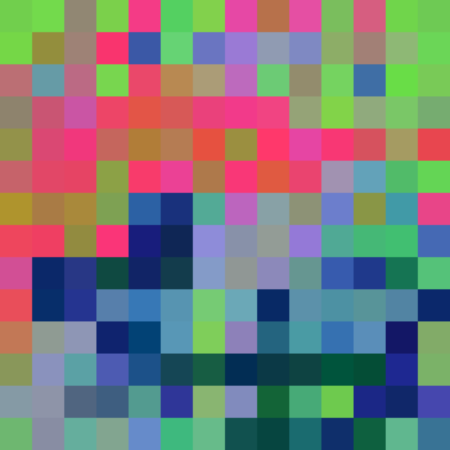} &
    \includegraphics[width=1\linewidth, valign=m]{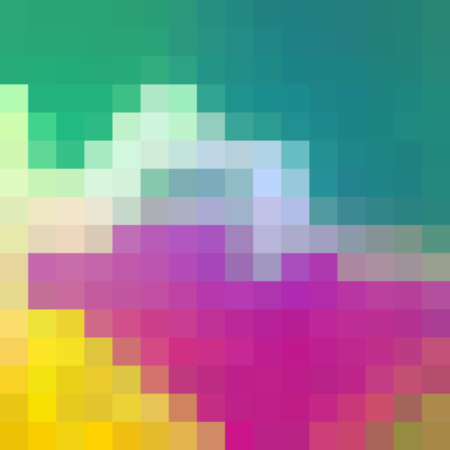} \\ \addlinespace[1ex]

% 16. Close-box
    \rotatebox[origin=c]{90}{\scriptsize Close-box} &
    \includegraphics[width=1\linewidth, valign=m]{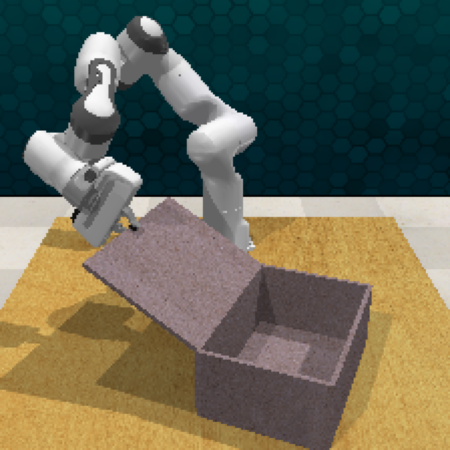} &
    \includegraphics[width=1\linewidth, valign=m]{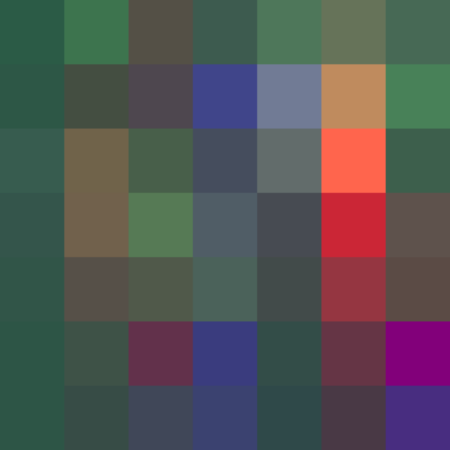} &
    \includegraphics[width=1\linewidth, valign=m]{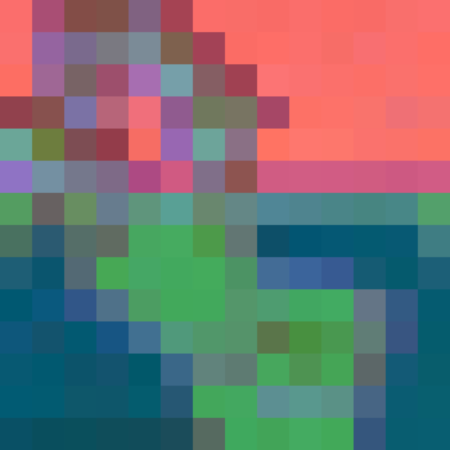} &
    \includegraphics[width=1\linewidth, valign=m]{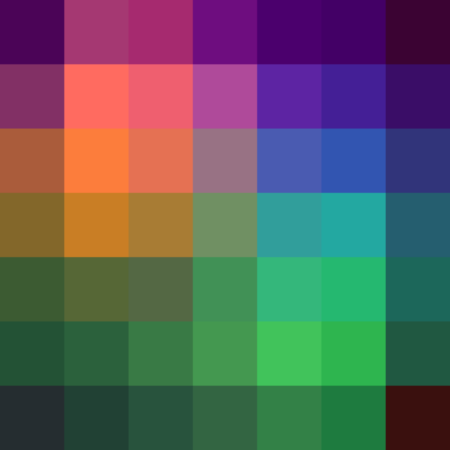} &
    \includegraphics[width=1\linewidth, valign=m]{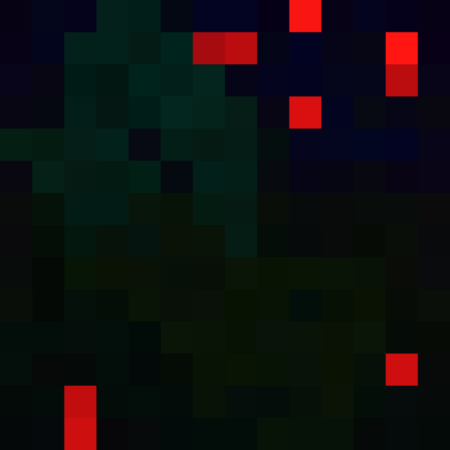} &
    \includegraphics[width=1\linewidth, valign=m]{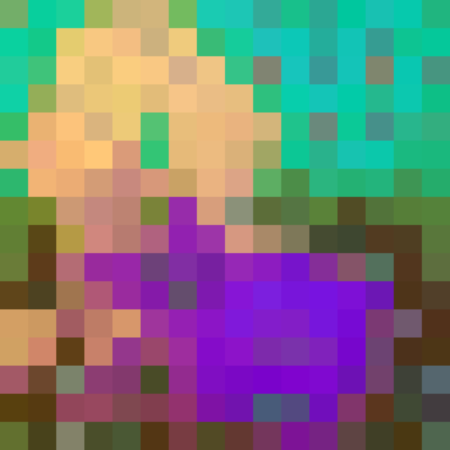} &
    \includegraphics[width=1\linewidth, valign=m]{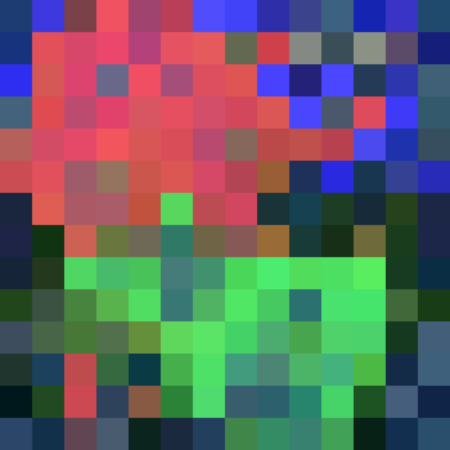} &
    \includegraphics[width=1\linewidth, valign=m]{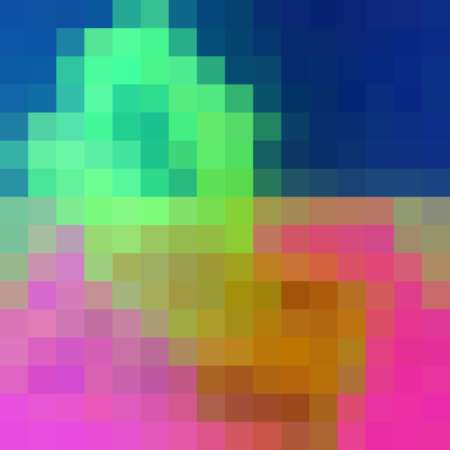} \\ \addlinespace[1ex]

% 17. Close-laptop-lid
    \rotatebox[origin=c]{90}{\scriptsize Close-laptop-lid} &
    \includegraphics[width=1\linewidth, valign=m]{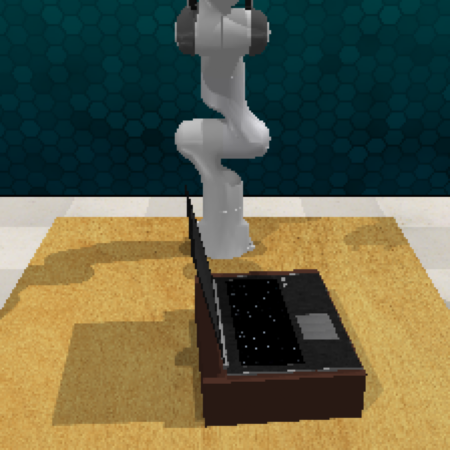} &
    \includegraphics[width=1\linewidth, valign=m]{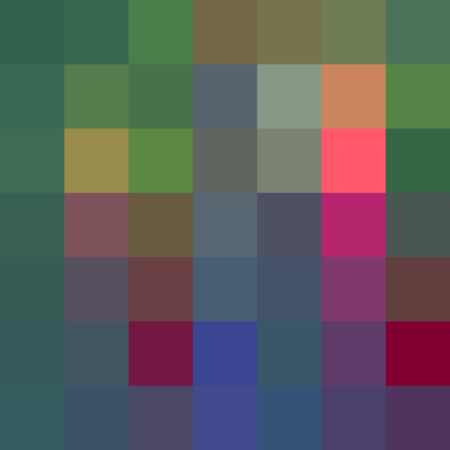} &
    \includegraphics[width=1\linewidth, valign=m]{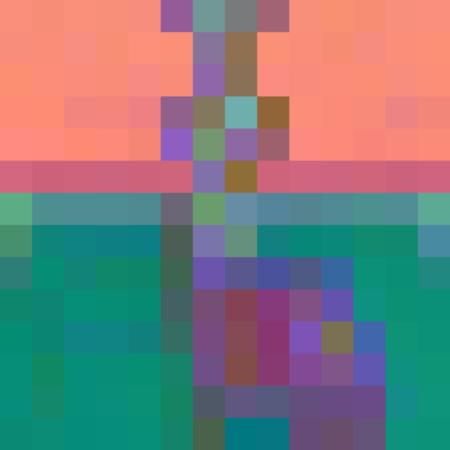} &
    \includegraphics[width=1\linewidth, valign=m]{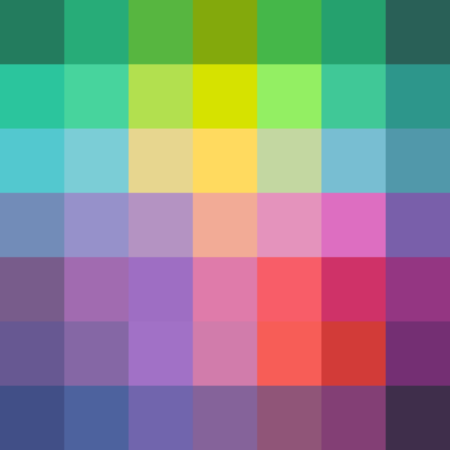} &
    \includegraphics[width=1\linewidth, valign=m]{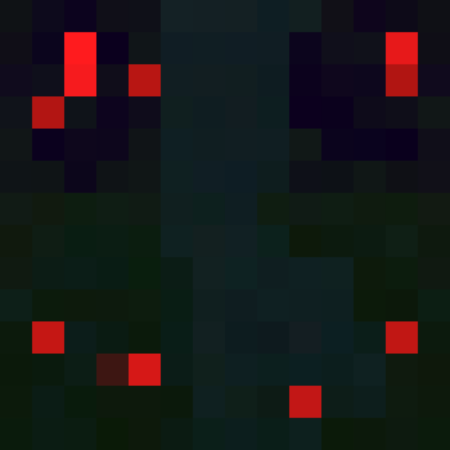} &
    \includegraphics[width=1\linewidth, valign=m]{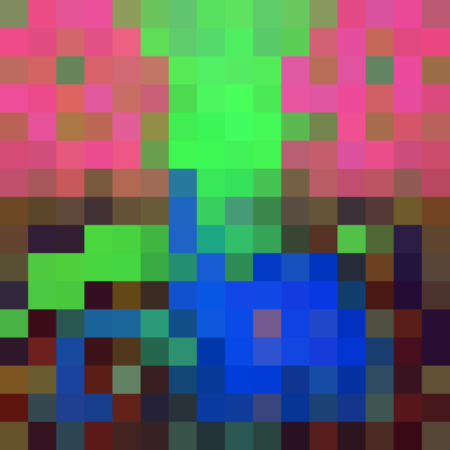} &
    \includegraphics[width=1\linewidth, valign=m]{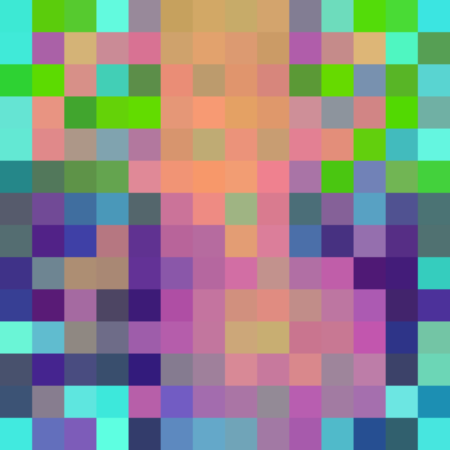} &
    \includegraphics[width=1\linewidth, valign=m]{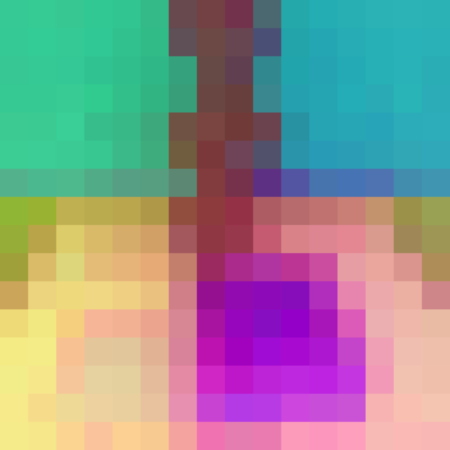} \\ \addlinespace[1ex]

% 18. Unplug-charger
    \rotatebox[origin=c]{90}{\scriptsize Unplug-charger} &
    \includegraphics[width=1\linewidth, valign=m]{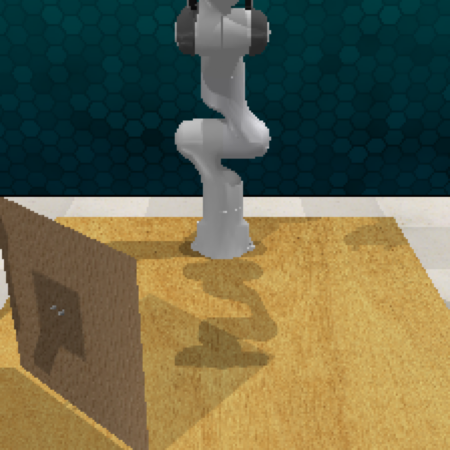} &
    \includegraphics[width=1\linewidth, valign=m]{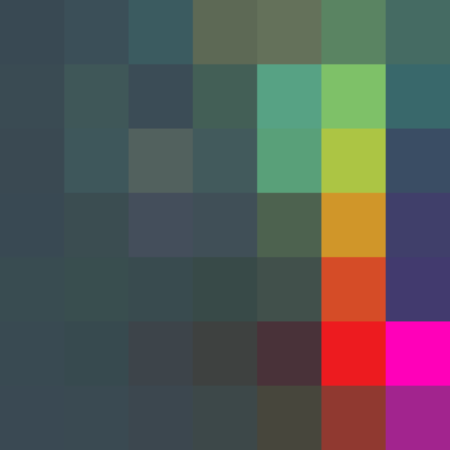} &
    \includegraphics[width=1\linewidth, valign=m]{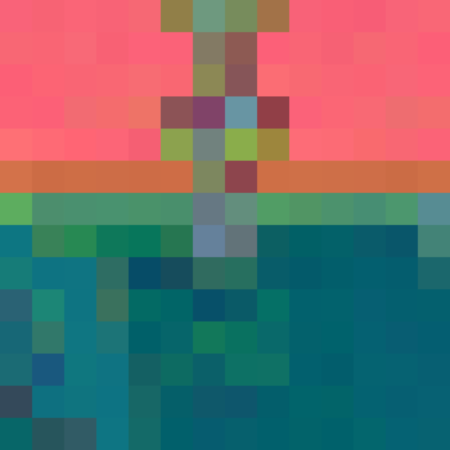} &
    \includegraphics[width=1\linewidth, valign=m]{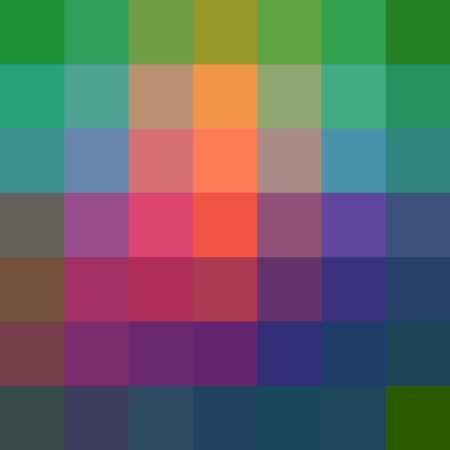} &
    \includegraphics[width=1\linewidth, valign=m]{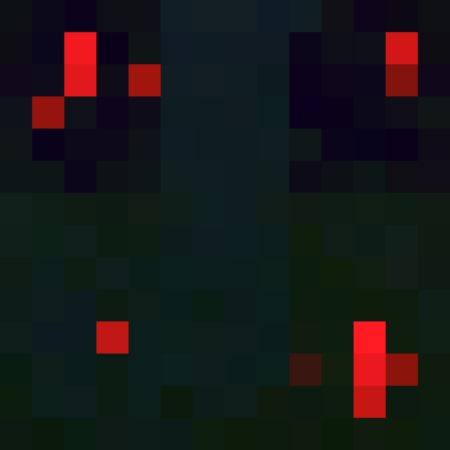} &
    \includegraphics[width=1\linewidth, valign=m]{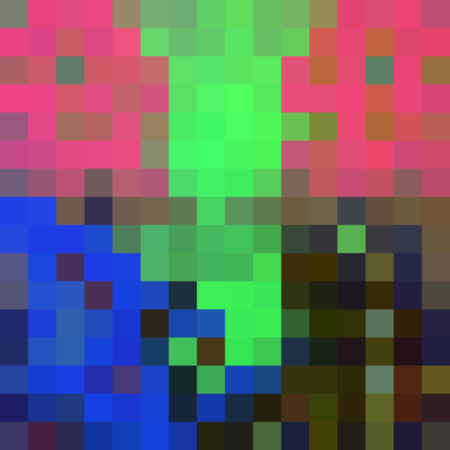} &
    \includegraphics[width=1\linewidth, valign=m]{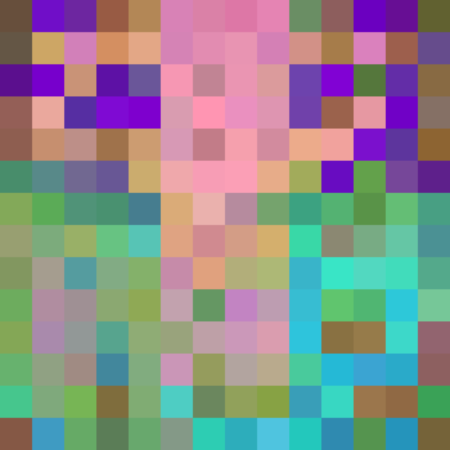} &
    \includegraphics[width=1\linewidth, valign=m]{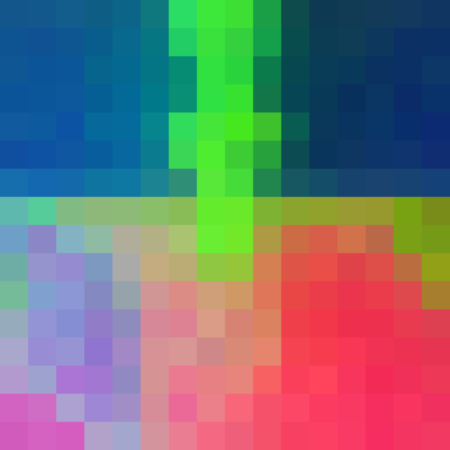} \\ \addlinespace[1ex]

% 19. Water-plants
    \rotatebox[origin=c]{90}{\scriptsize Water-plants} &
    \includegraphics[width=1\linewidth, valign=m]{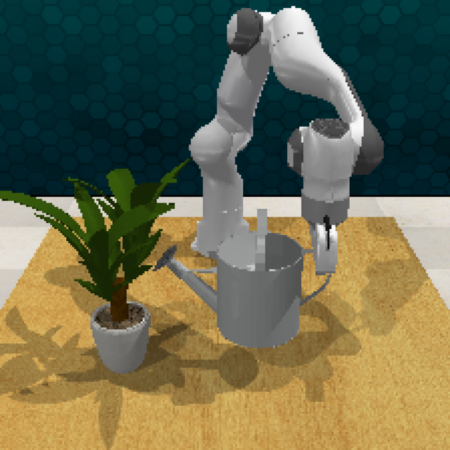} &
    \includegraphics[width=1\linewidth, valign=m]{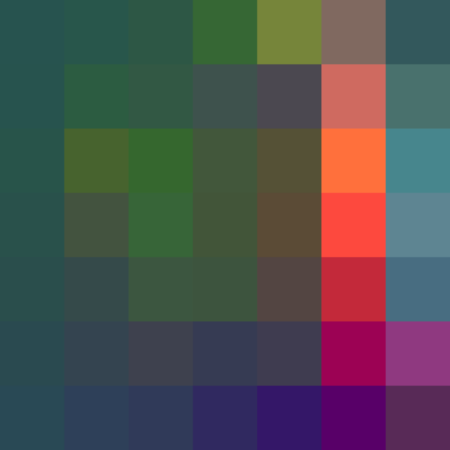} &
    \includegraphics[width=1\linewidth, valign=m]{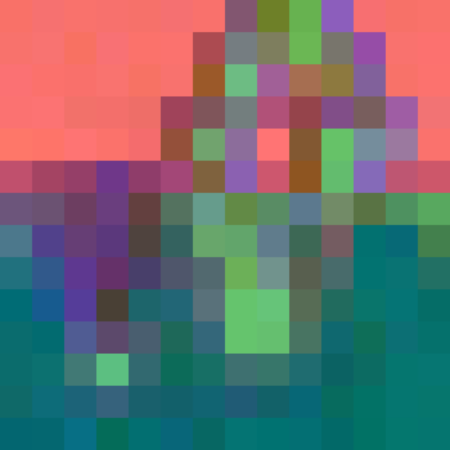} &
    \includegraphics[width=1\linewidth, valign=m]{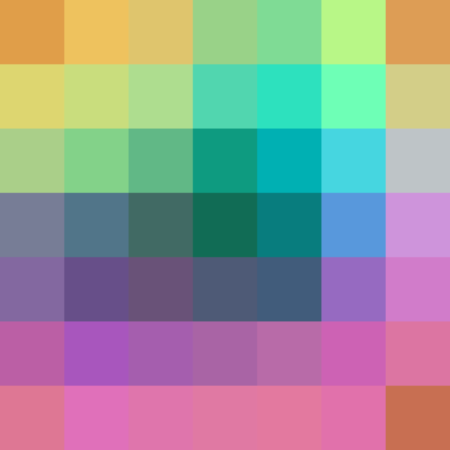} &
    \includegraphics[width=1\linewidth, valign=m]{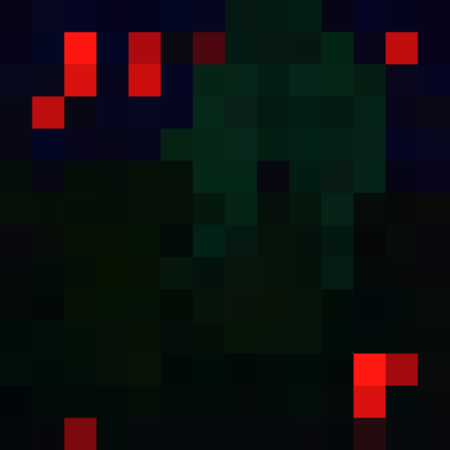} &
    \includegraphics[width=1\linewidth, valign=m]{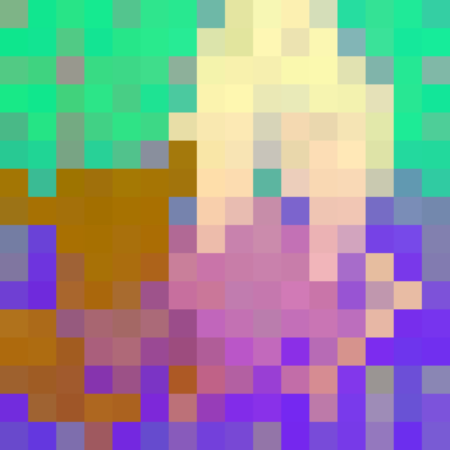} &
    \includegraphics[width=1\linewidth, valign=m]{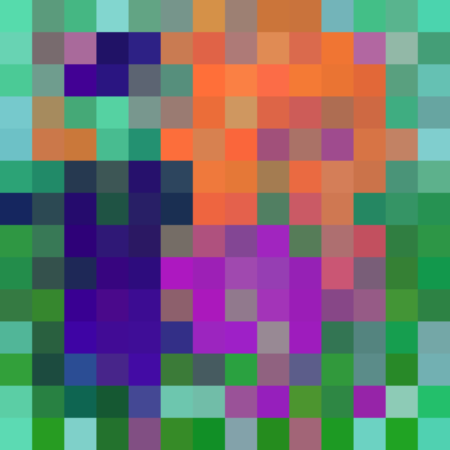} &
    \includegraphics[width=1\linewidth, valign=m]{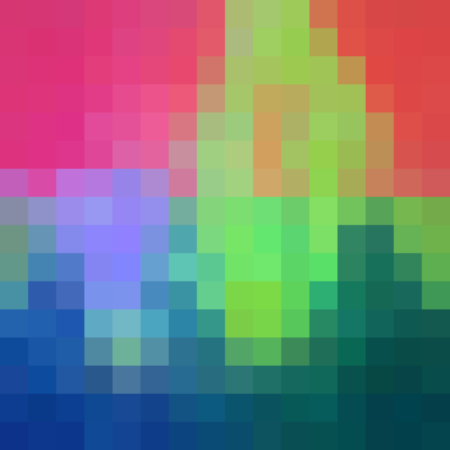} \\ \addlinespace[1ex]
\end{longtable}
}

\end{document}